\documentclass[nohyperref]{article}

\usepackage{microtype}
\usepackage{graphicx}
\usepackage{subfigure}
\usepackage{booktabs} 

\usepackage{hyperref}
\usepackage{caption}
\usepackage{tabularx}

\usepackage[accepted]{icml2022}

\usepackage{amsmath}
\usepackage{amssymb}
\usepackage{mathtools}
\usepackage{amsthm}
\usepackage{bbold}
\usepackage[capitalize,noabbrev]{cleveref}

\usepackage{color}
\usepackage{tcolorbox}
\newcommand{\PY}[1]{{\color{blue}[PY: #1]}}
\newcommand{\SL}[1]{{\color{orange}[SL: #1]}}
\newcommand{\JX}[1]{{\color{magenta}[JX: #1]}}
\newcommand{\MW}[1]{{\color{green}[MW: #1]}}
\theoremstyle{plain}
\newtheorem{theorem}{Theorem}[section]

\newtheorem{lemma}[theorem]{Lemma}

\theoremstyle{definition}

\newtheorem{assumption}[theorem]{Assumption}
\theoremstyle{remark}
\newtheorem{remark}[theorem]{Remark}

\usepackage[textsize=tiny]{todonotes}

\icmltitlerunning{Generalization Guarantee of Training   Graph Convolutional  Networks with  Graph Topology Sampling}

\begin{document}

\twocolumn[
\icmltitle{Generalization Guarantee of Training  Graph Convolutional   Networks   with   Graph Topology Sampling}



\icmlsetsymbol{equal}{*}

\begin{icmlauthorlist}
\icmlauthor{Hongkang Li}{rpi}
\icmlauthor{Meng Wang}{rpi}
\icmlauthor{Sijia Liu}{msu,ibm_mit}
\icmlauthor{Pin-Yu Chen}{ibm}
\icmlauthor{Jinjun Xiong}{buffalo}
\end{icmlauthorlist}

\icmlaffiliation{rpi}{Department of Electrical, Computer, and System Engineering, Rensselaer Polytechnic Institute, NY, USA}
\icmlaffiliation{msu}{Department of Computer Science and Engineering, Michigan State University, MI, USA}
\icmlaffiliation{ibm}{IBM Thomas J. Watson Research Center, Yorktown Heights, NY, USA}
\icmlaffiliation{ibm_mit}{MIT-IBM Watson AI Lab, IBM Research, MA, USA}
\icmlaffiliation{buffalo}{Department of Computer Science and Engineering, University at Buffalo, NY, USA}

\icmlcorrespondingauthor{Hongkang Li}{lih35@rpi.edu}
\icmlcorrespondingauthor{Meng Wang}{wangm7@rpi.edu}
\icmlcorrespondingauthor{Sijia Liu}{liusiji5@msu.edu}
\icmlcorrespondingauthor{Pin-yu Chen}{Pin-Yu.Chen@ibm.com}
\icmlcorrespondingauthor{Jinjun Xiong}{jinjun@buffalo.edu}

\icmlkeywords{Machine Learning, ICML}

\vskip 0.3in
]



\printAffiliationsAndNotice{} %

\begin{abstract}

 
Graph convolutional networks (GCNs) have recently achieved great empirical success in learning graph-structured data. To address its scalability issue due to the recursive embedding of neighboring features, graph topology sampling has been proposed to reduce the memory and computational cost of training GCNs, and  it has achieved comparable test performance to those without topology sampling in many empirical studies. To the best of our knowledge, this paper provides the first theoretical justification of graph topology sampling in training (up to) three-layer GCNs for semi-supervised node classification.  We formally characterize some sufficient conditions on graph topology sampling such that GCN training leads to a diminishing generalization error. Moreover, our method tackles the non-convex interaction  of weights across layers,   which is under-explored in the existing theoretical analyses of GCNs.  This paper characterizes the impact of graph structures and topology sampling on the generalization performance and sample complexity explicitly, and the theoretical findings are also justified through numerical experiments.  
 
\end{abstract}


\newcommand{\figleft}{{\em (Left)}}
\newcommand{\figcenter}{{\em (Center)}}
\newcommand{\figright}{{\em (Right)}}
\newcommand{\figtop}{{\em (Top)}}
\newcommand{\figbottom}{{\em (Bottom)}}
\newcommand{\captiona}{{\em (a)}}
\newcommand{\captionb}{{\em (b)}}
\newcommand{\captionc}{{\em (c)}}
\newcommand{\captiond}{{\em (d)}}

\newcommand{\newterm}[1]{{\bf #1}}

\newtheorem{rmk}{Remark}
\newtheorem{property}{Property}
\def\bfzero{{\boldsymbol{0}}}
\def\bfone{{{\bf1}}}
\def\bfa{{\boldsymbol a}}
\def\bfb{{\boldsymbol b}}
\def\bfc{{\boldsymbol c}}
\def\bfd{{\boldsymbol d}}
\def\bfe{{\boldsymbol e}}
\def\bff{{\boldsymbol f}}
\def\bfg{{\boldsymbol g}}
\def\bfh{{\boldsymbol h}}
\def\bfi{{\boldsymbol i}}
\def\bfj{{\boldsymbol j}}
\def\bfk{{\boldsymbol k}}
\def\bfl{{\boldsymbol l}}
\def\bfm{{\boldsymbol m}}
\def\bfn{{\boldsymbol n}}
\def\bfo{{\boldsymbol o}}
\def\bfp{{\boldsymbol p}}
\def\bfq{{\boldsymbol q}}
\def\bfr{{\boldsymbol r}}
\def\bfs{{\boldsymbol s}}
\def\bft{{\boldsymbol t}}
\def\bfu{{\boldsymbol u}}
\def\bfv{{\boldsymbol v}}
\def\bfw{{\boldsymbol w}}
\def\bfx{{\boldsymbol x}}
\def\bfy{{\boldsymbol y}}
\def\bfz{{\boldsymbol z}}
\def\bfmu{{\boldsymbol \mu}}
\def\bfsg{{\boldsymbol\sigma}}
\def\bfSg{{\boldsymbol\Sigma}}
\def\bfDT{{\boldsymbol\Delta}}
\def\bfLM{{\boldsymbol\Lambda}}

\def\bfA{{\boldsymbol A}}
\def\bfB{{\boldsymbol B}}
\def\bfC{{\boldsymbol C}}
\def\bfD{{\boldsymbol D}}
\def\bfE{{\boldsymbol E}}
\def\bfF{{\boldsymbol F}}
\def\bfG{{\boldsymbol G}}
\def\bfH{{\boldsymbol H}}
\def\bfI{{\boldsymbol I}}
\def\bfJ{{\boldsymbol J}}
\def\bfK{{\boldsymbol K}}
\def\bfL{{\boldsymbol L}}
\def\bfM{{\boldsymbol M}}
\def\bfN{{\boldsymbol N}}
\def\bfO{{\boldsymbol O}}
\def\bfP{{\boldsymbol P}}
\def\bfQ{{\boldsymbol Q}}
\def\bfR{{\boldsymbol R}}
\def\bfS{{\boldsymbol S}}
\def\bfT{{\boldsymbol T}}
\def\bfU{{\boldsymbol U}}
\def\bfV{{\boldsymbol V}}
\def\bfW{{\boldsymbol W}}
\def\bfX{{\boldsymbol X}}
\def\bfY{{\boldsymbol Y}}
\def\bfZ{{\boldsymbol Z}}

\def\figref#1{figure~\ref{#1}}
\def\Figref#1{Figure~\ref{#1}}
\def\twofigref#1#2{figures \ref{#1} and \ref{#2}}
\def\quadfigref#1#2#3#4{figures \ref{#1}, \ref{#2}, \ref{#3} and \ref{#4}}
\def\secref#1{section~\ref{#1}}
\def\Secref#1{Section~\ref{#1}}
\def\twosecrefs#1#2{sections \ref{#1} and \ref{#2}}
\def\secrefs#1#2#3{sections \ref{#1}, \ref{#2} and \ref{#3}}
\def\eqref#1{equation~\ref{#1}}
\def\Eqref#1{Equation~\ref{#1}}
\def\plaineqref#1{\ref{#1}}
\def\chapref#1{chapter~\ref{#1}}
\def\Chapref#1{Chapter~\ref{#1}}
\def\rangechapref#1#2{chapters\ref{#1}--\ref{#2}}
\def\algref#1{algorithm~\ref{#1}}
\def\Algref#1{Algorithm~\ref{#1}}
\def\twoalgref#1#2{algorithms \ref{#1} and \ref{#2}}
\def\Twoalgref#1#2{Algorithms \ref{#1} and \ref{#2}}
\def\partref#1{part~\ref{#1}}
\def\Partref#1{Part~\ref{#1}}
\def\twopartref#1#2{parts \ref{#1} and \ref{#2}}

\def\ceil#1{\lceil #1 \rceil}
\def\floor#1{\lfloor #1 \rfloor}
\def\1{\bm{1}}
\newcommand{\train}{\mathcal{D}}
\newcommand{\valid}{\mathcal{D_{\mathrm{valid}}}}
\newcommand{\test}{\mathcal{D_{\mathrm{test}}}}

\def\eps{{\epsilon}}

\def\reta{{\textnormal{$\eta$}}}
\def\ra{{\textnormal{a}}}
\def\rb{{\textnormal{b}}}
\def\rc{{\textnormal{c}}}
\def\rd{{\textnormal{d}}}
\def\re{{\textnormal{e}}}
\def\rf{{\textnormal{f}}}
\def\rg{{\textnormal{g}}}
\def\rh{{\textnormal{h}}}
\def\ri{{\textnormal{i}}}
\def\rj{{\textnormal{j}}}
\def\rk{{\textnormal{k}}}
\def\rl{{\textnormal{l}}}
\def\rn{{\textnormal{n}}}
\def\ro{{\textnormal{o}}}
\def\rp{{\textnormal{p}}}
\def\rq{{\textnormal{q}}}
\def\rr{{\textnormal{r}}}
\def\rs{{\textnormal{s}}}
\def\rt{{\textnormal{t}}}
\def\ru{{\textnormal{u}}}
\def\rv{{\textnormal{v}}}
\def\rw{{\textnormal{w}}}
\def\rx{{\textnormal{x}}}
\def\ry{{\textnormal{y}}}
\def\rz{{\textnormal{z}}}

\def\rvepsilon{{\mathbf{\epsilon}}}
\def\rvtheta{{\mathbf{\theta}}}
\def\rva{{\mathbf{a}}}
\def\rvb{{\mathbf{b}}}
\def\rvc{{\mathbf{c}}}
\def\rvd{{\mathbf{d}}}
\def\rve{{\mathbf{e}}}
\def\rvf{{\mathbf{f}}}
\def\rvg{{\mathbf{g}}}
\def\rvh{{\mathbf{h}}}
\def\rvu{{\mathbf{i}}}
\def\rvj{{\mathbf{j}}}
\def\rvk{{\mathbf{k}}}
\def\rvl{{\mathbf{l}}}
\def\rvm{{\mathbf{m}}}
\def\rvn{{\mathbf{n}}}
\def\rvo{{\mathbf{o}}}
\def\rvp{{\mathbf{p}}}
\def\rvq{{\mathbf{q}}}
\def\rvr{{\mathbf{r}}}
\def\rvs{{\mathbf{s}}}
\def\rvt{{\mathbf{t}}}
\def\rvu{{\mathbf{u}}}
\def\rvv{{\mathbf{v}}}
\def\rvw{{\mathbf{w}}}
\def\rvx{{\mathbf{x}}}
\def\rvy{{\mathbf{y}}}
\def\rvz{{\mathbf{z}}}

\def\erva{{\textnormal{a}}}
\def\ervb{{\textnormal{b}}}
\def\ervc{{\textnormal{c}}}
\def\ervd{{\textnormal{d}}}
\def\erve{{\textnormal{e}}}
\def\ervf{{\textnormal{f}}}
\def\ervg{{\textnormal{g}}}
\def\ervh{{\textnormal{h}}}
\def\ervi{{\textnormal{i}}}
\def\ervj{{\textnormal{j}}}
\def\ervk{{\textnormal{k}}}
\def\ervl{{\textnormal{l}}}
\def\ervm{{\textnormal{m}}}
\def\ervn{{\textnormal{n}}}
\def\ervo{{\textnormal{o}}}
\def\ervp{{\textnormal{p}}}
\def\ervq{{\textnormal{q}}}
\def\ervr{{\textnormal{r}}}
\def\ervs{{\textnormal{s}}}
\def\ervt{{\textnormal{t}}}
\def\ervu{{\textnormal{u}}}
\def\ervv{{\textnormal{v}}}
\def\ervw{{\textnormal{w}}}
\def\ervx{{\textnormal{x}}}
\def\ervy{{\textnormal{y}}}
\def\ervz{{\textnormal{z}}}

\def\rmA{{\mathbf{A}}}
\def\rmB{{\mathbf{B}}}
\def\rmC{{\mathbf{C}}}
\def\rmD{{\mathbf{D}}}
\def\rmE{{\mathbf{E}}}
\def\rmF{{\mathbf{F}}}
\def\rmG{{\mathbf{G}}}
\def\rmH{{\mathbf{H}}}
\def\rmI{{\mathbf{I}}}
\def\rmJ{{\mathbf{J}}}
\def\rmK{{\mathbf{K}}}
\def\rmL{{\mathbf{L}}}
\def\rmM{{\mathbf{M}}}
\def\rmN{{\mathbf{N}}}
\def\rmO{{\mathbf{O}}}
\def\rmP{{\mathbf{P}}}
\def\rmQ{{\mathbf{Q}}}
\def\rmR{{\mathbf{R}}}
\def\rmS{{\mathbf{S}}}
\def\rmT{{\mathbf{T}}}
\def\rmU{{\mathbf{U}}}
\def\rmV{{\mathbf{V}}}
\def\rmW{{\mathbf{W}}}
\def\rmX{{\mathbf{X}}}
\def\rmY{{\mathbf{Y}}}
\def\rmZ{{\mathbf{Z}}}

\def\ermA{{\textnormal{A}}}
\def\ermB{{\textnormal{B}}}
\def\ermC{{\textnormal{C}}}
\def\ermD{{\textnormal{D}}}
\def\ermE{{\textnormal{E}}}
\def\ermF{{\textnormal{F}}}
\def\ermG{{\textnormal{G}}}
\def\ermH{{\textnormal{H}}}
\def\ermI{{\textnormal{I}}}
\def\ermJ{{\textnormal{J}}}
\def\ermK{{\textnormal{K}}}
\def\ermL{{\textnormal{L}}}
\def\ermM{{\textnormal{M}}}
\def\ermN{{\textnormal{N}}}
\def\ermO{{\textnormal{O}}}
\def\ermP{{\textnormal{P}}}
\def\ermQ{{\textnormal{Q}}}
\def\ermR{{\textnormal{R}}}
\def\ermS{{\textnormal{S}}}
\def\ermT{{\textnormal{T}}}
\def\ermU{{\textnormal{U}}}
\def\ermV{{\textnormal{V}}}
\def\ermW{{\textnormal{W}}}
\def\ermX{{\textnormal{X}}}
\def\ermY{{\textnormal{Y}}}
\def\ermZ{{\textnormal{Z}}}

\def\vzero{{\bm{0}}}
\def\vone{{\bm{1}}}
\def\vmu{{\bm{\mu}}}
\def\vtheta{{\bm{\theta}}}
\def\va{{\bm{a}}}
\def\vb{{\bm{b}}}
\def\vc{{\bm{c}}}
\def\vd{{\bm{d}}}
\def\ve{{\bm{e}}}
\def\vf{{\bm{f}}}
\def\vg{{\bm{g}}}
\def\vh{{\bm{h}}}
\def\vi{{\bm{i}}}
\def\vj{{\bm{j}}}
\def\vk{{\bm{k}}}
\def\vl{{\bm{l}}}
\def\vm{{\bm{m}}}
\def\vn{{\bm{n}}}
\def\vo{{\bm{o}}}
\def\vp{{\bm{p}}}
\def\vq{{\bm{q}}}
\def\vr{{\bm{r}}}
\def\vs{{\bm{s}}}
\def\vt{{\bm{t}}}
\def\vu{{\bm{u}}}
\def\vv{{\bm{v}}}
\def\vw{{\bm{w}}}
\def\vx{{\bm{x}}}
\def\vy{{\bm{y}}}
\def\vz{{\bm{z}}}

\def\evalpha{{\alpha}}
\def\evbeta{{\beta}}
\def\evepsilon{{\epsilon}}
\def\evlambda{{\lambda}}
\def\evomega{{\omega}}
\def\evmu{{\mu}}
\def\evpsi{{\psi}}
\def\evsigma{{\sigma}}
\def\evtheta{{\theta}}
\def\eva{{a}}
\def\evb{{b}}
\def\evc{{c}}
\def\evd{{d}}
\def\eve{{e}}
\def\evf{{f}}
\def\evg{{g}}
\def\evh{{h}}
\def\evi{{i}}
\def\evj{{j}}
\def\evk{{k}}
\def\evl{{l}}
\def\evm{{m}}
\def\evn{{n}}
\def\evo{{o}}
\def\evp{{p}}
\def\evq{{q}}
\def\evr{{r}}
\def\evs{{s}}
\def\evt{{t}}
\def\evu{{u}}
\def\evv{{v}}
\def\evw{{w}}
\def\evx{{x}}
\def\evy{{y}}
\def\evz{{z}}

\def\mA{{\bm{A}}}
\def\mB{{\bm{B}}}
\def\mC{{\bm{C}}}
\def\mD{{\bm{D}}}
\def\mE{{\bm{E}}}
\def\mF{{\bm{F}}}
\def\mG{{\bm{G}}}
\def\mH{{\bm{H}}}
\def\mI{{\bm{I}}}
\def\mJ{{\bm{J}}}
\def\mK{{\bm{K}}}
\def\mL{{\bm{L}}}
\def\mM{{\bm{M}}}
\def\mN{{\bm{N}}}
\def\mO{{\bm{O}}}
\def\mP{{\bm{P}}}
\def\mQ{{\bm{Q}}}
\def\mR{{\bm{R}}}
\def\mS{{\bm{S}}}
\def\mT{{\bm{T}}}
\def\mU{{\bm{U}}}
\def\mV{{\bm{V}}}
\def\mW{{\bm{W}}}
\def\mX{{\bm{X}}}
\def\mY{{\bm{Y}}}
\def\mZ{{\bm{Z}}}
\def\mBeta{{\bm{\beta}}}
\def\mPhi{{\bm{\Phi}}}
\def\mLambda{{\bm{\Lambda}}}
\def\mSigma{{\bm{\Sigma}}}

\newcommand{\tens}[1]{\bm{\mathsfit{#1}}}
\def\tA{{\tens{A}}}
\def\tB{{\tens{B}}}
\def\tC{{\tens{C}}}
\def\tD{{\tens{D}}}
\def\tE{{\tens{E}}}
\def\tF{{\tens{F}}}
\def\tG{{\tens{G}}}
\def\tH{{\tens{H}}}
\def\tI{{\tens{I}}}
\def\tJ{{\tens{J}}}
\def\tK{{\tens{K}}}
\def\tL{{\tens{L}}}
\def\tM{{\tens{M}}}
\def\tN{{\tens{N}}}
\def\tO{{\tens{O}}}
\def\tP{{\tens{P}}}
\def\tQ{{\tens{Q}}}
\def\tR{{\tens{R}}}
\def\tS{{\tens{S}}}
\def\tT{{\tens{T}}}
\def\tU{{\tens{U}}}
\def\tV{{\tens{V}}}
\def\tW{{\tens{W}}}
\def\tX{{\tens{X}}}
\def\tY{{\tens{Y}}}
\def\tZ{{\tens{Z}}}

\def\gA{{\mathcal{A}}}
\def\gB{{\mathcal{B}}}
\def\gC{{\mathcal{C}}}
\def\gD{{\mathcal{D}}}
\def\gE{{\mathcal{E}}}
\def\gF{{\mathcal{F}}}
\def\gG{{\mathcal{G}}}
\def\gH{{\mathcal{H}}}
\def\gI{{\mathcal{I}}}
\def\gJ{{\mathcal{J}}}
\def\gK{{\mathcal{K}}}
\def\gL{{\mathcal{L}}}
\def\gM{{\mathcal{M}}}
\def\gN{{\mathcal{N}}}
\def\gO{{\mathcal{O}}}
\def\gP{{\mathcal{P}}}
\def\gQ{{\mathcal{Q}}}
\def\gR{{\mathcal{R}}}
\def\gS{{\mathcal{S}}}
\def\gT{{\mathcal{T}}}
\def\gU{{\mathcal{U}}}
\def\gV{{\mathcal{V}}}
\def\gW{{\mathcal{W}}}
\def\gX{{\mathcal{X}}}
\def\gY{{\mathcal{Y}}}
\def\gZ{{\mathcal{Z}}}

\def\sA{{\mathbb{A}}}
\def\sB{{\mathbb{B}}}
\def\sC{{\mathbb{C}}}
\def\sD{{\mathbb{D}}}
\def\sF{{\mathbb{F}}}
\def\sG{{\mathbb{G}}}
\def\sH{{\mathbb{H}}}
\def\sI{{\mathbb{I}}}
\def\sJ{{\mathbb{J}}}
\def\sK{{\mathbb{K}}}
\def\sL{{\mathbb{L}}}
\def\sM{{\mathbb{M}}}
\def\sN{{\mathbb{N}}}
\def\sO{{\mathbb{O}}}
\def\sP{{\mathbb{P}}}
\def\sQ{{\mathbb{Q}}}
\def\sR{{\mathbb{R}}}
\def\sS{{\mathbb{S}}}
\def\sT{{\mathbb{T}}}
\def\sU{{\mathbb{U}}}
\def\sV{{\mathbb{V}}}
\def\sW{{\mathbb{W}}}
\def\sX{{\mathbb{X}}}
\def\sY{{\mathbb{Y}}}
\def\sZ{{\mathbb{Z}}}

\def\emLambda{{\Lambda}}
\def\emA{{A}}
\def\emB{{B}}
\def\emC{{C}}
\def\emD{{D}}
\def\emE{{E}}
\def\emF{{F}}
\def\emG{{G}}
\def\emH{{H}}
\def\emI{{I}}
\def\emJ{{J}}
\def\emK{{K}}
\def\emL{{L}}
\def\emM{{M}}
\def\emN{{N}}
\def\emO{{O}}
\def\emP{{P}}
\def\emQ{{Q}}
\def\emR{{R}}
\def\emS{{S}}
\def\emT{{T}}
\def\emU{{U}}
\def\emV{{V}}
\def\emW{{W}}
\def\emX{{X}}
\def\emY{{Y}}
\def\emZ{{Z}}
\def\emSigma{{\Sigma}}

\newcommand{\etens}[1]{\mathsfit{#1}}
\def\etLambda{{\etens{\Lambda}}}
\def\etA{{\etens{A}}}
\def\etB{{\etens{B}}}
\def\etC{{\etens{C}}}
\def\etD{{\etens{D}}}
\def\etE{{\etens{E}}}
\def\etF{{\etens{F}}}
\def\etG{{\etens{G}}}
\def\etH{{\etens{H}}}
\def\etI{{\etens{I}}}
\def\etJ{{\etens{J}}}
\def\etK{{\etens{K}}}
\def\etL{{\etens{L}}}
\def\etM{{\etens{M}}}
\def\etN{{\etens{N}}}
\def\etO{{\etens{O}}}
\def\etP{{\etens{P}}}
\def\etQ{{\etens{Q}}}
\def\etR{{\etens{R}}}
\def\etS{{\etens{S}}}
\def\etT{{\etens{T}}}
\def\etU{{\etens{U}}}
\def\etV{{\etens{V}}}
\def\etW{{\etens{W}}}
\def\etX{{\etens{X}}}
\def\etY{{\etens{Y}}}
\def\etZ{{\etens{Z}}}

\newcommand{\pdata}{p_{\rm{data}}}
\newcommand{\ptrain}{\hat{p}_{\rm{data}}}
\newcommand{\Ptrain}{\hat{P}_{\rm{data}}}
\newcommand{\pmodel}{p_{\rm{model}}}
\newcommand{\Pmodel}{P_{\rm{model}}}
\newcommand{\ptildemodel}{\tilde{p}_{\rm{model}}}
\newcommand{\pencode}{p_{\rm{encoder}}}
\newcommand{\pdecode}{p_{\rm{decoder}}}
\newcommand{\precons}{p_{\rm{reconstruct}}}

\newcommand{\laplace}{\mathrm{Laplace}} 

\newcommand{\E}{\mathbb{E}}
\newcommand{\Ls}{\mathcal{L}}
\newcommand{\R}{\mathbb{R}}
\newcommand{\emp}{\tilde{p}}
\newcommand{\lr}{\alpha}
\newcommand{\reg}{\lambda}
\newcommand{\rect}{\mathrm{rectifier}}
\newcommand{\softmax}{\mathrm{softmax}}
\newcommand{\sigmoid}{\sigma}
\newcommand{\softplus}{\zeta}
\newcommand{\KL}{D_{\mathrm{KL}}}
\newcommand{\Var}{\mathrm{Var}}
\newcommand{\standarderror}{\mathrm{SE}}
\newcommand{\Cov}{\mathrm{Cov}}
\newcommand{\normlzero}{L^0}
\newcommand{\normlone}{L^1}
\newcommand{\normltwo}{L^2}
\newcommand{\normlp}{L^p}
\newcommand{\normmax}{L^\infty}

\newcommand{\parents}{Pa} 

\let\ab\allowbreak

\section{Introduction}\label{sec:intro}

Graph convolutional neural networks (GCNs) 
aggregate the embedding of each node 
with the embedding of
its neighboring nodes in each layer. 
GCNs can  model graph-structured data more accurately and compactly than conventional neural networks and have demonstrated great empirical advantage in text analysis \citep{HYL17, KW17, VCCRLB18, PPQT17}, computer vision \citep{SE18, WYG18, HGZD18}, recommendation systems \citep{YHCEHL18, BKW18}, physical reasoning \citep{BPLR16, SHSM18}, 
and biological science
\citep{DMIBH15}. Such empirical success is often  achieved at a cost of higher computational and memory costs, especially for large graphs, because the embedding of one node depends recursively on the neighbors.   To alleviate the exponential increase of computational cost in training deep GCNs, various graph topology sampling methods  
  have been proposed  to only aggregate  the embeddings of a   selected subset of neighbors in training GCNs.  
 Node-wise neighbor-sampling methods such as  GraphSAGE \citep{HYL17}, VRGCN \citep{CZS18}, and Cluster-GCN \citep{CLSL19} sample  a subset of neighbors for each node. Layer-wise importance sampling methods such as FastGCN \citep{CMX18} and LADIES \citep{ZHWJ19} sample a fixed number of nodes for each layer based on the   estimate of   node importance.  Another line of works such as \citep{ZZCS20, LZTJ20, CSCZ21} employ graph sparsification or pruning to reduce the computational and memory cost. 
 Surprisingly, these sampling methods often have comparable or even better testing performance compared to   training with the original graph in many empirical studies   \cite{CMX18, CSCZ21}. 
 
 
 In contrast to the empirical success, the theoretical foundation of training GCNs with graph sampling is much less investigated. Only \citet{CRM21} analyzes the convergence rate of graph sampling, but no generalization analysis is provided. 
 One fundamental question about training GCNs is still vastly open, which is:
 
\begin{center}
 \textit{Under what conditions does a GCN learned with graph topology sampling   achieve satisfactory generalization?}
\end{center}

\textbf{Our contributions:}  
To the best of our knowledge, this paper provides the first generalization analysis of training GCNs with graph topology sampling.  We focus on semi-supervised node classification problems where, with all node features and partial node labels, the objective is to predict unknown node labels. We summarize our contributions from the following dimensions. 

 \textit{First}, 
 this paper proposes a training framework that implements both stochastic gradient descent (SGD) and graph topology sampling, and   the learned GCN model with  Rectified Linear Unit (ReLU) activation is guaranteed to approach  the best generalization performance of a large class of target functions. Moreover, as the number of labeled nodes and the number of neurons increase, the class of target function enlarges, indicating   improved   generalization.  

 \textit{Second}, this paper explicitly characterizes the impact of graph topology sampling on the generalization performance through the proposed \textit{effective adjacency matrix} $\bfA^*$ of a directed graph that models the node correlations.  $\bfA^*$ depends on both the given normalized graph adjacency matrix in GCNs and the graph sampling strategy. We provide the general insights 
 that (1) if a node is sampled with a low frequency, its impact on other nodes is reduced in $\bfA^*$ compared with $\bfA$; (2) graph sampling on a highly-unbalanced $\bfA$, where some nodes have a dominating impact in the graph, results in a more balanced $\bfA^*$. Moreover, these insights apply to other graph sampling methods such as FastGCN~\citep{CMX18}.

We show that learning with topology sampling  has the same generalization performance as training GCNs using $\bfA^*$. Therefore, a satisfactory generalization can still be achieved even when the number of sampled nodes is small, provided that the resulting $\bfA^*$ still characterizes the data correlations properly. This is the first theoretical explanation of the empirical success of graph topology sampling.  
 
 \textit{Third}, this paper shows that the required number of labeled nodes, referred to as the sample complexity, is   a polynomial of $\|\bfA^*\|_\infty$ and the maximum  node degree, where $\|\cdot\|_\infty$ measures the maximum absolute row sum. Moreover, our sample complexity is only logarithmic in the number of neurons $m$ and consistent with the practical over-parameterization  of GCNs,  in contrast to  the loose bound of poly($m$) in \citep{ZWLC20} in the restrictive setting of two-layer  (one-hidden-layer) GCNs without graph topology sampling.

\subsection{Related Works}


\textbf{Generalization analyses of GCNs without graph sampling}. Some recent works analyze GCNs trained on the original graph. 
\citet{XHLJ19, CRM21} characterize the expressive power of GCNs. 
   \citet{XZJK21} analyzes the convergence of gradient descent 
   in training linear GCNs.
  \citet{L21, LUZ21, GJJ20, OS20}  characterize the generalization gap, which is the difference  between the training error and testing error,  through Rademacher complexity. \citet{VZ19, CRM21, ZW21}  analyze  the generalization gap of  training GCNs using   SGD 
  via the notation of algorithmic stability.
  
  To analyze the training error and generalization performance simultaneously, \citet{DHPSWX19} uses the   neural tangent kernel (NTK) approach,
 where the neural network width is infinite and the step size is infinitesimal, shows that the training error is zero, and characterizes the generalization bound.  
\citet{ZWLC20} proves that gradient descent can learn a model with zero population risk, provided that all data are generated by an unknown target model. The result in  \citep{ZWLC20}  is limited to two-layer GCNs   and requires a proper initialization in the local convex region of the optimal solution. 

\textbf{Generalization analyses of feed-forward neural networks.} The NTK approach was first developed to analyze fully connected neural networks (FCNNs), see, e.g., \citep{JGH18}. The works of \citet{ZSJB17, FCL20, LZW22}  analyze one-hidden-layer neural networks with Gaussian input data.  \citet{D17} analyzes multi-layer FCNNs but focuses on training the last layer only, while the changes in the hidden layers are negligible.  \citet{ALL19} 
provides  the optimization and generalization of three-layer FCNNs. 
Our proof framework is built upon \citep{ALL19} but makes two important technical contributions. First, this paper provides the first generalization analysis of  graph topology sampling in training GCNs, while \citet{ALL19} considers FCNNs with neither graph topology nor graph  sampling. Second, \citet{ALL19}  considers i.i.d. training samples, while this paper considers semi-supervised GCNs where the training data are correlated through graph convolution. 


\subsection{Notations}


Vectors are in bold lowercase, matrices and tensors in
are bold uppercase. Scalars are in normal fonts. 
For instance, $\bfZ$ is a matrix, and $\bfz$ is a vector. 
$z_i$ denotes the $i$-th entry of $\bfz$, and $Z_{i,j}$ denotes the $(i,j)$-th entry of $\bfZ$. 
$[K]$ ($K>0$) denotes the set including integers from $1$ to $K$. $\bfI_d\in\mathbb{R}^{d\times d}$ and $\bfe_i$ represent the identity matrix in $\mathbb{R}^{d \times d}$ and the $i$-th standard basis vector, respectively. 
We denote the column $\ell_p$ norm for $\bfW\in\mathbb{R}^{d\times N}$ (for $p\geq1$) as
\begin{equation}
    \|\bfW\|_{2,p}=(\sum_{i\in[m]}\|\bfw_i\|_2^p)^\frac{1}{p}
\end{equation}
Hence, $\|\bfW\|_{2,2}=\|\bfW\|_F$ is the Frobenius norm of $\bfW$. We use $\bfw_i$ ($\tilde{\bfw}_i$) to denote the $i$-th column (row) vector of $\bfW$. 
We follow the convention that $f(x)=O(g(x))$ (or $\Omega(g(x))$, $\Theta(g(x)))$ means that $f(x)$ increases at most (or at least, or in the same, respectively,) order of $g(x)$.  With high probability (w.h.p.) means with probability 
$1-e^{-c\log^2(m_1, m_2)}$ for a sufficient large constant $c$ 
where $m_1$ and $m_2$ are the number of neurons in the two hidden layers. 

\textbf{Function complexity. }
For any  smooth function $\phi(z)$ with its  power series representation as $\phi(z)=\sum_{i=0}^\infty c_i z^i$,   define two useful parameters  as follows,  
\begin{equation} 
    \mathcal{C}_\epsilon(\phi, R)=\sum_{i=0}^\infty \Big((C^*R)^i+(\frac{\sqrt{\log(1/\epsilon)}}{\sqrt{i}}C^*R)^i\Big)|c_i|\label{C_e}
\end{equation}
\begin{equation} 
    \mathcal{C}_s(\phi, R)=C^*\sum_{i=0}^\infty (i+1)^{1.75} R^i|c_i|\label{C_s}
\end{equation}
where $R\geq0$ and $C^*$ is a  sufficiently  large constant. 
These two quantities are used 
in the model complexity and sample complexity, which represent the required number of model parameters and  training samples to learn $\phi$ up to $\epsilon$ error, respectively.
Many population functions have bounded complexity. For instance, if $\phi(z)$ is $\exp(z)$, $\sin(z)$, $\cos(z)$ or polynomials of $z$, then $\mathcal{C}_\epsilon(\phi,O(1))\leq O(\text{poly}(1/\epsilon))$ and $\mathcal{C}_s(\phi, O(1))\leq O(1)$. 

The main notations   are summarized in Table  \ref{tab:notation} in Appendix.


\section{Training GCNs with   Topology Sampling: Formulation and Main Components}


\textbf{GCN setup.} Let $\mathcal{G}=\{\mathcal{V},\mathcal{E}\}$ denote an un-directed graph, where $\mathcal{V}$ is the set of nodes with size $|\mathcal{V}|=N$ and $\mathcal{E}$ is the set of edges. Let $\tilde{\bfA}\in\{0,1\}^{N\times N}$ be the adjacency matrix of $\mathcal{G}$ with added self-connections. Let $\bfD$ be the degree matrix with diagonal elements $D_{i,i}=\sum_{j} \tilde{A}_{i, j}$ and zero entries otherwise. $\bfA$ denotes the normalized adjacency matrix with $\bfA=\bfD^{-\frac{1}{2}}\tilde{\bfA}\bfD^{-\frac{1}{2}}$. 
Let $\bfX\in\mathbb{R}^{N\times d}$ denote the matrix of the features of $N$ nodes, where the $n$-th row of $\bfX$, denoted by $\Tilde{\bfx}_n \in \mathbb{R}^{1\times d}$, represents the feature of node $n$. 
Assume $\|\Tilde{\bfx}_n\|=1$ for all $n$ without loss of generality.
  $y_n\in \mathcal{Y}$ represents the label of node $n$, where $\mathcal{Y}$ is a set of all labels.  $y_n$ depends on not only $\bfx_n$ but the neighbors. 
Let $\Omega \subset \mathcal{V}$ denote the set of labeled nodes. 
Given $\bfX$ and  labels in $\Omega$, 
the objective of semi-supervised node-classification is to predict the unknown labels in $\mathcal{V}/\Omega$.

\textbf{Learner network} We consider the setting of training a three-layer  GCN  $F: \mathbb{R}^N\times   \mathbb{R}^{N\times d}\rightarrow \mathbb{R}^{1\times K}$ with
\begin{equation}\label{eqn:threelayer}
\begin{aligned}
F_{\bfA}(\bfe_g,\bfX;\bfW,\bfV)&=\bf\bfe_g^\top \bfA\sigma(  \bfr+\bfB_2)\bfC~\text{~and~} \\
\bfr &= \bfA\sigma(\bfA\bfX\bfW+\bfB_1)\bfV 
\end{aligned}
\end{equation}
where $\sigma(x)= \max(x,0)$ is the ReLU activation function, 
$\bfW\in\mathbb{R}^{d\times m_1}$ and $\bfV\in\mathbb{R}^{m_1\times m_2}$ represent the weights of $m_1$ and $m_2$ hidden nodes in the first and second layer, respectively.   $\bfB_1\in\mathbb{R}^{N\times m_1}$  and  $\bfB_2\in\mathbb{R}^{m_1\times m_2}$ represent the bias matrices.  $\bfC\in\mathbb{R}^{m\times K}$ is the output weight vector.   $\bfe_g\in\mathbb{R}^N$ 
  belongs to $\{\bfe_i\}_{i=1}^N$ and selects the index of the node label. 
We write $F$ 
as $F_{\bfA}(\bfe_g,\bfX; \bfW, \bfV)$, 
because we only update $\bfW$ and $\bfV$ in training, and $\bfA$ represents the graph topology.  
Note that in conventional GCNs such as \cite{KW17}, $\bfC$ is 
a learnable parameter, and $\bfB_1$ and $\bfB_2$ can be zero. Here for the analytical purpose, we consider a slightly different model   where   $\bfC$, $\bfB_1$ and $\bfB_2$ are fixed as randomly selected values. 

Consider a loss function $L: \mathbb{R}^{1\times k}\times \mathcal{Y} \rightarrow \mathbb{R}$ such that for every $y\in\mathcal{Y}$, the function $L(\cdot, y)$ is nonnegative, convex, 1-Lipschitz continuous and 1-Lipschitz smooth and $L(0, y)\in[0, 1]$. This includes both the cross-entropy loss and the $\ell_2$-regression loss (for bounded $\mathcal{Y}$). 
The learning problem solves the following empirical risk minimization problem:
\begin{equation}\label{eqn:risk}
\min_{\bfW, \bfV}
L_{\Omega}(\bfW, \bfV)= \frac{1}{|\Omega|}\sum_{i \in \Omega} L(F_\bfA(\bfe_i,\bfX; \bfW, \bfV), y^i)
\end{equation}
where $L_{\Omega}$ is the empirical risk    of the labeled nodes in  $\Omega$.  The trained weights are used to estimate the unknown labels  on $\mathcal{V}/\Omega$.
Note that the results in this paper are \textit{distribution-free}, and no assumption is made on the distributions of $\tilde{x}_n$ and $y_n$.

%

\textbf{Training with SGD}. In practice, (\ref{eqn:risk}) is often solved by gradient type of methods, where   in iteration $t$, the currently estimations 
are updated  by subtracting the product of a positive step size and the gradient of  $L_{\Omega}$ evaluated at the current estimate. 
To reduce the computational complexity in estimating the gradient,  an SGD method is often employed to compute the gradient of the risk of a  randomly selected subset of $\Omega$ 
rather than using the whole set $\Omega$. 

However, due to the recursive embedding of neighboring features in GCNs, see the concatenations of $\bfA$ in (\ref{eqn:threelayer}), the computation and memory cost  of computing 
the gradient can be high. Thus, graph topology sampling methods have been proposed to further reduce the computational cost.

\textbf{Graph topology sampling}. A  node sampling method randomly removes a   subset of nodes and the incident edges from $\mathcal{G}$ in each iteration independently, 
and the  embedding aggregation is based on  the reduced graph.  Mathematically,  in iteration $s$, replace $\bfA$ in (\ref{eqn:threelayer})  with\footnote{Here we use the same sampled matrix $\bfA^{s}$ in all three layers in (\ref{eqn:threelayer})  to simplify the  representation. Our analysis applies to the more general setting that each layer uses a different sampled adjacency matrix, i.e.,
the three $\bfA$ matrices in (\ref{eqn:threelayer})  are replaced with $\bfA^{s(1)}=\bfA\bfP^{s(1)},\ \bfA^{s(2)}=\bfA\bfP^{s(2)},\ \bfA^{s(3)}=\bfA\bfP^{s(3)}$, respectively, as in \citep{ZHWJ19, RCMS20}, where $\bfP^{s(1)}$, $\bfP^{s(2)}$, and $\bfP^{s(3)}$ are independently sampled following the same sampling strategy. } $\bfA^s=\bfA\bfP^s$, where $\bfP^s$ is a diagonal matrix, and the $i$th diagonal entry 
 is 0, if node $i$ is removed in iteration $s$. The non-zero diagonal entries of  $\bfP^s$  are selected differently based on different sampling methods. 
 Because $\bfA^s$ is  much more sparse than $\bfA$, the computation and memory cost of embedding neighboring features is significantly reduced. 
 
 This paper will analyze the generalization performance, i.e., the prediction accuracy of unknown labels, of 
 our algorithm framework that implements both SGD and graph topology sampling to solve (\ref{eqn:risk}). The details of our algorithm are discussed in Section \ref{sec:sampling}-\ref{sec:algorithm}, and the generalization performance is presented in Section \ref{sec:result}.

\section{Main Algorithmic and Theoretical Results}


\subsection{Informal Key Theoretical Findings}

We first summarize the main insights of our results before presenting them formally. 

\textbf{1. A provable generalization guarantee of GCNs beyond  two layers and with graph topology sampling.} 
The learned GCN by our Algorithm \ref{alg: 1} can approach the best performance of label prediction using a large class of target functions. 
Moreover, the prediction performance improves when the number of labeled nodes and the number of neurons $m_1$ and $m_2$ increase.  This is the first generalization performance guarantee of training GCNs with graph topology sampling.  

\textbf{2. The explicit characterization of the impact of graph sampling through the effective adjacency matrix $\bfA^*$}. We show that training with graph sampling returns a model that has the same label prediction performance as that of a model trained by replacing $\bfA$ with $\bfA^*$ in (\ref{eqn:threelayer}), where $\bfA^*$ depends on both $\bfA$ and the graph sampling strategy. As long as $\bfA^*$ can characterize the correlation among nodes properly, the learned GCN maintains a desirable prediction performance. This explains the empirical success of graph topology sampling in many datasets. 


\textbf{3. The explicit  sample complexity bound on graph properties}. We provide explicit bounds  on  
the sample complexity and the required number of neurons, both of which grow as the node correlation increase. 
Moreover, the sample complexity depends on the number of 
neurons 
only logarithmically, which is consistent with the practical over-parameterization. To the best of our knowledge,    \citep{ZWLC20} is the only existing work that provides a sample complexity bound based on the graph topology,   but  in the non-practical and restrictive setting of two-layer GCNs. Moreover, the sample complexity bound by  \citep{ZWLC20} is polynomial in the number of neurons.

\textbf{4. Tackling the non-convex interaction of weights between different layers}. The convexity plays a critical role in many exiting analyses of GCNs. For instance, 
the analyses in \cite{ZWLC20}  require 
a special initialization in the local convex region of the global minimum, and the  results  
only apply to two-layer GCNs. The NTK approach in \cite{DHPSWX19} considers the limiting case that the interactions across layers are negligible. Here, we directly address the non-convex interaction of weights $\bfW$ and $\bfV$ in both algorithmic design and theoretical analyses.

\subsection{Graph Topology Sampling Strategy} \label{sec:sampling}

Here we describe our   graph topology sampling strategy using $\bfA^s$, which we randomly generate  to replace $\bfA$ in the $s$th SGD iteration. Although our method is motivated for analysis and   different from the existing graph sampling strategies, our insights   generalize to other sampling methods like FastGCN \citep{CMX18}.  The outline of our algorithmic framework of training GCNs with graph sampling is deferred to Section \ref{sec:algorithm}. 

 Suppose the node  degrees in $\mathcal{G}$ can be divided into $L$ groups with $L \geq 1$, where the degrees of nodes in group $l$ are in the order of $d_l$, i.e., between $cd_l$ and $Cd_l$ for some constants $c \leq C$, and $d_l$ is order-wise smaller than $d_{l+1}$, i.e.,  $d_l=o(d_{l+1})$. Let $N_l$ denote the number of nodes in group $l$.

 \textbf{Graph sampling strategy\footnote{Here we discuss asymmetric sampling as a general case.  The special case of symmetric sampling is introduced in Section \ref{sec: symmetric_sampling}}.}.  We consider a group-wise uniform sampling strategy, where $S_l$ out of $N_l$ nodes are sampled uniformly from each group $l$.  For all unsampled nodes, we set the corresponding diagonal entries of a  diagonal matrix $\bfP^s$  to be zero. 
If node $i$ is sampled in this iteration and  belongs to group $l$ for any $i$ and $l$, the $i$th diagonal entry of  $\bfP^s$ is set as $p^*_lN_l/S_l$ for some non-negative constant $p^*_l$. Then  $\bfA^s=\bfA\bfP^s$.  $N_l/S_l$ can be viewed as the scaling to compensate for the unsampled nodes in group $l$. $p^*_l$ can be viewed as the  scaling to reflect the impact of sampling on nodes with different importance that will be discussed in detail soon.  

\textbf{Effective adjacency matrix $\bfA^*$ by graph sampling.}
To analyze  the impact of graph topology sampling on the learning performance, we define the effective  adjacency matrix as follows:   
 \begin{equation}\label{eqn:Astar}
     \bfA^{*}=\bfA \bfP^* 
 \end{equation}
where $\bfP^*$ is  a diagonal matrix defined  as 
 \begin{equation}\label{eqn:pstar}
 \bfP^*_{ii}=p^*_l  \quad \textrm{ if node } i \textrm{ belongs to degree group } l
 \end{equation}
Therefore, compared with $\bfA$, all the columns with indices corresponding to group $l$ are scaled by a factor of $p^*_l$. We will formally analyze the impact of graph topology sampling on the generalization performance in Section \ref{sec:result}, but an intuitive understanding is that our graph sampling strategy effectively changes the normalized adjacency matrix $\bfA$ in the GCN network model (\ref{eqn:threelayer})
to $\bfA^*$. 

$\bfA^*$ can be viewed as an adjacency matrix of a weighted directed graph $\mathcal{G'}$ that reflects the node correlations, where each un-directed edge in $\mathcal{G}$ corresponds to two directed edges in $\mathcal{G'}$ with possibly different weights.  
 $\bfA^*_{ji}$ measures the impact of the feature of node $i$   on the label of node $j$. 
 If $p^*_l$ is in the range of $(0,1)$,  the corresponding entries  of columns with indices in group $l$ in $\bfA^*$ are smaller than those in $\bfA$. That means  the impact of a node in group $l$ on all other nodes is reduced from those in $\bfA$.   Conversely, if $p^*_l>1$, then the impact of nodes in group $l$ 
 in $\bfA^*$ is enhanced from that in $\bfA$.

\textbf{Parameter selection and insights} 

(1) The scaling factor  $p^*_l$ should satisfy
\begin{equation}\label{eqn:pstar}
0 \leq p^*_l  \leq \frac{c_1}{L\psi_l}, \quad \forall l
\end{equation}
for a positive constant $c_1$ that can be sufficiently large. 
  $\psi_l$ is defined as follows, 
 \begin{equation}\label{eqn:psi}
 \psi_l := \frac{\sqrt{d_Ld_l}\bfN_l}{\sum_{i=1}^L d_i \bfN_i} \quad \quad
    \forall l\in [L]
\end{equation}

Note that (\ref{eqn:pstar}) is a minor requirement for most graphs.  
  To see this, suppose $L$ is a constant, and every $N_l$ is in the order of $N$. Then $\psi_l$ is less than $O(1)$ for all $l$. 
Thus, all constant values  of $p^*_{\hat{l}}$
  satisfy (\ref{eqn:pstar}) with $\psi_l$ from (\ref{eqn:psi}). A special example is that $p^*_l$ are all equal, i.e., $\bfA^*=c_2\bfA$ for some constant $c_2$. Because one can scale $\bfW$ and $\bfV$ by $1/c_2$ in (\ref{eqn:threelayer}) without changing the results, 
  $\bfA^*$ is equivalent to $\bfA$ in this case.


The upper bound in  (\ref{eqn:psi}) 
only becomes active in highly unbalanced graphs where there exists a dominating group $\hat{l}$ such that $\sqrt{d_{\hat{l}}} N_{\hat{l}}\gg  \sqrt{d_l}N_l$ for all other $l$. Then the upper bound of $p^*_{\hat{l}}$ is much smaller than those for other $p^*_l$. Therefore, the columns of $\bfA^*$ that correspond to group $\hat{l}$ are scaled down more significantly than other columns, indicating that the impact of group $\hat{l}$   is reduced more significantly than other groups in $\bfA^*$. Therefore, the takeaway is that  \textbf{graph topology sampling reduces the impact of dominating nodes more than other nodes, resulting in a more balanced $\bfA^*$ compared with $\bfA$}. 


(2) The number of sampled nodes shall satisfy 
\begin{equation}\label{eqn:ll}
 \frac{S_l}{N_l} \geq \ (1+ \frac{c_1 \text{poly}(\epsilon)}{Lp^*_l \psi_l})^{-1} \quad \quad \forall l\in[L]
\end{equation}
where $\epsilon$ is a small positive value. 
The sampling requirement in (\ref{eqn:ll}) has two takeaways. \textbf{First, the higher-degree groups shall be sampled more frequently than lower-degree groups.} 
To see this, consider a special case that 
$p^*_l=1$, and $N_l=N/L$   for all $l$. 
Then  (\ref{eqn:ll})  indicates that $S_l$ is larger in a group $l$ with a larger $d_l$. 
This  intuition is the same   as FastGCN \citep{CMX18}, which also samples high-degree nodes with a higher probability in many cases. Therefore, the insights from our graph sampling method also apply to other sampling methods such as FastGCN. We will show the connection to FastGCN empirically in Section \ref{sec:varyA}. 
\textbf{Second,  reducing the number of samples in group $l$ corresponds to reducing the impact of group $l$ in $\bfA^*$.} To see this, 
note that decreasing  $p^*_l$ reduces the right-hand side of   (\ref{eqn:ll}).  

\subsection{The  Algorithmic Framework of Training  GCNs}\label{sec:algorithm}
Because (\ref{eqn:risk})  is non-convex,   solving it directly using SGD can get stuck at  a bad local minimum in theory. The main idea in the theoretical analysis to address this non-convexity is to add weight decay and regularization in the objective of (\ref{eqn:risk})  such that with a proper regularization, any second-order critical point is \textit{almost} a global minimum.

\begin{algorithm}[t]

\caption{Training with SGD and graph topology sampling}

\begin{algorithmic}[1]\label{alg: 1}
\STATE{\textbf{Input: }} 
Normalized adjacency matrix $\bfA$, node features $\bfX$, known node labels in $\Omega$, the step size $\eta$, the number of inner iterations $T_w$, the number of outer iterations $T$,   $\sigma_w$, $\sigma_v$, $\lambda_w$, $\lambda_v$.
\STATE Initialize
  $\bfW^{(0)}$, $\bfV^{(0)}$, $\bfB_1$, $\bfB_2$, $\bfC$.  
\STATE $\bfW_0=0$, $\bfV_0=0$.
\FOR{$t=0,1,\cdots,T-1$}
\STATE Apply noisy SGD with step size $\eta$ on the stochastic objective 
$\hat{L}_\Omega(\lambda_{t};\bfW,\bfV)$ in (\ref{eqn:hatL}) for   $T_w$ steps. To generate  the stochastic objective in each step $s$, 
randomly sample a batch of labeled nodes $\Omega^s$ from $\Omega$; generate $\bfA^{s}$ using graph sampling; randomly generate $\bfW^\rho$, $\bfV^\rho$ and $\boldsymbol{\Sigma}$. \\
Let the starting point be $\bfW=\bfW_t$, $\bfV=\bfV_t$ and suppose it reaches $\bfW_{t+1}$ and $\bfV_{t+1}$.
\STATE $\lambda_{t+1}=\lambda_t\cdot(1-\eta)$.
\ENDFOR
\STATE{\textbf{Output:}}\\ $\bfW^{(out)}=\sqrt{\lambda_{T-1}}(\bfW^{(0)}+\bfW^{\rho}+\bfW_{T} \boldsymbol{\Sigma})$ \\ $\bfV^{(out)}=\sqrt{\lambda_{T-1}}(\bfV^{(0)}+\bfV^{\rho}+ \boldsymbol{\Sigma} \bfV_{T})$.
\end{algorithmic}
\end{algorithm}


 Specifically, for initialization,  
 entries of $\bfW^{(0)}$ are i.i.d.    from $\mathcal{N}(0,\frac{1}{m_1})$, and entries of $\bfV^{(0)}$ are i.i.d.     from $\mathcal{N}(0,\frac{1}{m_2})$. 
 $\bfB_1$ (or $\bfB_2$) is initialized to be an all-one vector multiplying a row vector with i.i.d. samples from    $\mathcal{N}(0,\frac{1}{m_1})$  (or $\mathcal{N}(0, \frac{1}{m_2})$). Entries of $\bfC$ are  drawn i.i.d.   from $\mathcal{N}(0, 1)$. 
 
 

 In each outer loop $t=0, ..., T-1$, we use noisy   SGD\footnote{Noisy SGD is vanilla SGD plus Gaussian perturbation. It is a common trick  in the theoretical analyses of non-convex optimization  \citep{GHJY15} and is not needed in practice.} with step size $\eta$ for $T_w$ iterations to minimize the stochastic objective function $\hat{L}_\Omega$ in (\ref{eqn:hatL}) with some fixed $\lambda_{t-1}$, where $\lambda_0=1$, and the weight decays with $\lambda_{t+1}=(1-\eta)\lambda_{t}$. 
 {\small
\begin{equation}\label{eqn:hatL}
\begin{aligned}
    &\hat{L}_\Omega(\lambda_{t};\bfW,\bfV)\\
    =&L_\Omega( \sqrt{\lambda_{t}}(\bfW^{(0)}+\bfW^\rho+\bfW\boldsymbol{\Sigma}), \sqrt{\lambda_{t}}(\bfV^{(0)}+\bfV^\rho+\boldsymbol{\Sigma}\bfV))\\
    &+\lambda_w\|\sqrt{\lambda_t}\bfW\|_{2,4}^4+\lambda_v\|\sqrt{\lambda_t}\bfV\|_F^2
\end{aligned}
\end{equation}}%
 $\hat{L}_\Omega(\lambda_{t};\bfW,\bfV)$ is stochastic because in each inner iteration $s$, (1) we randomly sample a subset $\Omega^s$ of labeled nodes; (2) we randomly sample $\bfA^s$ from the graph topology sampling method in Section \ref{sec:sampling}; (3)
 $\bfW^{\rho}$ and $\bfV^{\rho}$ are small perturbation matrices with entries i.i.d. drawn from $\mathcal{N}(0, \sigma_w^2)$ and $\mathcal{N}(0, \sigma_v^2)$, respectively; and (4) $\boldsymbol{\Sigma} \in \mathbb{R}^{m_1\times m_1}$ is   a random diagonal matrix with diagonal entries uniformly drawn from $\{1, -1\}$. 
  $\bfW^{\rho}$ and $\bfV^{\rho}$ are standard Gaussian smoothing 
 in the literature
of theoretical analyses of non-convex optimization, see, e.g. \citep{GHJY15},  and are not needed in practice.  
 $\boldsymbol{\Sigma}$ is similar to the practical Dropout \citep{SHK14} technique that randomly masks out neurons and is also introduced for the theoretical analysis only.

The last two terms in (\ref{eqn:hatL}) are additional regularization terms
for some positive $\lambda_w$ and $\lambda_v$. 
As shown in \citep{ALL19}, $\|\cdot\|_{2,4}$ is used for the analysis to drive the weights to be evenly distributed among neurons. The practical regularization $\|\cdot\|_F$ has the same effect in empirical results, while the theoretical justification is open. 

Algorithm \ref{alg: 1} summarizes the algorithm with the parameter selections in Table \ref{tab:table1}.
Let $\bfW^{out}$ and $\bfV^{out}$ denote the returned weights. 
We use $F_{\bfA^*}(\bfe_i,\bfX;\bfW^{out},\bfV^{out})$ to predict the label of node $i$. 
This might sound different from the conventional practice which uses $\bfA$ in predicting unknown labels. However, note that $\bfA^*$ only differs from $\bfA$ by  a column-wise scaling as from (\ref{eqn:Astar}). Moreover, $\bfA^*$ can be set as $\bfA$ in many practical datasets based on our discussion after (\ref{eqn:psi}). Here we use the general form of $\bfA^*$ for the purpose of analysis.


We remark that our framework of algorithm and analysis can be easily applied to  the simplified setup of two-layer GCNs. 
The resulting algorithm is much simplified to a  vanilla SGD plus graph topology sampling. All the additional components   above are introduced
   to address the non-convex interaction of $\bfW$ and $\bfV$ theoretically and may not be needed for practical implementation. We skip the discussion of two-layer GCNs in this paper.

\begin{table}[h!]
  \begin{center}
        \caption{Parameter choices for Algorithm \ref{alg: 1}}    \label{tab:table1}
    \begin{tabular}{l|c|l|c} 
 \hline
 $\lambda_v$ & $\scriptstyle 2\epsilon_0 m_2/m_1^{1-0.01}$ & $\sigma_v$ & $\scriptstyle 1/m_2^{1/2+0.01}$\\
 \hline
 $\lambda_w$ & $\scriptstyle 2\epsilon_0 m_1^{3-0.002}/C_0^4$ & $\sigma_w$ & $\scriptstyle 1/m_1^{1-0.01}$\\
 \hline
  $C$ & $\scriptstyle \mathcal{C}_\epsilon(\phi,\|\bfA^*\|_\infty)\sqrt{\|\bfA^*\|_\infty^2+1}$ & $C'$ & $\scriptstyle 10C\sqrt{p_2}$\\
 \hline
  $C''$ & $\scriptstyle \mathcal{C}_\epsilon(\Phi,C')\sqrt{\|\bfA^*\|_\infty^2+1}$ &
  $C_0$ & $\scriptstyle \tilde{O}(p_1^2p_2K^2 CC'')$\\
 \hline
    \end{tabular}

  \end{center}
\end{table}

\subsection{Generalization Guarantee}\label{sec:result}


Our formal generalization analysis shows that our learning method returns a GCN model 
  that approaches the minimum prediction error that can be achieved  by the best function in a large concept class of target functions, which have two  important properties: (1) the prediction error decreases as size of the function class increases; 
  and (2) the concept class uses $\bfA^*$ in (\ref{eqn:Astar})  as the adjacency matrix of the graph topology. Therefore, the result implies that if $\bfA^*$ accurately captures the correlations among node features and labels, the learned GCN model can achieve a small prediction error of unknown labels. Moreover,  no other functions in a large concept class can perform better than the learned GCN model.  
To formalize the results, we first define the target functions as follows.

\textbf{Concept class and target function $F^*$}. 
Consider a concept class consisting of target functions $F^*: \mathbb{R}^N\times  \mathbb{R}^{N\times d}\rightarrow \mathbb{R}^{1\times K}$:
\begin{equation}\label{eqn:concept2}
\begin{aligned}
F^*_{\bfA^*} (\bfe_g, \bfX)=\bfe_g^\top \bfA^*\big(\Phi (\bfr_1) \odot \bfr_2\big) \bfC^* \\  
\bfr_1= \bfA^* \phi_1 (\bfA^*\bfX \bfW_1^*)\bfV_1^*\\
\bfr_2=\bfA^* \phi_2(\bfA^* \bfX \bfW_2^*)\bfV_2^*
\end{aligned}
  \end{equation}
 \noindent
  where   $\phi_{1}$, $\phi_{2}$, $\Phi$: $\mathbb{R}\rightarrow\mathbb{R}$ all infinite-order smooth\footnote{When $\Phi$ is operated on a matrix $\bfr_1$, $\Phi(\bfr_1)$ means applying $\Phi$ on each entry of $\bfr_1$. In fact, our results still hold for a more general case that a different function $\Phi_j$ is applied to every entry of the $j$th column of  $\bfr_1$, $j\in [p_2]$. We keep the simpler model to have a more compact representation. The similar arguments hold for $\phi_{1}$, $\phi_{2}$.  }. 
  The parameters $\bfW_{1}^*, \bfW_{2}^*\in\mathbb{R}^{d \times p_2}$,  $\bfV_{1}^*, \bfV_{2}^*\in\mathbb{R}^{p_2\times p_1}$,  $\bfC^*\in\mathbb{R}^{p_1 \times k}$  satisfy that every column of $\bfW_{1}^*$, $\bfW_{2}^*$, $\bfV_{1}^*$, $\bfV_{2}^*$ is unit norm, and the maximum absolute value of $\bfC^*$ is at most $1$.   The effective adjacency matrix  $\bfA^*$ is defined in (\ref{eqn:Astar}). 
      Define 
      \begin{align}\label{eqn:ce1}
       \mathcal{C}_\epsilon(\phi, R) = \max\big(\mathcal{C}_\epsilon(\phi_1, R),\mathcal{C}_\epsilon(\phi_2, R)\big),
       \\
      \label{eqn:cs1}
         \mathcal{C}_s(\phi, R) =\max\big(\mathcal{C}_s(\phi_1, R),\mathcal{C}_2(\phi_1, R)\big).
   \end{align}
  
    We  focus on target functions where the function complexity
    $\mathcal{C}_\epsilon(\Phi, R)$, $\mathcal{C}_s(\Phi, R)$, $\mathcal{C}_\epsilon(\phi, R)$, $\mathcal{C}_s(\phi, R)$, defined in (\ref{C_e})-(\ref{C_s}), (\ref{eqn:ce1})-(\ref{eqn:cs1}), as well as $p_1$ and $p_2$, are all bounded.   
  


  (\ref{eqn:concept2}) is more general than GCNs. If $\bfr_2$ is a constant matrix,   (\ref{eqn:concept2})  models a GCN, where $\bfW^*_1$ and $\bfV^*_1$ are 
  weight matrices in the first and second layer, respectively, and $\phi_1$ and $\Phi$ are the activation functions in each layer.

  \textbf{Modeling the prediction error of unknown labels.} We will show that the learned GCN by our method performs almost the same as the best function in the concept class in (\ref{eqn:concept2})
in predicting unknown labels.   Because the practical datasets usually contain noise in features and labels,  
  we employ a probabilistic model to model the data. Note that our result is  distribution-free , and the following distributions are introduced for the presentation of the results. 
  
  Specifically, let $\mathcal{D}_{\tilde{x}_n}$ denote the distribution from which  the feature $\tilde{x}_n$ of node $n$  is drawn. 
  For example, when the noise level is low, $\mathcal{D}_{\tilde{x}_n}$ can be a distribution centered at the observed feature of node $n$ with a  small variance. Similarly,  let $\mathcal{D}_{y_n}$ denote the distribution from which the label $y_n$ at node $n$  is drawn. Let   $\bfe_g$ 
 be uniformly selected   from $\{\bfe_i\}_{i=1}^N\in \mathbb{R}^N$. Let $\mathcal{D}$ denote the concatenation of   these distributions of a data point 
\begin{equation}
z=(\bfe_g, \bfX, y)\in \mathbb{R}^N \times \mathbb{R}^{N\times d}\times\mathcal{Y}. 
\end{equation}

Then the given feature matrix $\bfX$ and partial labels in   $\Omega$     can be viewed as  $|\Omega|$    identically distributed but \textit{correlated} samples from $\mathcal{D}$. 
The correlation results from the fact that the label of node $i$ depends on not only the feature of node $i$ but also neighboring features. 
This model of correlated samples is different from the conventional assumption of i.i.d. samples in supervised learning and makes our analyses more involved.  

 Let 
\begin{equation}
\mathrm{OPT}_{\bfA^*}= \mathop{\rm{\min}}_{\bfW^*_1,\ \bfW^*_2, \atop
\bfV^*_1,\ \bfV^*_2,\ \bfC^*}\mathbb{E}_{(\bfe_g, \bfX, y)\sim \mathcal{D}} L (F^*_{\bfA^*}(\bfe_g, \bfX), y)
\end{equation}
be the smallest population risk achieved by the best target function (over the choices of $\bfW^*_1$, $\bfW^*_2$, $\bfV^*_1$, $\bfV^*_2$, $\bfC^*$) in the concept class  $F^*_{\bfA^*}$ in  (\ref{eqn:concept2}). $\textrm{OPT}_{\bfA^*}$ measures the average loss of predicting the unknown labels if the estimates are computed using the best target function in  (\ref{eqn:concept2}). 
Clearly, $\textrm{OPT}_{\bfA^*}$   decreases as the size of  the concept increases, i.e., when $p_1$ and $p_2$ increase. Moreover, if $\bfA^*$ indeed models the node correlations accurately, $\textrm{OPT}_{\bfA^*}$ can be very small, indicating a desired generalization performance.
We next show that the population risk of the learned GCN model by our method can be arbitrarily close to $\textrm{OPT}_{\bfA^*}$.

\begin{theorem}
\label{thm: node_3}

For 
every $\epsilon_0\in(0,\frac{1}{100}]$, every $\epsilon\in(0,(Kp_1p_2^2\mathcal{C}_s(\Phi, p_2\mathcal{C}_s(\phi,O(1)))
\mathcal{C}_s(\phi,O(1))\\
\cdot\|\bfA^*\|_\infty^2)^{-1}\epsilon_0)$, as long as
\begin{equation}\label{eqn:neuron}
\begin{aligned}
&m_1=m_2=m\\
\geq  &\text{poly}\Big(\mathcal{C}_\epsilon\big(\Phi, \mathcal{C}_\epsilon(\phi,O(1))\big),p_2, \|\bfA^*\|_\infty, \frac{1}{\epsilon}\Big)
\end{aligned}
\end{equation}
\begin{equation}\label{eqn:complexity}
\begin{aligned}
  |\Omega| \geq   &\Theta (\epsilon_0^{-2}\|\bfA^*\|_\infty^8K^6(1+p_1^4p_2^5\mathcal{C}_\epsilon(\Phi, \sqrt{p_2}\mathcal{C}_\epsilon(\phi, O(1)))\\
  &\cdot \mathcal{C}_\epsilon(\phi,O(1))(\|\bfA^*\|_\infty+1)^4)(1+\delta)^4\log N \log m), 
\end{aligned}
\end{equation}
(\ref{eqn:pstar}) and (\ref{eqn:ll}) 
hold,    
there is a choice $\eta=1/\text{poly}(\|\bfA^*\|_\infty, K, m)$ and $T=\text{poly}(\|\bfA^*\|_\infty, K, m)$ such that with probability at least $ 0.99$ 
,
 \begin{equation}
 \begin{aligned}
     &\mathbb{E}_{(\bfe_g,\bfX,y)\in\mathcal{D}}L(F_{\bfA^*}(\bfe_g,\bfX;\bfW^{(out)}, \bfV^{(out)}),y)\\
     \leq &(1+\epsilon_0)\mathrm{OPT}_{\bfA^*}+\epsilon_0,
\end{aligned}
 \end{equation}
 where $\bfA^*$  is the effective adjacency matrix in (\ref{eqn:concept2}). 
 \end{theorem}

Theorem \ref{thm: node_3} shows that the required sample complexity is  polynomial in $\|\bfA^*\|$ and $\delta$, where $\delta$ is the maximum node degree without self-connections in $\bfA$. Note that condition (\ref{eqn:pstar}) implies that   $\|\bfA^*\|_\infty$ is $O(1)$. Then as long as $\delta$ is   $O(N^\alpha)$ for some small $\alpha$ in $(0,1)$, say $\alpha=1/5$,  then 
one can accurately infer the unknown labels   from a small percentage of labeled nodes. Moreover, our sample complexity is sufficient but not necessary. It is possible to achieve a desirable generalization performance  if the number of labeled nodes is less than the bound in (\ref{eqn:complexity}).

Graph topology sampling affects the generalization performance through $\bfA^*$. From the discussion in Section \ref{sec:sampling}, graph sampling reduces the node correlation in $\bfA^*$, especially for dominating nodes. The generalization performance does not degrade when $\textrm{OPT}_{\bfA^*}$ is small, i.e., the resulting $\bfA^*$ is sufficient to characterize the node correlation in a given dataset. That explains the empirical success of graph sampling in many datasets.

\section{ Numerical Results} 
 
To unveil how our theoretical results are aligned with GCN's generalization performance in experiments, we will focus on numerical evaluations on synthetic data where we can control target functions and compare with $\bfA^*$ explicitly. We also evaluate both our graph sampling method and FastGCN \citep{CMX18} to validate that insights for our graph sampling method also apply to FastGCN. 

We generate a graph $\mathcal{G}$ with $N=2000$ nodes.  $\mathcal{G}$   has two degree groups. 
Group 1 has $N_1$ nodes, and every node degree approximately equals $d_1$. Group 2 has $N_2$ nodes, and every node degree approximately equals  $d_2$. The edges between nodes are randomly selected. 
 $\bfA$ is the normalized adjacency matrix of $\mathcal{G}$.

The node labels are  generated by the target function 
\begin{equation}\label{simu:y}
y=(\sin(\hat{\bfA}\bfX\bfW^*)\odot \tanh(\hat{\bfA}\bfX\bfW^*))\bfC^*,
\end{equation}
where $\hat{\bfA}\in\mathbb{R}^{N\times N}$, $\bfX \in\mathbb{R}^{N\times d}$, $\bfW^*\in\mathbb{R}^{d\times p}$ and $\bfC^*\in\mathbb{R}^{p\times K}$.  The  feature dimension $d=10$. $p=10$, and $K=2$. $\bfX$, $\bfW^*$ and $\bfC^*$ are all randomly generated with each entry i.i.d. from $\mathcal{N}(0,1)$. 

We consider a regression task with the $\ell_2$-regression loss function. A three-layer GCN as defined in (\ref{eqn:threelayer}) with $m$ neurons in each hidden layer is trained on  a  randomly selected set $\Omega$ of labeled nodes. The rest $N-|\Omega|$  labels are used for testing.  The learning rate   $\eta = 10^{-3}$. The mini-batch size is  $5$, and the dropout rate as $0.4$. 
The total number of iterations is $TT_w=4|\Omega|$.  Our graph topology sampling method samples $S_1=0.9 N_1$  and $S_2=0.9 N_2$ nodes for both groups in each iteration.

\subsection{Sample Complexity and Neural Network Width with respect to $\|\bfA^*\|_\infty$}


We fix $N_1=100$, $N_2=1900$ and  vary $\bfA$ by changing   node degrees   $d_1$ and $d_2$.   In the graph topology sampling method,  
  $p^*_1=0.7$ and  $p^*_2=0.3$. For every fixed $\bfA$, the effective adjacency matrix $\bfA^*$ is computed based on (\ref{eqn:Astar}) 
  using  $p^*_1$ and $p^*_2$.  Synthetic labels are generated based on (\ref{simu:y}) using  $\bfA^*$ as $\hat{\bfA}$. 
  
\begin{figure}[htpb]
\centering
 \centering
\includegraphics[width=0.8\linewidth]{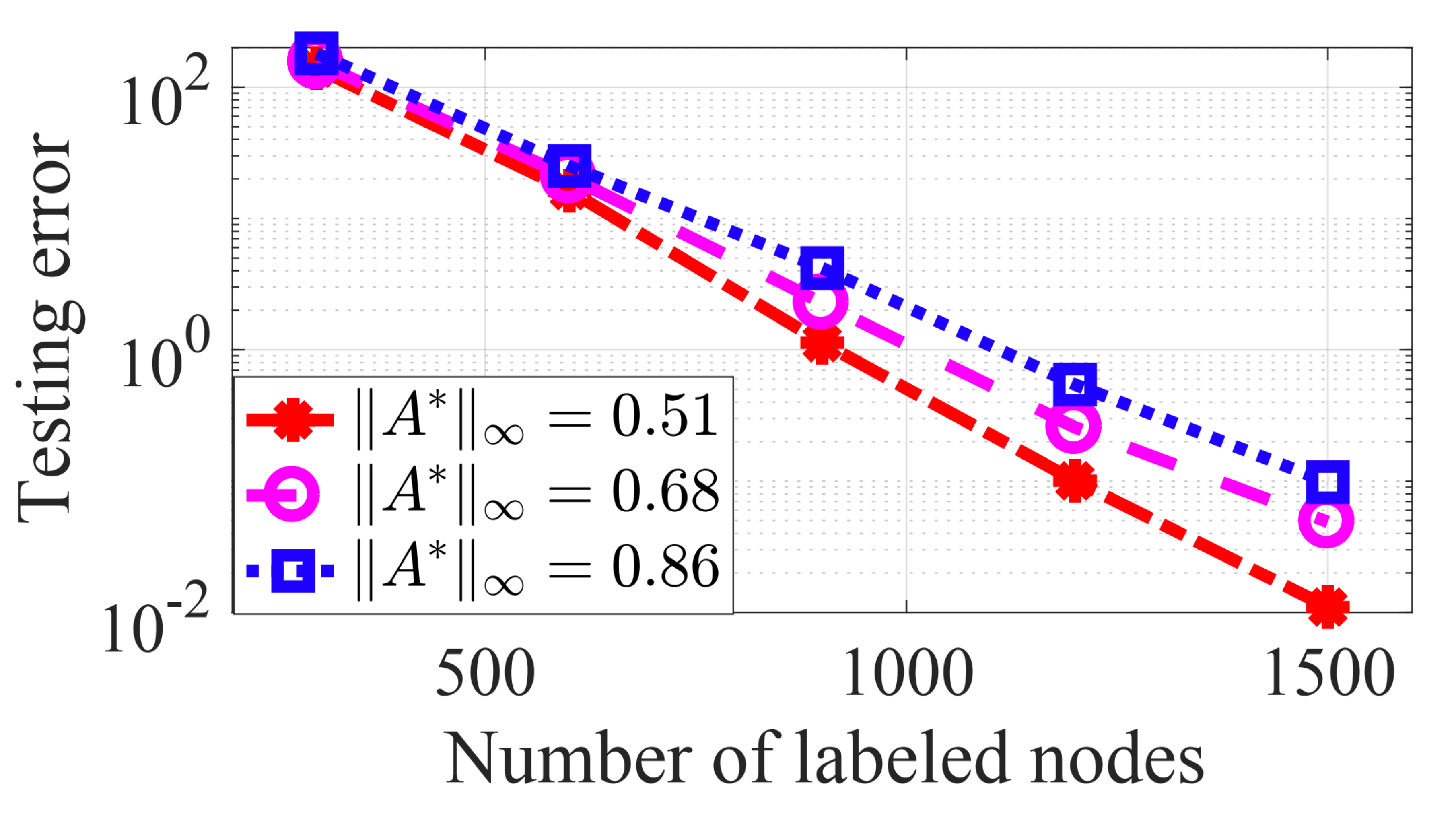}
\caption{The testing error when $|\Omega|$ and  $\|\bfA^*\|_\infty$ change. $m=500$} 
\label{fig: A_N}
\end{figure}

Figure \ref{fig: A_N} shows the testing error decreases as the number of labeled nodes  $|\Omega|$ increases, when the number of neurons per layer $m$ is fixed as $500$. Moreover, as $\|\bfA^*\|_\infty$ increases, the required number of labeled nodes increases to achieve the same level of testing error. This verifies our sample complexity bound in (\ref{eqn:complexity}). 

\begin{figure}[htpb]
\centering
 \centering
\includegraphics[width=0.8\linewidth]{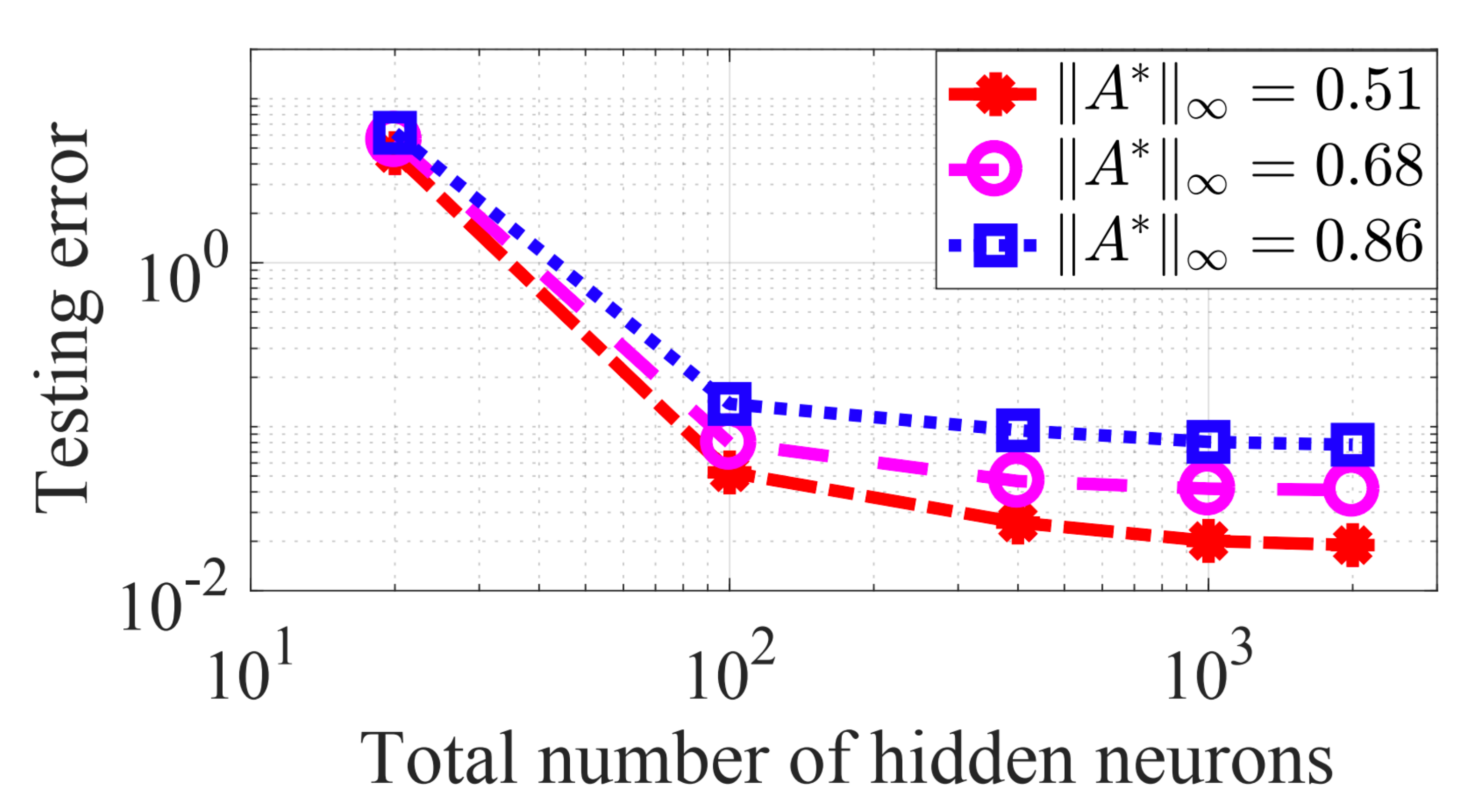}
\caption{The testing error when $m$ and $\|\bfA^*\|_\infty$ change. $|\Omega|=1500$.}
\label{fig: A_m}
\end{figure}

Figure \ref{fig: A_m} shows the testing error decreases as  $m$ increases when $|\Omega|$ is fixed as $1500$.  Moreover,   as $\|\bfA^*\|_\infty$ increases, a larger $m$ is needed to  achieve the same level of testing error. This verifies our bound on the number of neurons in (\ref{eqn:neuron}).

\subsection{Graph Sampling Affects   $\bfA^*$}\label{sec:varyA}

Here we fix $\bfA$ and the graph sampling strategy, and evaluate the prediction performance on datasets generated by (\ref{simu:y}) using different $\hat{\bfA}$.  We generate
  $\hat{\bfA}$ from  $\hat{\bfA}=\bfA \hat{\bfP}$, where $\hat{\bfP}$ is  a diagonal matrix with $\hat{ \bfP}_{ii}=\hat{p}_1$ for nodes $i$ in group $1$ and $\hat{ \bfP}_{ii}=\hat{p}_2$ for nodes $i$ in group $2$.   We vary  $\hat{p}_1$  and $\hat{p}_2$ to generate three different datasets from (\ref{simu:y}).  We consider both our graph sampling method in Section \ref{sec:sampling} and FastGCN \citep{CMX18}. 

\begin{figure}[h]
    \centering
    \subfigure[]{
        \begin{minipage}{0.2\textwidth}
        \centering
        \includegraphics[width=1\textwidth]{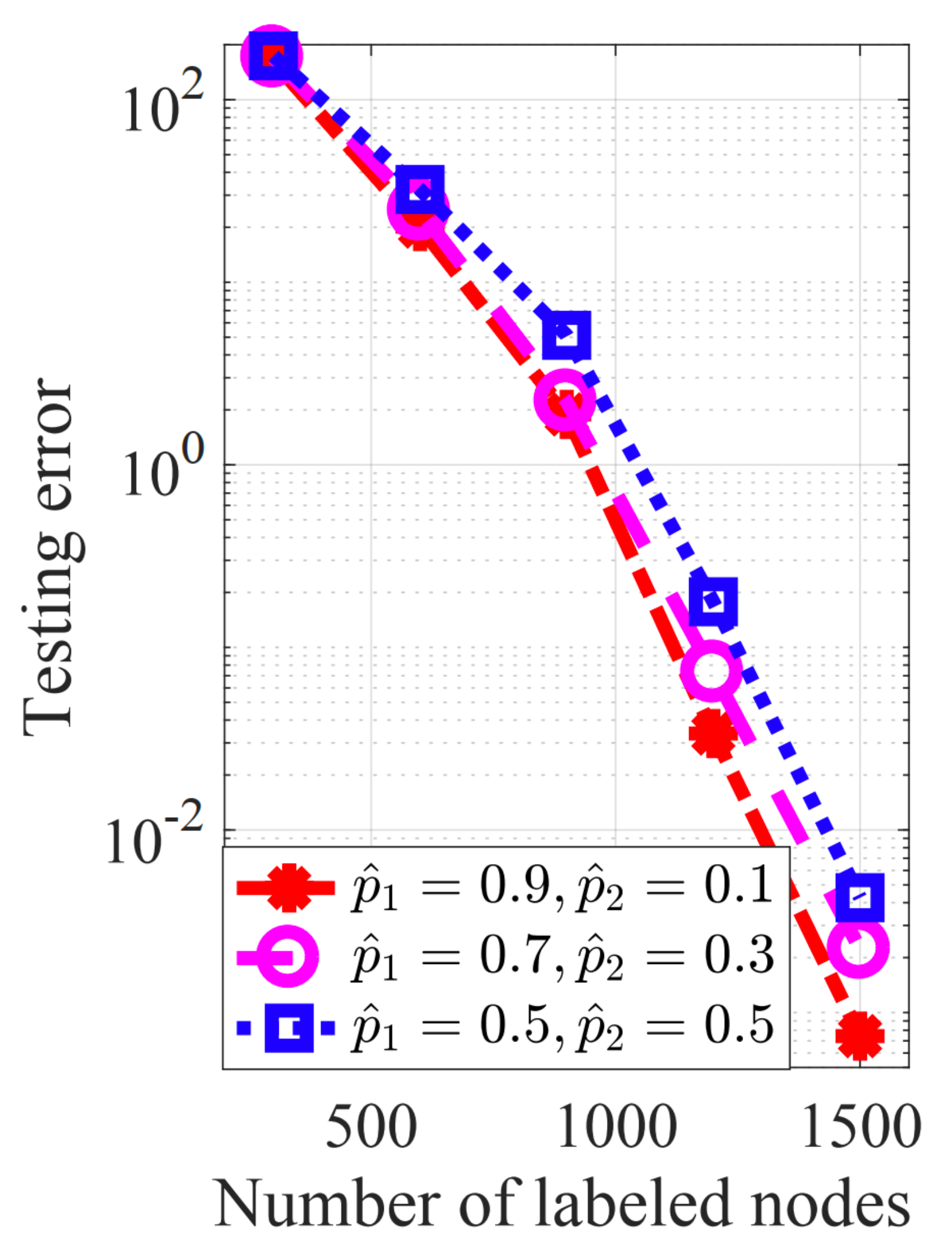}
        \end{minipage}
    }
    ~
    \subfigure[]{
        \begin{minipage}{0.22\textwidth}
        \centering
        \includegraphics[width=1\textwidth]{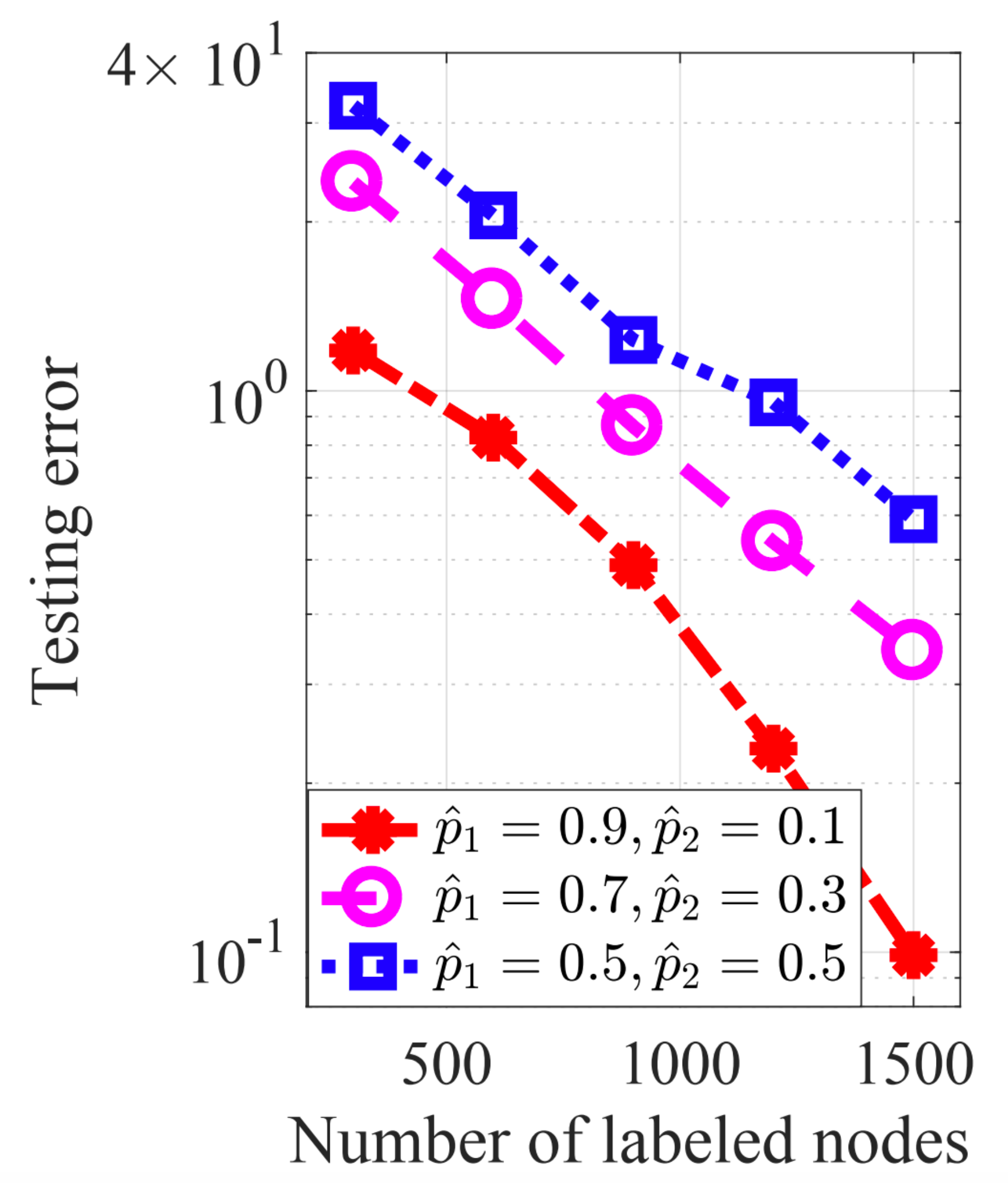}
        \end{minipage}
    }
    \caption{Generalization performance of learned GCNs on datasets generated from different $\hat{\bfA}$  by (a) our graph sampling strategy and   (b) FastGCN. $\bfA$ is very unbalanced.}
    \label{figure:unbalanced}
\end{figure}

In Figure~\ref{figure:unbalanced}, $N_1=100$ and $N_2=1900$. $d_1=10$ and $d_2=1$. 
Figure~\ref{figure:unbalanced}(a) shows the testing performance of a learned GCN by Algorithm 1, where $p_1^*=0.9$ and $p_2^*=0.1$. 
the method indeed performs the best on  Dataset 1 when $\hat{\bfA}$ is generated using $\hat{p}_1=0.9$ and $\hat{p}_2=0.1$, in which case $\bfA^*=\hat{\bfA}$. This verifies our theoretical result that graph sampling affects $\bfA^*$ in the target functions, i.e., it achieves the best performance if $\bfA^*$ is the same as $\hat{\bfA}$ in the target function. 

Fig.~\ref{figure:unbalanced} (b) shows the performance on the same three datasets where in each iteration of Algorithm 1, the graph sampling strategy is replaced with FastGCN \citep{CMX18}. The method also performs the best in Dataset 1 when $\bfA^*$ is generated using $\hat{p}_1=0.9$ and $\hat{p}_2=0.1$. 
The reason is that the graph topology is highly unbalanced  in the sense that $\sqrt{d_2}N_2 \gg \sqrt{d_1}N_1$, which means group $2$ has a much higher impact on other nodes in group $1$ in $\bfA$. The graph sampling reduces the impact of group $2$ nodes more significantly than group $1$ nodes, as discussed in Section \ref{sec:sampling}. 

\begin{figure}[h]
    \centering
    \subfigure[]{
        \begin{minipage}{0.2\textwidth}
        \centering
        \includegraphics[width=1\textwidth]{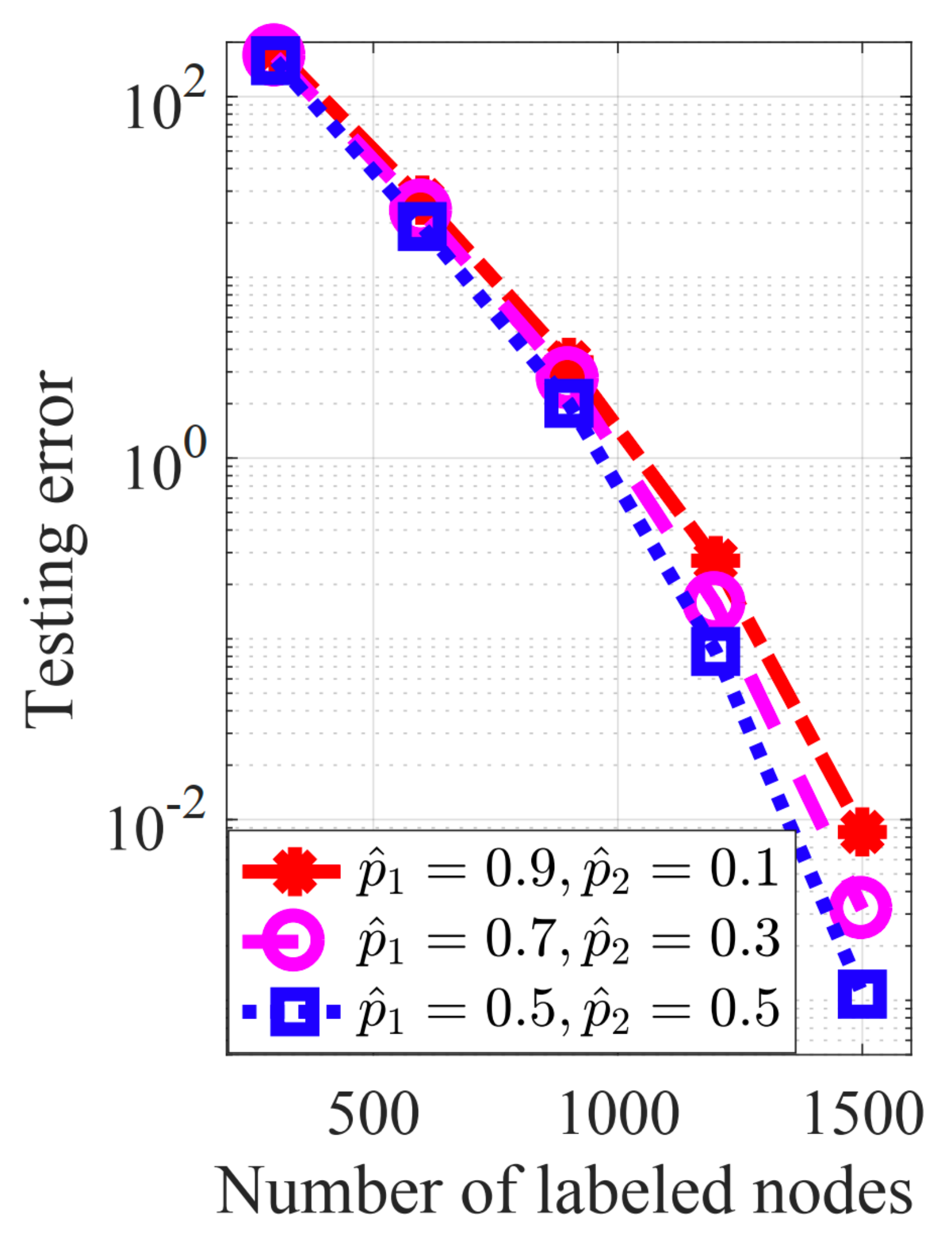}
        \end{minipage}
    }
    ~
    \subfigure[]{
        \begin{minipage}{0.22\textwidth}
        \centering
        \includegraphics[width=1\textwidth]{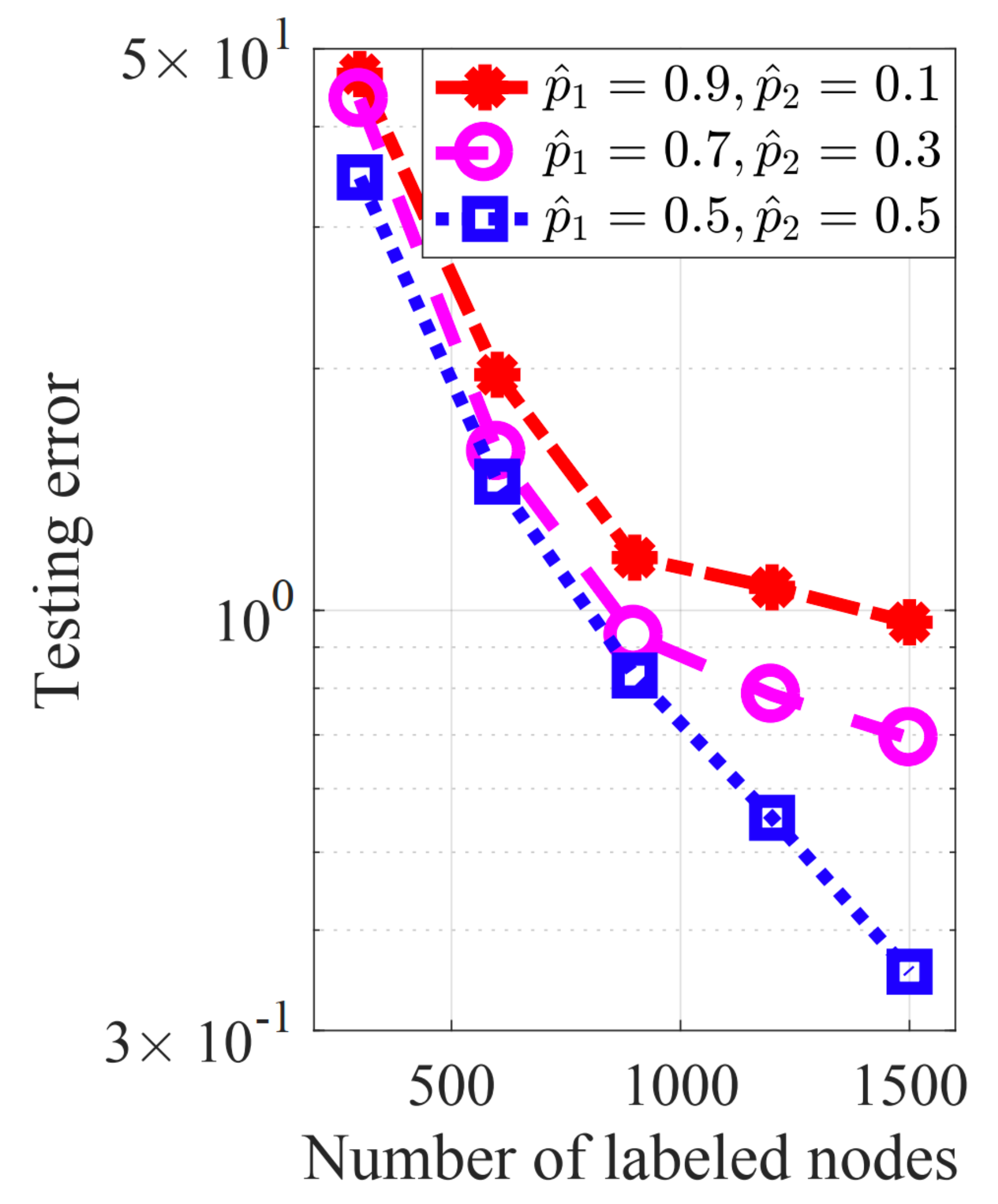}
        \end{minipage}
    }
    \caption{Generalization performance  of learned GCNs on datasets generated from different $\bfA^*$   by (a) our graph sampling strategy and (b) FastGCN. $\bfA$ is balanced.  }
    \label{figure:balanced}
\end{figure}

To further illustrate this, in Figure~\ref{figure:balanced}  we change the graph topology by setting $N_1=1000$ and $N_2=1000$, and all the other settings remain the same. 
In this case, the graph is  balanced  
because  $\sqrt{d_2}N_2$ and  $\sqrt{d_1}N_1$ are in the same order. We generate  different datasets using  the new $\bfA$ following the same method and evaluate the performance of both our graph sampling method and FastGCN. 
Both methods perform the best in Dataset 3 when $\hat{\bfA}$ is generated using $\hat{p}_1=0.5$ and $\hat{p}_2=0.5$. That is because on a balanced graph,  graph sampling reduces the impact of both groups equally.  

\section{Conclusion}

This paper  provides a new theoretical framework for explaining the empirical success of graph sampling in training   GCNs.  It quantifies the impact of graph sampling explicitly through the effective adjacency matrix and provides generalization and sample complexity analyses. One future direction is to develop active graph sampling strategies based on the presented insights and analyze its generalization performance. Other potential extension includes the construction of statistical-model-based characterization of $\bfA^*$ and fitness to real-world data, and the generalization analysis of deep
GCNs, graph auto-encoders, and jumping knowledge networks.  

\section*{Acknowledgements}



This work was supported by AFOSR FA9550-20-1-0122, ARO W911NF-21-1-0255, NSF 1932196 and the Rensselaer-IBM AI Research Collaboration (http://airc.rpi.edu), part of the IBM AI Horizons Network (http://ibm.biz/AIHorizons). We thank Ruisi Jian, Haolin Xiong at Rensselaer Polytechnic Institute for the help in formulating numerical experiments. We thank all anonymous reviewers for their
constructive comments.

\nocite{langley00}


\newpage
\appendix
\onecolumn

\section{Preliminaries}

\begin{lemma}
$\|\tilde{\bfa}_n\bfX\|\leq \|\bfA\|_\infty$.
\end{lemma}
Proof:\\
\begin{equation}
    \begin{aligned}
        \|\tilde{\bfa}_n\bfX\|&=\|\sum_{k=1}^N a_{n,k}\Tilde{\bfx}_k\|\\
        &= \|\sum_{k=1}^N \frac{a_{n,k}}{\sum_{k=1}^N a_{n,k}}\Tilde{\bfx}_k\|\cdot \sum_{k=1}^N a_{n,k}\\
        &\leq \sum_{k=1}^N \frac{a_{n,k}}{\sum_{k=1}^N a_{n,k}}\|\tilde{\bfx}_k\|\cdot \|\bfA\|_\infty\\
        &= \|\bfA\|_\infty
    \end{aligned}
\end{equation}
where the second to last step is by the convexity of $\|\cdot\|$.

\begin{lemma}\label{lm: sampling}
Given a graph $\mathcal{G}$ with $L(\geq1)$ groups of nodes, where the group $i$ with node degree $d_i$ is denoted as $\mathcal{N}_i$. Suppose that in iteration $t$, $\bfA^t$ (or any of $\bfA^{t(1)}$, $\bfA^{t(2)}$, $\bfA^{t(3)}$ in the general setting) is generated from the sampling strategy in Section \ref{sec:sampling}, if the number of sampled nodes satisfies $l_i\geq |\mathcal{N}_i|/(1+\frac{c_1\text{poly}(\epsilon)}{L p^*_i\Psi_i})$, we have
\begin{equation}
    \|\bfA^t-\bfA^*\|_\infty\leq \text{poly}(\epsilon)
\end{equation}
\end{lemma}

\textbf{Proof:}\\
From Section \ref{sec:sampling}, we can rewrite that
\begin{equation}
    \tilde{\bfa}_n^{t}=\begin{cases}\frac{|\mathcal{N}_k|}{l_k}p^*_k A_{n,j}, &\text{if the nodes }n, j\text{ are connected and }j\text{ is selected and }j\in\mathcal{N}_k\\
    0, &\text{else}\end{cases}
\end{equation}
\begin{equation}
    \tilde{\bfa^*}_n=\begin{cases}p^*_k A_{n,j}, &\text{if the nodes }n, j\text{ are connected and }j\in\mathcal{N}_k\\
    0, &\text{else}\end{cases}
\end{equation}
Let $\bfA^*=(\tilde{\bfa^*}_1^\top,\tilde{\bfa^*}_2^\top,\cdots,\tilde{\bfa^*}_{n})^\top$. Since that we need that $\sum_{j=1}^N A_{n,j}^*\leq O(1)$, we require 
\begin{equation}
    p^*_i\sum_{j\in\mathcal{N}_i}A_{n,j}\leq O(1/L), \text{ holds for any }i\in[L], n\in[N]\\
\label{p_i''constraints_ini}
\end{equation}
We first roughly compute the ratio of edges that one node is connected to the nodes in another group. For the node with degree $\text{deg}(i)$, it has $\text{deg}(i)-1$ open edges except the self-connection. Hence, the group with degree $\text{deg}(j)$ has $(\text{deg}(j)-1)|\mathcal{N}_j|$ open edges except self-connections in total. Therefore, the ratio of the edges connected to the group $j$ to all groups is
\begin{equation}
    \frac{(\text{deg}(j)-1)|\mathcal{N}_j|}{\sum_{l=1}^L (\text{deg}(l)-1)|\mathcal{N}_l|}\approx \frac{d_j|\mathcal{N}_j|}{\sum_{l=1}^L d_l|\mathcal{N}_l|}\label{p_i''constraints_ini}
\end{equation}
Define
\begin{equation}
    \Psi(n,i)=\sqrt{\frac{d_n}{d_{i}}}\cdot \frac{d_i|\mathcal{N}_i|}{\sum_{l=1}^L d_l|\mathcal{N}_l|}
\end{equation}
Then, as long as 
\begin{equation}
    p^*_i\sum_{j\in|\mathcal{N}_i|}A_{n,j}\approx p^*_i\frac{1}{\sqrt{d_{i} d_{n}}}\cdot \frac{d_i|\mathcal{N}_i|}{\sum_{l=1}^L d_l|\mathcal{N}_l|} d_{n}\lesssim p^*_i\Psi(n,i)\leq O(1/L)\label{p_Phi_1}
\end{equation}
i.e., 
\begin{equation}
    p^*_i\leq \frac{c_1}{L\cdot \max_{n\in[L]}\{\Psi(n,i)\}}=\frac{c_1}{L\cdot \Psi(L,i)}=\frac{c_1}{L}\sqrt{\frac{d_i}{d_L}}\frac{\sum_{l=1}^L d_l |\mathcal{N}_l|}{d_i|\mathcal{N}_i|}\label{p_i''upper}
\end{equation}
for some constant $c_1>0$, we can obtain that $\|\bfA^*\|_\infty\leq O(1)$. 
Since that
\begin{equation}
    \sum_{j\in\mathcal{S}_k}A_{n,j}\approx \frac{1}{\sqrt{d_{i} d_{n}}}\cdot \frac{d_i|\mathcal{N}_i|}{\sum_{l=1}^L d_l|\mathcal{N}_l|} d_{n}\frac{l_k}{|\mathcal{N}_k|}\approx \sum_{j\in\mathcal{N}_k}A_{n,j}\frac{l_k}{|\mathcal{N}_k|}\label{in_sk}
\end{equation}
\begin{equation}
    \sum_{j\notin\mathcal{S}_k}A_{n,j}\approx \frac{1}{\sqrt{d_{i} d_{n}}}\cdot \frac{d_i|\mathcal{N}_i|}{\sum_{l=1}^L d_l|\mathcal{N}_l|} d_{n}(1-\frac{l_k}{|\mathcal{N}_k|})\approx \sum_{j\in\mathcal{N}_k}A_{n,j}(1-\frac{l_k}{|\mathcal{N}_k|}),\label{notin_sk}
\end{equation}
the difference between $\tilde{\bfa}_n^{t}$ and $\tilde{\bfa^*}_n$ can then be derived as
\begin{equation}
    \begin{aligned}
    &\|\tilde{\bfa}_n^{t}-\tilde{\bfa^*}_{n}\|_1\\
    =&\Big| \sum_{k=1}^L\sum_{j\in\mathcal{S}_k}A_{n,j} p^*_k(\frac{|\mathcal{N}_k|}{l_k}- 1)+\sum_{k=1}^L \sum_{j\notin\mathcal{S}_k}A_{n,j} p^*_k\Big|\\
    \lesssim &\sum_{k=1}^L(p^*_k(\frac{|\mathcal{N}_k|}{l_k}-1)\frac{l_k}{|\mathcal{N}_k|}\sum_{j\in\mathcal{N}_k} A_{n,j}+ (1-\frac{l_k}{|\mathcal{N}_k|})p^*_k\sum_{j\in\mathcal{N}_k} A_{n,j})\\
    \lesssim & \text{poly}(\epsilon)\sum_{k=1}^L \frac{1}{L\Psi(L,k)} \sum_{j\in \mathcal{N}_k}A_{n,j}\\
    :=&\text{poly}(\epsilon)\Gamma(\bfA^*)\label{a_n,t-a_n, p*}
    \end{aligned}
\end{equation}
where the first inequality is by (\ref{in_sk}, \ref{notin_sk}) and the second inequality holds as long as $l_i\geq |\mathcal{N}_i|/(1+\frac{c_1\text{poly}(\epsilon)}{L p^*_i\Psi(L,i)})$.
Combining (\ref{p_Phi_1}), we have
\begin{equation}
    \sum_{i=1}^L p^*_i\sum_{j\in\mathcal{N}_i}A_{n,j}\lesssim \sum_{i=1}^L \frac{1}{L \Psi(L,i)}\sum_{j\in\mathcal{N}_i}A_{n,j}=\Gamma(\bfA^*)\leq O(1)
\end{equation}
Hence, (\ref{a_n,t-a_n, p*}) can be bounded by $\text{poly}(\epsilon)$.\\

\subsection{Symmetric graph sampling method}\label{sec: symmetric_sampling}
We provide and discuss a symmetric graph sampling method in this section. The insights behind this version of sampling strategy is the same as in Section \ref{sec:sampling}.

Similar to the asymmetric construction in Section \ref{sec:sampling}, we consider a group-wise uniform sampling strategy, where $S_l$ nodes are sampled uniformly from $N_l$ nodes.  For all unsampled nodes, we set the corresponding diagonal entries of a  diagonal matrix $\bfP^s$  to be zero. 
If node $i$ is sampled in this iteration and  belongs to group $l$ for any $i$ and $l$, the $i$th diagonal entry of  $\bfP^s$ is set as $\sqrt{p^*_lN_l/S_l}$ for some non-negative constant $p^*_l$. Then  $\bfA^s=\bfP^s\bfA\bfP^s$.

Based on this symmetric graph sampling method, we define the effective adjacency matrix as 
\begin{equation}
    \bfA^*=\bfP^*\bfA\bfP^*,
\end{equation}
where $\bfP^*$ is a diagonal matrix defined as
\begin{equation}
    \bfP_{ii}^*=\sqrt{p_l^*}\ \ \ \ \text{if node }i\text{ belongs to degree group }l
\end{equation}

The scaling factor  $p^*_l$ should satisfy
\begin{equation}
0 \leq p^*_l  \leq \frac{c_2}{L^2\psi_l^2}, \quad \forall l
\end{equation}
for a positive constant $c_2$ that can be sufficiently large. 
  $\psi_l$ is defined in (\ref{eqn:psi}).
The number of sampled nodes shall satisfy 
\begin{equation}
 \frac{S_l}{N_l} \geq \ (1+ \frac{c_2 \text{poly}(\epsilon)}{L\sqrt{p^*_l} \psi_l})^{-2} \quad \quad \forall l\in[L]
\end{equation}
where $\epsilon$ is a small positive value. 

\begin{lemma}\label{lm: symmetric_sampling}
Given a graph $\mathcal{G}$ with $L(\geq1)$ groups of nodes, where the group $i$ with node degree $d_i$ is denoted as $\mathcal{N}_i$. Suppose $\bfA^{t}$ (or any of $\bfA^{t(1)}$, $\bfA^{t(2)}$, $\bfA^{t(3)}$ in the general setting) is generated from the sampling strategy in Section \ref{sec: symmetric_sampling}, if the number of sampled nodes satisfies $l_i\geq |\mathcal{N}_i|/(1+\frac{c_2\text{poly}(\epsilon)}{L p^*_i\Psi_i})$, then we have
\begin{equation}
    \|\bfA^{t}-\bfA^*\|_\infty\leq \text{poly}(\epsilon)
\end{equation}
\end{lemma}

\textbf{Proof:}\\

From Section \ref{sec: symmetric_sampling}, we can rewrite that
\begin{equation}
    \tilde{\bfa}_n^{t}=\begin{cases}\sqrt{\frac{|\mathcal{N}_k||\mathcal{N}_u|}{l_k l_u}p^*_k p^*_u} A_{n,j}, &\text{if the nodes }n, j\text{ are connected and }j\text{ is selected and }n\in\mathcal{N}_u,\ j\in\mathcal{N}_k\\
    0, &\text{else}\end{cases}
\end{equation}
\begin{equation}
    \tilde{\bfa^*}_n=\begin{cases}\sqrt{p^*_k p^*_u} A_{n,j}, &\text{if the nodes }n, j\text{ are connected and }n\in\mathcal{N}_u, j\in\mathcal{N}_k\\
    0,\ &\text{else}\end{cases}
\end{equation}
Then $\bfA^*=(\tilde{\bfa^*}_1^\top,\tilde{\bfa^*}_2^\top,\cdots,\tilde{\bfa^*}_{n})^\top$. Then, for $n\in\mathcal{N}_u$, as long as 
\begin{equation}
    \sum_{j\in|\mathcal{N}_i|}\sqrt{p_i^* p_u^*}A_{n,j}\approx \sqrt{p_u^*p^*_i}\frac{1}{\sqrt{d_{i} d_{n}}}\cdot \frac{d_i|\mathcal{N}_i|}{\sum_{l=1}^L d_l|\mathcal{N}_l|} d_{n}\lesssim \sqrt{p^*_u p^*_i}\Psi(n,i)\leq \sqrt{ p^*_i}\Psi(n,i)\leq O(1/L)\label{p_Phi_1}
\end{equation}
i.e., 
\begin{equation}
    \sqrt{p^*_i}\leq \frac{c_2}{L\cdot \max_{n\in[L]}\{\Psi(n,i)\}}=\frac{c_1}{L\cdot \Psi(L,i)}=\frac{c_2}{L}\sqrt{\frac{d_i}{d_L}}\frac{\sum_{l=1}^L d_l |\mathcal{N}_l|}{d_i|\mathcal{N}_i|}\label{p_i''upper}
\end{equation}
for some constant $c_2>0$, we can obtain that $\|\bfA^*\|_\infty\leq O(1)$. 

The difference between $\tilde{\bfa}_n^{t}$ and $\tilde{\bfa^*}_n$ can then be derived as
\begin{equation}
    \begin{aligned}
    &\|\tilde{\bfa}_n^{t}-\tilde{\bfa^*}_{n}\|_1\\
    =&\Big| \sum_{k=1}^L\sum_{j\in\mathcal{S}_k}A_{n,j} \sqrt{p^*_u p^*_k}(\sqrt{\frac{|\mathcal{N}_k||\mathcal{N}_u|}{l_k l_u}}- 1)+\sum_{k=1}^L \sum_{j\notin\mathcal{S}_k}A_{n,j} \sqrt{ p^*_u p^*_k}\Big|\\
    \approx &\Big| \sum_{k=1}^L\sum_{j\in\mathcal{N}_k}A_{n,j} \sqrt{p^*_u p^*_k}(\sqrt{\frac{|\mathcal{N}_k||\mathcal{N}_u|}{l_k l_u}}- 1)\frac{l_k}{|\mathcal{N}_k|}+\sum_{k=1}^L \sum_{j\in\mathcal{N}_k}A_{n,j} \sqrt{ p^*_u p^*_k}(1-\frac{l_k}{|\mathcal{N}_k|})\Big|\\
    \lesssim&\text{poly}(\epsilon)
    \end{aligned}
\end{equation}
as long as $l_i\geq |\mathcal{N}_i|/(1+\frac{c_2\text{poly}(\epsilon)}{L \sqrt{p^*_i}\Psi(L,i)})^2$.

\section{Node classification for three layers}

In the whole proof, we consider a more general target function compared to (\ref{eqn:concept2}). We write $F^*: \mathbb{R}^N\times  \mathbb{R}^{N\times d}\rightarrow \mathbb{R}^K$:
\begin{equation}\label{eqn:concept2_new}
\begin{aligned}
&F^*_{\bfA^*}=(f_1^*,f_2^*,\cdots, f_K^*), \\
&f_r^*(\bfe_g, \bfX)=\bfe_g^\top \sum_{k\in[p_1]}c_{k,r}^*\Phi\Big(\bfA^*\sum_{j\in[p_2]}v_{1,k,j}^*\phi_{1,j}(\bfA^*\bfX\bfw_{1,j}^*)\Big)\odot \Big(\bfA^*\sum_{l\in[p_2]}v_{2,k,l}^*\phi_{2,l}(\bfA^*\bfX\bfw_{2,l}^*)\Big),\ 
\forall r\in[K],
\end{aligned}
\end{equation}
where each $\phi_{1,j}$, $\phi_{2,j}$, $\Phi_i$: $\mathbb{R}\rightarrow\mathbb{R}$ is infinite-order smooth.\\
Table \ref{tab:notation} shows some important notations used in our theorem and algorithm. Table \ref{tab:table2} gives the full parameter choices for the three-layer GCN.  
$\text{ploy}(\log(m_1 m_2))$ in the following analysis.

\begin{table}[h!]
 
  \begin{center}
        \caption{Summary of notations}
        \label{tab:notation}
\begin{tabularx}{\textwidth}{lX} 
 \hline

 $\mathcal{G}=\{\mathcal{V},\mathcal{E}\}$ & $\mathcal{G}$ is an un-directed graph consisting of a set of nodes $\mathcal{V}$ and a set of edges $\mathcal{E}$.\\ 
 \hline
 $N$ & The total number of nodes in a graph.\\
 \hline
  $\bfA=\bfD^{-\frac{1}{2}}\tilde{\bfA}\bfD^{-\frac{1}{2}}$ & {$\bfA\in\mathbb{R}^{N\times N}$ is the normalized adjacency matrix computed by the degree matrix $\bfD$ and the \newline initial adjacency matrix $\tilde{\bfA}$}.\\
 \hline
  $\bfA^*$ & The effective adjacency matrix.\\
 \hline
 $\bfA^t$ & The sampled adjacency matrix using our sampling strategy in Section \ref{sec:sampling} at the $t$-th iteration.\\
 \hline
 $\bfe_g$, $\bfX$ $y_n$ &  $\bfe_g$ belongs to $\{\bfe_i\}_{i=1}^N$ and selects the index of the node label. $\bfX\in\mathbb{R}^{N\times d}$ is the feature matrix. $y_n$ is the label of the $n$-th node.\\
 \hline
  $m_1$, $m_2$ & $m_1$, $m_2$ are the number of neurons in the first and second hidden layer, respectively.\\
 \hline
 $\bfW$, $\bfV$, $\bfB_1$, $\bfB_2$ & $\bfX\in\mathbb{R}^{N\times d}$ is the data matrix. $\bfW$, $\bfV$ are the weight matrices of the first and second hidden layer, respectively. $\bfB_1$, $\bfB_2$ are the corresponding bias matrices.\\
 \hline
 $\bfW^{(0)}$, $\bfV^{(0)}$ & $\bfW^{(0)}$ and $\bfV^{(0)}$ are random initializations of $\bfW$ and $\bfV$, respectively. \\
 \hline
 $\bfW^\rho$, $\bfV^\rho$ &  $\bfW^\rho$ and $\bfV^\rho$ are two random matrices used for Gaussian smoothing. \\
 \hline
 $\boldsymbol{\Sigma}$ & The Dropout technique.\\
 \hline
$\Omega$, $\Omega^t$ &  $\Omega$ is the set of labeled nodes and $\Omega_t$ is the batch of labeled nodes at the $t$-th iteration.\\
 \hline
 $T$, $T_w$, $\eta$, $\lambda_t$ & In Algorithm \ref{alg: 1}, $T$ is the number of outer iterations for the weight decay step, while $T_w$ is the number of inner iterations for the SGD steps. $\eta$ is the step size and $\lambda_t$ is the weight decay coefficient at the $t$-th iteration.\\
 \hline

 $L$, $d_l$, $S_l$, $N_l$ & $L$ is the number of node groups in a graph. $d_l$ is the order-wise degree in the $l$-th group. $N_l$ is the number of nodes in group $l$.\\
 \hline
$S_l$ & The number of nodes we sample in group $l$.\\
  \hline

\end{tabularx}
\end{center}

\end{table}

\begin{table}[h!]
  \begin{center}
        \caption{Full parameter choices for three-layer GCN}   \label{tab:table2}
    \begin{tabular}{l|c} 
 \hline
 $\tau_v'$ & $m_1^{1/2-0.005}/(\sqrt{\epsilon_0}m_2^{1/2})$ \\ 
 \hline
 $\tau_w'$ & $C_0/(\epsilon_0^{1/4}m_1^{3/4-0.005})$\\
 \hline
 $\tau_v$ & $m_1^{1/2-0.001}/m_2^{1/2}\gg \tau_v'$\\
 \hline
 $\tau_w$ & $1/m_1^{3/4-0.01}\gg\tau_w'$ \\
 \hline
 $\lambda_v$ & $2/(\tau_v')^2$\\
 \hline
 $\lambda_w$ & $2/(\tau_w')^4$\\
 \hline
 $\sigma_v$ & $1/m_2^{1/2+0.01}$\\
 \hline
 $\sigma_w$ & $\sigma_w=1/m_1^{1-0.01}$\\
 \hline
 $C$ & $\mathcal{C}_\epsilon(\phi,\|\bfA\|_\infty)\sqrt{\|\bfA\|_\infty^2+1}$\\
 \hline
  $C'$ & $10C\sqrt{p_2}$\\
  \hline 
  $C''$ & $\mathcal{C}_\epsilon(\Phi,C')\sqrt{\|\bfA\|_\infty^2+1}$\\
  \hline
  $C_0$ & $\tilde{O}(p_1^2p_2K^2 CC'')$\\
  \hline
  $\epsilon_c$ & $1$\\
  \hline
    \end{tabular}

  \end{center}
\end{table}

\subsection{Lemmas}

\subsubsection{Function approximation}
To show that the target function can be learnt by the learner network with the Relu function, a good approach is to firstly find a function $h(\cdot)$ such that the $\phi$ functions in the target function can be approximated by $h(\cdot)$ with an indicator function. In this section, Lemma \ref{lm: h_function} provides the existence of such $h(\cdot)$ function. Lemma \ref{lm: claim1_h} and \ref{lm: claim2_h} are two supporting lemmas to prove Lemma \ref{lm: h_function}.
\begin{lemma}\label{lm: h_function}
For every smooth function $\phi$, every $\epsilon\in(0,\frac{1}{\mathcal{C}(\phi,a)\sqrt{a^2+1}})$, there exists a function $h: \mathbb{R}^2\rightarrow [-\mathcal{C}_\epsilon(\phi,a)\sqrt{a^2+1}, \mathcal{C}_\epsilon(\phi,a)\sqrt{a^2+1}]$ that is also $\mathcal{C}_\epsilon(\phi,a)\sqrt{a^2+1}$-Lipschitz continuous on its first coordinate with the following two (equivalent) properties:\\
(a) For every $x_1\in[-a,a]$ where $a>0$:
$$\Big|\mathbb{E}\Big[\mathbb{1}_{\alpha_1x_1+\beta_1\sqrt{a^2-x_1^2}+b_0\geq0}h(\alpha_1,b_0)\Big]-\phi(x_1)\Big|\leq\epsilon$$
where $\alpha_1,\beta_1,b_0\sim\mathcal{N}(0,1)$ are independent random variables.\\
(b) For every $\bfw^*, \bfx\in\mathbb{R}^d$ with $\|\bfw^*\|_2=1$ and $\|\bfx\|\leq a$:
$$\Big|\mathbb{E}\Big[\mathbb{1}_{\bfw\bfX+b_0\geq0}h(\bfw^\top\bfw^*,b_0)\Big]-\phi({\bfw^*}^\top\bfx)\Big|\leq\epsilon$$
where $\bfw\sim\mathcal{N}(0,\bfI)$ is an d-dimensional Gaussian, $b_0\sim\mathcal{N}(0,1)$.\\
Furthermore, we have $\mathbb{E}_{\alpha_1,b_0\sim\mathcal{N}(0,1)}[h(\alpha_1,b_0)^2]\leq (\mathcal{C}_s(\phi,a))^2(a^2+1)$.\\
(c) For every $\bfw^*, \bfx\in\mathbb{R}^d$ with $\|\bfw^*\|_2=1$, let $\tilde{\bfw}=(\bfw,b_0)\in\mathbb{R}^{d+1}$, $\tilde{\bfx}=(\bfx,1)\in\mathbb{R}^{d+1}$ with $\|\tilde{\bfx}\|\leq \sqrt{a^2+1}$, then we have
$$\Big|\mathbb{E}\Big[\mathbb{1}_{\tilde{\bfw}^\top\tilde{\bfx}\geq0}h({\tilde{\bfw}[1:d]}^\top\bfw^*,\tilde{\bfw}[d+1])\Big]-\phi({\bfw^*}^\top\tilde{\bfx}[1:d])\Big|\leq\epsilon$$
where $\tilde{\bfw}\sim\mathcal{N}(0,\bfI_{d+1})$ is an d-dimensional Gaussian.\\
We also have $\mathbb{E}_{\tilde{\bfw}\in\mathcal{N}(0,\bfI_{d+1})}[h({\tilde{\bfw}[1:d]}^\top\bfw^*,\tilde{\bfw}[d+1])^2]\leq (\mathcal{C}_s(\phi,a))^2(a^2+1)$.\\
\end{lemma}
\textbf{Proof:}\\
Firstly, since we can assume $\bfw^*=(1,0,\cdots,0)$ without loss of generality by rotating $\bfx$ and $\bfw$, it can be derived that $\bfx$, $\bfw$, $\bfw^*$ are equivalent to that they are two-dimensional. Therefore, proving Lemma \ref{lm: h_function}b suffices in showing Lemma \ref{lm: h_function}a.\\
Let $\bfw_0=(\alpha,\beta)$, $\bfx=(x_1, \sqrt{t^2-x_1^2})$ where $\alpha$ and $\beta$ are independent. Following the idea of Lemma 6.3 in \citep{ALL19}, we use another randomness as an alternative, i.e., we write $\bfx^\perp=(\sqrt{t^2-x_1^2},-x_1)$, $\bfw_0=\alpha\frac{\bfx}{t}+\beta\frac{\bfx^\perp}{t}\sim\mathcal{N}(0,\bfI)$. Then $\bfw_0\bfX=t\alpha$. Let $\alpha_1=w_{01}=\alpha \frac{x_1}{t}+\beta\sqrt{1-\frac{x_1^2}{t^2}}$, where $\alpha,\beta\sim\mathcal{N}(0,1)$. Hence, $\alpha_1\sim\mathcal{N}(0,1)$.\\
We first use Lemma \ref{lm: claim1_h} to fit $\phi(x_1)$.
By Taylor expansion, we have
\begin{equation}
\begin{aligned}
\phi(x_1)
&=c_0+\sum_{i=1, \text{\ odd\ } i}^\infty c_i x_1^i+\sum_{i=2, \text{\ even\ }i}^\infty c_i x_1^i\\
&=c_0+\sum_{i=1}^\infty c_i'\mathbb{E}_{\alpha,\beta\sim\mathcal{N}(0,1)}[h_i(\alpha_1)\mathbb{1}[q_i(b_0)]\mathbb{1}[\bfw_0\bfX+b_0\geq0]]
\end{aligned}
\end{equation}
where $h_i(\cdot)$ is the Hermite polynomial defined in Definition A.5 in \citep{ALL19}, and 
\begin{equation}
    c_i'=\frac{c_i}{p_i'},\ |c_i'|\leq \frac{200i^2|c_i|}{(i-1)!!}\frac{\sqrt{t^2+1}}{t^{1-i}} \text{ and } q_i(b_0)=\begin{cases}|b_0|\leq t/(2i), & i\mbox{ is odd} \\ 0<-b_0\leq t/(2i), & i\mbox{ is even}\end{cases}\label{ci_qi}
\end{equation}
Let $B_i=100i^\frac{1}{2}+10\sqrt{\log(\frac{1}{\epsilon}\frac{\sqrt{t^2+1}}{t^{1-i}})}$. Define $\hat{h}_i(\alpha_1)=h_i(\alpha_1)\cdot \mathbb{1}[|\alpha_1|\leq B_i]+h_i(\text{sign}(\alpha_1)B_i)\cdot\mathbb{1}[|\alpha_1|> B_i]$ as the truncated version of the Hermite polynomial. Then we have
$$\phi(x_1)=c_0+R(x_1)+\sum_{i=1}^\infty c_i'\mathbb{E}_{\alpha,\beta\sim\mathcal{N}(0,1)}[\hat{h}_i(\alpha_1)\mathbb{1}[q_i(b_0)]\mathbb{1}[\bfw_0\bfX+b_0\geq0]],$$
where 
$$R(x_1)=\sum_{i=1}^\infty c_i'\mathbb{E}_{\alpha,\beta\sim\mathcal{N}(0,1)}\Big[\big(h_i(\alpha_1)\cdot\mathbb{1}[|\alpha_1|>B_i]-h_i(\text{sign}(\alpha_1)B_i\cdot\mathbb{1}[|\alpha|> B_i])\big)\mathbb{1}[q_i(b_0)]\mathbb{1}[\bfw_0\bfX+b_0\geq0]\Big]$$
Define
$$h(\alpha_1,b_0)=c_0+\sum_{i=1}^\infty c_i'\cdot \hat{h}_i(\alpha_1)\cdot\mathbb{1}[q_i(b_0)]$$
Then by Lemma \ref{lm: claim2_h}, we have
$$|\mathbb{E}_{\alpha,\beta,b_0\sim\mathcal{N}(0,1)}[\mathbb{1}[\bfw_0\bfX+b_0\geq0]\cdot h(\alpha_1,b_0)-\phi(x_1)|\leq|R(x_1)|\leq\frac{\epsilon}{4}$$
We also have
\begin{equation}
    \begin{aligned}
    \mathbb{E}_{\alpha_1,b_0\sim\mathcal{N}(0,1)}[h(\alpha_1,b_0)^2]&\leq (\epsilon^2+c_0^2)+O(1)\cdot\sum_{i=1}^\infty\frac{i!\cdot |c_i|^2 i^3}{((i-1)!!)^2}\cdot(\frac{\sqrt{t^2+1}}{t^{1-i}})^2\\
    &\leq (\epsilon^2+c_0^2)+\sum_{i=1}^\infty i^{3.5}\cdot|c_i|^2\cdot(\frac{\sqrt{t^2+1}}{t^{1-i}})^2\\
    &\leq (\epsilon^2+c_0^2)+\Big(\sum_{i=0}^\infty (i+1)^{1.75}\cdot|c_i|\cdot t^i\sqrt{t^2+1}\Big)^2\\
    &\leq \mathcal{C}_s(\phi,t)^2(t^2+1)
    \end{aligned}
\end{equation}

\begin{lemma}\label{lm: claim1_h}
Denote $h_i(x)$ as the degree-i Hermite polynomial as in Definition A.5 in \citep{ALL19}. For every integer $i\geq1$, there exists constant $p_i'$ with $|p_i'|\geq \frac{t^{1-i}}{\sqrt{t^2+1}}\frac{(i-1)!!}{100i^2}$ such that
\begin{equation}
    \text{for even }i:\ \ x_1^i=\frac{1}{p_i'}\mathbb{E}_{\bfw_0\sim\mathcal{N}(0,\bfI), b_0\sim\mathcal{N}(0,1)}[h_i(\alpha_1)\mathbb{1}[\alpha\geq -\frac{b_0}{t}]\mathbb{1}[0<-b_0\leq \frac{t}{2i}]]
\end{equation}
\begin{equation}
    \text{for odd }i:\ \ x_1^i=\frac{1}{p_i'}\mathbb{E}_{\bfw_0\sim\mathcal{N}(0,\bfI), b_0\sim\mathcal{N}(0,1)}[h(\alpha_1)\mathbb{1}[\alpha\geq -\frac{b_0}{t}]\mathbb{1}[|b_0|\leq \frac{t}{2i}]]
\end{equation}
for $\|\bfx\|\leq t$.
\end{lemma}
\textbf{Proof:}\\
For even $i$, by Lemma A.6 in \citep{ALL19}, we have
$$\mathbb{E}_{\bfw_0\sim\mathcal{N}(0,\bfI), b_0\sim\mathcal{N}(0,1)}[h_i(\alpha_1)\mathbb{1}[\alpha\geq -\frac{b_0}{t}]\mathbb{1}[0<-b_0\leq \frac{t}{2i}]]=\mathbb{E}_{b_0\sim\mathcal{N}(0,1)}[p_i\cdot \mathbb{1}[0<-b_0\leq \frac{t}{2i}]]\cdot \frac{x_1^i}{t^i}$$, 
where $$p_i=(i-1)!!\frac{\exp(-b_0^2/(2t^2))}{\sqrt{2\pi}}\sum_{r=1, r\text{ odd}}^{i-1}\frac{(-1)^\frac{i-1-r}{2}}{r!!}\binom{i/2-1}{(r-1)/2}(-b_0/t)^r$$
Define $c_r=\frac{(-1)^\frac{i-1-r}{2}}{r!!}\binom{i/2-1}{(r-1)/2}$. Then $\text{sign}(c_r)=-\text{sign}(c_{r+2})$. We can derive
$$\Big|\frac{c_{r}(-b_0/t)^{r}}{c_{r-2}(-b_0/t)^{r-2}}\Big|=\Big|(\frac{b_0}{t})^2\frac{i+1-r}{r(r-1)}\Big|\leq \frac{1}{4i}\leq\frac{1}{4}$$
Therefore, 
$$\Big|\sum_{r=1, r\text{ odd}}^{i-1}c_r(-b_0/t)^r\Big|\geq \frac{3}{4}|b_0/t|$$
\begin{equation}
\begin{aligned}
    &|\mathbb{E}_{b_0\sim\mathcal{N}(0,1)}[p_i\cdot\mathbb{1}[0\leq -b_0/t\leq 1/(2i)]]|\cdot t^{-i}\\
    \geq &|\mathbb{E}_{b_0\sim\mathcal{N}(0,1)}[(i-1)!!\frac{\exp(-b_0^2/2t^2)}{\sqrt{2\pi}}\cdot\frac{3}{4}|b_0/t|\cdot\mathbb{1}[0\leq -b_0/t\leq 1/(2i)]]|\cdot t^{-i}\\
    =& t^{-i}\cdot\int_{-\frac{t}{2i}}^0(i-1)!!\frac{\exp(-\frac{b_0^2}{2}(1+\frac{1}{t^2}))}{2\pi}\cdot\frac{3}{4}(-\frac{b_0}{t})db_0\\
    =& t^{-i}\cdot\frac{t}{t^2+1}\exp(-\frac{b_0^2}{2}(1+\frac{1}{t^2}))(i-1)!!\frac{3}{8\pi}\Big|^0_{-\frac{t}{2i}}\\
    =& t^{-i}\frac{t}{t^2+1}(i-1)!!\frac{3}{8\pi}\Big(1-\exp(-\frac{t^2+1}{8i^2})\Big)\\
    \geq & t^{1 -i}\frac{(i-1)!!}{100i^2}\\
\end{aligned}
\end{equation}
For odd $i$, similarly by Lemma A.6 in \citep{ALL19}, we can obtain
$$\mathbb{E}_{\bfw_0\sim\mathcal{N}(0,\bfI), b_0\sim\mathcal{N}(0,1)}[h(\alpha_1)\mathbb{1}[\alpha\geq -\frac{b_0}{t}]\mathbb{1}[|b_0|\leq \frac{t}{2i}]]=\mathbb{E}_{b_0\sim\mathcal{N}(0,1)}[p_i\cdot \mathbb{1}[|b_0|\leq \frac{t}{2i}]]\cdot \frac{x_1^i}{t^{i}}$$, 
where $$p_i=(i-1)!!\frac{\exp(-b_0^2/(2t^2))}{\sqrt{2\pi}}\sum_{r=1, r\text{ even}}^{i-1}\frac{(-1)^\frac{i-1-r}{2}}{r!!}\binom{i/2-1}{(r-1)/2}(-b_0/t)^r$$
Then we also have
$$\Big|\frac{c_{r}(-b_0/t)^{r}}{c_{r-2}(-b_0/t)^{r-2}}\Big|=\Big|(\frac{b_0}{t})^2\frac{i+1-r}{r(r-1)}\Big|\leq \frac{1}{4i}\leq\frac{1}{4}$$
Therefore, 
$$\Big|\sum_{r=1, r\text{ odd}}^{i-1}c_r(-b_0/t)^r\Big|\geq \frac{3}{4}|c_0|=\frac{3}{4}\frac{(\frac{i}{2}-1)!}{\pi(\frac{1}{2}\cdot\frac{3}{2}\cdots\frac{i-1}{2})}\geq \frac{3}{4}\frac{1}{\pi\frac{i-1}{2}}\geq \frac{3}{2\pi i}$$
\begin{equation}
\begin{aligned}
    &|\mathbb{E}_{b_0\sim\mathcal{N}(0,1)}[p_i\cdot\mathbb{1}[|b_0|/t\leq 1/(2i)]]|\cdot t^{-i}\\
    \geq &t^{-i}\cdot|\mathbb{E}_{b_0\sim\mathcal{N}(0,1)}[(i-1)!!\frac{\exp(-b_0^2/2t^2)}{\sqrt{2\pi}}\cdot\frac{3}{2\pi i}\cdot\mathbb{1}[|b_0|/t\leq 1/(2i)]]|\\
    =& t^{-i}\cdot\int_{-\frac{t}{2i}}^{\frac{t}{2i}}(i-1)!!\frac{\exp(-\frac{b_0^2}{2}(1+\frac{1}{t^2}))}{2\pi}\cdot\frac{3}{2\pi i}db_0\\
    = & t^{-i}\cdot(i-1)!!\frac{3}{4\pi^2 i}\cdot \frac{t}{\sqrt{t^2+1}}\cdot\sqrt{2\pi}\cdot\Big(2\Phi(\frac{\sqrt{t^2+1}}{2i})-1\Big)\\
    = & t^{-i}\cdot(i-1)!!\frac{3}{4\pi^2 i}\cdot \frac{t}{\sqrt{t^2+1}}\cdot\sqrt{2\pi}\cdot\frac{2\Phi(\frac{\sqrt{t^2+1}}{2})-1}{i}\\
    \geq & \frac{t^{1-i}}{\sqrt{t^2+1}}\frac{(i-1)!!}{100i^2}
\end{aligned}
\end{equation}

\begin{lemma}\label{lm: claim2_h}
For $B_i=100i^{1/2}+10\sqrt{\log(t^{i-1}\sqrt{t^2+1}/\epsilon_i^2)}$ where $\epsilon_i^2= t^{i-1}\sqrt{t^2+1}\epsilon^2$, we have
\begin{enumerate}
    \item  $\sum_{i=1}^\infty|c_i'|\cdot|\mathbb{E}_{x\sim\mathcal{N}(0,1)}[|h_i(x)|\cdot\mathbb{1}[|x|\geq b]]|\leq \frac{\epsilon}{8}\sqrt{t^2+1}$
    \item $\sum_{i=1}^\infty|c_i'|\cdot|\mathbb{E}_{x\sim\mathcal{N}(0,1)}[|h_i(b)|\cdot\mathbb{1}[|x|\geq b]]|\leq \frac{\epsilon}{8}\sqrt{t^2+1}$
    \item $\sum_{i=1}^\infty |c_i'|\mathbb{E}_{z\in\mathcal{N}(0,1)}[|h_i(z)|\mathbb{1}[|z|\leq B_i]]\leq\mathcal{C}_\epsilon(\phi,t)\sqrt{t^2+1}$
    \item $\sum_{i=1}^\infty |c_i'|\mathbb{E}_{z\in\mathcal{N}(0,1)}[|\frac{d}{dz} h_i(z)|\mathbb{1}[|z|\leq B_i]]\leq \mathcal{C}_\epsilon(\phi, t)\sqrt{t^2+1}$
\end{enumerate}

\end{lemma}

\noindent\textbf{Proof:}\\
By the definition of Hermite polynomial in Definition A.5 in \citep{ALL19}, we have
$$h_i(x)\leq \sum_{j=1}^{\left\lfloor i/2\right\rfloor}\frac{|x|^{i-2j}i^{2j}}{j!}$$
Combining (\ref{ci_qi}), we can obtain
\begin{equation}|c_i'h_i(x)|\leq O(1)|c_i|\frac{\sqrt{t^2+1}}{t^{1-i}}\frac{i^4}{i!!}\sum_{j=1}^{\left\lfloor i/2\right\rfloor}\frac{|x|^{i-2j}i^{2j}}{j!}\label{c_i'h_i}\end{equation}
(1) Let $b=100i^\frac{1}{2}\theta_i$ and $\theta_i=1+\frac{\sqrt{\log (\frac{1}{\epsilon_i^2}\frac{\sqrt{t^2+1}}{t^{1-i}})}}{10\sqrt{i}}$ for $\epsilon_i^2=\frac{\sqrt{t^2+1}}{t^{1-i}}\epsilon^2$ where $i\geq 1$, then we have
\begin{equation}
\begin{aligned}
    (\theta_i\cdot e^{-10^2\theta_i^2})^i= & \Big(\big(1+\frac{\sqrt{\log (\frac{1}{\epsilon_i^2}\frac{\sqrt{t^2+1}}{t^{1-i}})}}{10\sqrt{i}}\big)\cdot e^{-10^2}e^{-2\cdot10^2\frac{\sqrt{\log (\frac{1}{\epsilon_i^2}\frac{\sqrt{t^2+1}}{t^{2-i}})}}{10\sqrt{i}}}\cdot e^{-10^2 \frac{\log (\frac{1}{\epsilon_i^2}\frac{\sqrt{t^2+1}}{t^{2-i}})}{100i}}\Big)^i\\
    =& \epsilon_i^2\frac{t^{1-i}}{\sqrt{t^2+1}}\cdot e^{-10^2i}\cdot (1+\frac{\sqrt{\log (\frac{1}{\epsilon_i^2}\frac{\sqrt{t^2+1}}{t^{1-i}})}}{10\sqrt{i}})e^{-2\cdot 10^2 \frac{\sqrt{\log (\frac{1}{\epsilon_i^2}\frac{\sqrt{t^2+1}}{t^{1-i}}})}{10\sqrt{i}}}\\
    \leq & \frac{\epsilon_i^2}{100000^i}\frac{t^{1-i}}{\sqrt{t^2+1}}
\end{aligned}\label{theta_i_power}
\end{equation}
where the second step comes from that $(1+s)\cdot e^{-2\cdot10^4\cdot s}\leq 1$ for any $s>0$. Combining the equation C.6, C.7 in \citep{ALL19} and (\ref{theta_i_power}), we can derive
\begin{equation}
\begin{aligned}
&\sum_{i=1}^\infty|c_i'|\cdot\mathbb{E}_{x\sim\mathcal{N}(0,1)}[|h_i(z)|\cdot\mathbb{1}[|x|\geq b]]\\
\leq & \sum_{i=1}^\infty O(1)|c_i|\frac{\sqrt{t^2+1}}{t^{1-i}}\frac{i^4}{i!!}\cdot i^\frac{i}{2}\cdot 1200^i\cdot (\theta_i \cdot e^{-10^2\theta_i^{2}})^i\\
\leq & \frac{\epsilon}{8}\sqrt{t^2+1}
\end{aligned}\label{claim1_h_a_epsilon}
\end{equation}
for any $\epsilon>0$ and $t\leq O(1)$. \\
(b) Similarly, following (\ref{theta_i_power}) and (\ref{claim1_h_a_epsilon}), we have
$$\sum_{i=1}^\infty|c_i'|\cdot|\mathbb{E}_{x\sim\mathcal{N}(0,1)}[|h_i(b)|\cdot\mathbb{1}[|x|\geq b]]|\leq \sum_{i=1}^\infty O(1)\frac{\sqrt{t^2+1}}{t^{1-i}}|c_i|\frac{i^4}{i!!}\cdot e^{-\frac{b^2}{2}}(3b)^i\leq\frac{\epsilon}{8}\sqrt{t^2+1}$$
(c) Similar to (\ref{c_i'h_i}), 
\begin{equation}
    \begin{aligned}
    \sum_{i=1}^\infty |c_i'|\mathbb{E}_{z\in\mathcal{N}(0,1)}[|h_i(z)|\mathbb{1}[|z|\leq B_i]]& \leq O(1) \sum_{i=1}^\infty |c_i|\frac{i^4}{i!!}\sum_{j=0}^{\left\lfloor i/2\right\rfloor}\frac{B_i^{i-2j}i^{2j}}{j!}t^{i-1}\sqrt{t^2+1}\\
    & \leq \sum_{i=1}^\infty |c_i|(O(1)\theta_i)^i t^{i-1}\sqrt{t^2+1}\\
    &\leq \mathcal{C}_\epsilon(\phi, t)\sqrt{t^2+1},
    \end{aligned}
\end{equation}
where the step follows from Claim C.2 (c) in \citep{ALL19}.\\
(d) Since we have
\begin{equation}
    |\frac{d}{dx}h_i(x)|\leq \sum_{j=0}^{\left\lfloor i/2\right\rfloor}|x|^{i-2j}i^{2j}
\end{equation}
by Definition A.5 in \citep{ALL19}, we can derive
\begin{equation}
    \sum_{i=1}^\infty |c_i'|\mathbb{E}_{z\in\mathcal{N}(0,1)}[|\frac{d}{dz} h_i(z)|\mathbb{1}[|z|\leq B_i]]\leq \mathcal{C}_\epsilon(\phi, t)\sqrt{t^2+1}
\end{equation}

\subsubsection{Existence of a good pseudo network}
We hope to find some good pseudo network that can approximate the target network. In such a pseudo network, the activation $\mathbb{1}_{\bfx\geq0}$ is replaced by $\mathbb{1}_{\bfx^{(0)}\geq0}$ where $\bfx^{(0)}$ is the value at the random initialization. We can define a pseudo network without bias as
\begin{equation}
    g_r^{(0)}(\bfq,\bfA,\bfW,\bfV, \bfB)=\sum_{n=1}^N \bfq^\top\bfa_n\sum_{i\in[m_2]}c_{i,r}\mathbb{1}_{\bfr_{n,i}+B_{2(n,i)}\geq0}\sum_{j=1}^N a_{n,j}\sum_{l\in[m_1]}v_{i,l}\mathbb{1}_{\bfa_j\bfX\bfw_l+B_{1(j,l)}}\bfa_j\bfX\bfw_l
\end{equation}
Lemma \ref{lm: existence} shows the target function can be approximated by the pseudo network with some parameters. Lemma \ref{lm: approximation} to \ref{lm: C5} provides how the existence of such a pseudo network is developed step by step.
\begin{lemma}\label{lm: existence}
For every $\epsilon\in(0,\frac{1}{K\|\bfq\|_1 p_1p_2^2\mathcal{C}_s(\Phi, p_2\mathcal{C}_s(\phi,\|\bfA\|_\infty))\mathcal{C}_s(\phi,\|\bfA\|_\infty)\sqrt{\|\bfA\|_\infty^2+1}})$, there exists $$M=\text{poly}(\mathcal{C}_\epsilon(\Phi, \sqrt{p_2}\mathcal{C}_\epsilon(\phi,\|\bfA\|_\infty)\sqrt{\|\bfA\|_\infty^2+1}), 1/\epsilon)$$
\begin{equation}
    C=\mathcal{C}_\epsilon(\phi,\|\bfA\|_\infty)\sqrt{\|\bfA\|_\infty^2+1}
\end{equation}
\begin{equation}
    C'=10 C\sqrt{p_2}
\end{equation}
\begin{equation}
    C''=\mathcal{C}_\epsilon(\Phi, C')\sqrt{\|\bfA\|_\infty^2+1}
\end{equation}
\begin{equation}C_0=\tilde{O}(p_1^2p_2K^2 CC'')
\end{equation}such that with high probability, there exists
$\widehat{\bfW}$, $\widehat{\bfV}$ with $m_1, m_2\geq M$, 
$$\|\widehat{\bfW}\|_{2,\infty}\leq \frac{C_0}{m_1},\ \ \ \ \|\widehat{\bfV}\|_{2,\infty}\leq \frac{\sqrt{m_1}}{m_2}$$
such that
$$\mathbb{E}_{(\bfX,y)\in\mathcal{D}}\Big[\sum_{r=1}^K|f_r^*(\bfq,\bfA,\bfX)-g_r^{(0)}(\bfq,\bfA,\bfX,\widehat{\bfW}.\widehat{\bfV})|\Big]\leq \epsilon$$
$$\mathbb{E}_{(\bfX,y)\in\mathcal{D}}[|L(G^{(0)}(\bfq,\bfA,\bfX,\widehat{\bfW},\widehat{\bfV}))|]\leq OPT+\epsilon$$
\end{lemma}
\textbf{Proof:}\\
For each $\phi_{2,j}$, we can construct $h_{\phi,j}:\ \mathbb{R}^2\rightarrow [-C,C]$ where $C=\mathcal{C}_\epsilon(\phi,\|\bfA\|_\infty)\sqrt{\|\bfA\|_\infty^2+1}$ using Lemma \ref{lm: h_function} satisfying
\begin{equation}
    \mathbb{E}[h_{\phi,j}({\bfw_{2,j}^*}^\top\bfw_i^{(0)}, B_{1(n,i)}^{(0)})\mathbb{1}_{{\tilde{\bfa}_n}\bfX\bfw_i^{(0)}+B_{1(n,i)\geq0}}]=\phi_{2,j}({\tilde{\bfa}_n}\bfX\bfw_{2,j})\pm \epsilon
\end{equation}
for $i\in[m_1]$. Consider any arbitrary $\bfb\in\mathbb{R}^{m_1}$ with $v_i\in\{-1,1\}$. Define
\begin{equation}\widehat{\bfW}=\frac{(C_0C''/C)^{\frac{1}{2}}}{\epsilon_c^2 m_1}(v_i\sum_{j\in[p_2]}v_{2,j}^*h_{\phi,j}({\bfw_{2,j}^*}^\top\bfw_i^{(0)}, B_{1(i)}^{(0)})\bfe_d)_{i\in[m_1]}\label{W_hat}
\end{equation}
\begin{equation}
    \widehat{\bfV}=(C_0C''/C)^{-\frac{1}{2}}\sum_{k\in[p_1]}\frac{c_k^*}{m_2}(\bfv h(\sqrt{m_2}\sum_{j\in[p_2]}v_{1,j}^*\alpha_{i,j}, B_{2(i)}^{(0)})\sum_{r=1}^K c_{i,r})_{i\in[m_2]}\label{V_hat}
\end{equation}
Then,
\begin{equation}
    \begin{aligned}
    &g_r^{(0)}(\bfq,\bfA,\widehat{\bfW},\widehat{\bfV}, \bfB)\\
    =&\sum_{n=1}^N \bfq^\top\bfa_n\sum_{i\in[m_1]}c_{i,r}\mathbb{1}_{\bfr_{n,i}+B_{2(n,i)}\geq0}\sum_{i'\in [m_2]}\sum_{j=1}^N a_{n,j}\mathbb{1}_{\bfa_j\bfX\bfw^{(0)}_i+B_{1(j,i)}\geq0}\bfa_j\bfX\widehat{\bfW}_{i} \widehat{V}_{i,i'}\\
    =&\sum_{k\in[p_1]}\frac{c_k^*}{m_2\epsilon_c^2}\sum_{n=1}^N \bfq^\top\bfa_n\sum_{i\in[m_1]}c_{i,r}^2\mathbb{1}_{\bfr_{n,i}+B_{2(n,i)}\geq0}h(\sqrt{m_2}\sum_{j\in[p_2]}v_{1,j}^*\alpha_{i,j}, B_{2(i)}^{(0)})\sum_{j=1}^N a_{n,j}\sum_{l\in[p_2]}v_{2,l}^*\phi_{2,l}(\bfa_j\bfX\bfw_{2,l}^*)\\
    =& \sum_{k\in[p_1]}\sum_{n=1}^N \bfq^\top\bfa_n c_k^*\Phi(\sum_{j\in[p_2]}v_{1,j}^*\sum_{m=1}^N a_{m,n}\phi_{1,j}(\bfa_m\bfX\bfw_{1,j}^*))\sum_{j=1}^N a_{n,j}\sum_{l\in[p_2]}v_{2,l}^*\phi_{2,l}(\bfa_j\bfX\bfw_{2,l}^*)\\
    &\pm O(p_1p_2^2\mathcal{C}_s(\Phi, p_2\mathcal{C}_s(\phi,\|\bfA\|_\infty))\mathcal{C}_s(\phi,\|\bfA\|_\infty)\sqrt{\|\bfA\|_\infty^2+1}\epsilon)\\
    =& \sum_{n=1}^N \bfq^\top\bfa_n \sum_{k\in[p_1]}c_k^*\Phi({\tilde{\bfa}_n}\sum_{j\in[p_2]}v_{1,j}^*\phi_{1,j}(\bfA\bfX\bfw_{1,j}^*)){\tilde{\bfa}_n}\sum_{l\in[p_2]}v_{2,l}^*\phi_{2,l}(\bfA\bfX\bfw_{2,l}^*)\\
    &\pm O(\|\bfq\|_1 p_1p_2^2\mathcal{C}_s(\Phi, p_2\mathcal{C}_s(\phi,\|\bfA\|_\infty))\mathcal{C}_s(\phi,\|\bfA\|_\infty)\sqrt{\|\bfA\|_\infty^2+1}\epsilon)\\
    \end{aligned}
\end{equation}
where the first step comes from definition of $g^{(0)}$, the second step is derived from (\ref{W_hat}) and (\ref{V_hat}) and the second to last step is by Lemma \ref{lm: C5}.

\begin{lemma}\label{lm: approximation}
For every smooth function $\phi$, every $\bfw^*\in\mathbb{R}^d$ with $\|\bfw^*\|=1$, for every $\epsilon\in(0,\frac{1}{\mathcal{C}_s(\phi,\|\bfA\|_\infty)\sqrt{\|\bfA\|_\infty^2+1}})$, there exists real-valued functions $\rho(\bfv_1^{(0)}, \bfW^{(0)}, \bfB_{1(n)}^{(0)})$, $J(\tilde{\bfa}_n\bfX, \bfv_1^{(0)}, \bfW^{(0)}, \bfB_{1(n)}^{(0)})$, $R(\tilde{\bfa}_n\bfX, \bfv_1^{(0)}, \bfW^{(0)}, \bfB_{1(n)}^{(0)})$ and $\phi_\epsilon(\tilde{\bfa}_n\bfX)$ such that for every $\bfX$
$$r_{n,1}(\bfX)=\rho(\bfv_1^{(0)}, \bfW^{(0)}, \bfB_{1(n)}^{(0)})\sum_{j=1}^N a_{j,n}\phi_\epsilon(\bfa_j\bfX\bfw^*)+J(\bfX, \bfv_1^{(0)}, \bfW^{(0)}, \bfB_{1(n)}^{(0)})+R(\bfX, \bfv_1^{(0)}, \bfW^{(0)}, \bfB_{1(n)}^{(0)})$$
Moreover, letting $C=\mathcal{C}_\epsilon(\phi,\|\bfA\|_\infty)\sqrt{\|\bfA\|_\infty^2+1}$ be the complexity of $\phi$, and if $v_{1,i}\sim\mathcal{N}(0,\frac{1}{m_2})$ and $w_{i,j}^{(0)}, \bfB_{1(n)}^{(0)}\sim\mathcal{N}(0,\frac{1}{m_1})$ are at random initialization, then we have\\
1. for every fixed $\tilde{\bfa}_n\bfX$, $\rho(\bfv_1^{(0)}, \bfW^{(0)}, \bfB_{1(n)}^{(0)})$ is independent of $J(\tilde{\bfa}_n\bfX, \bfv_1^{(0)}, \bfW^{(0)}, \bfB_{1(n)}^{(0)})$.\\
2. $\rho(\bfv_1^{(0)}, \bfW^{(0)}, \bfB_{1(n)}^{(0)})\sim\mathcal{N}(0,\frac{1}{100C^2m_2})$.\\
3. $|\phi_\epsilon(\tilde{\bfa}_n\bfX\bfw_i^*)-\phi(\tilde{\bfa}_n\bfX\bfw_{i}^*)|\leq \epsilon$\\
4. with high probability, $|R(\bfX, \bfv_1^{(0)}, \bfW^{(0)}, \bfB_{1(n)}^{(0)})|\leq \tilde{O}(\frac{\|\bfA\|_\infty}{\sqrt{m_1m_2}})$, $|J(\bfX, \bfv_1^{(0)}, \bfW^{(0)}, \bfB_{1(n)}^{(0)})|\leq \tilde{O}(\frac{\|\bfA\|_\infty(1+\|\bfA\|_\infty)}{\sqrt{m_2}})$ and $\mathbb{E}[J(\bfX, \bfv_1^{(0)}, \bfW^{(0)}, \bfB_{1(n)}^{(0)})]=0$.\\
With high probability, we also have
$$\tilde{\rho}(v_1^{(0)})\sim\mathcal{N}(0,\frac{\tau}{C^2m_2})$$
$$\mathcal{W}_2(\rho|_{\bfW^{(0)},\bfB_{1(n)}^{(0)}},\tilde{\rho})\leq \tilde{O}(\frac{1}{C\sqrt{m_2}})$$
\end{lemma}
\textbf{Proof:}\\
By Lemma \ref{lm: h_function}, we have
$$\mathbb{E}_{\bfw_i^{(0)}\sim\mathcal{N}(0,\frac{\bfI}{m_1}), b_{1(n,i)}\sim\mathcal{N}(0,\frac{1}{m_1})}[h(\sqrt{m_1}{\bfw_i^{(0)}}^\top\bfw^* , b_{1(n,i)})\mathbb{1}[\tilde{\bfa}_n\bfX\bfw_i^{(0)}+b_{1(n,i)}\geq0]]=\frac{\phi_\epsilon(\tilde{\bfa}_n\bfX\bfw_i^*)}{C}$$
with $$|\phi_\epsilon(\tilde{\bfa}_n\bfX\bfw^*)-\phi(\tilde{\bfa}_n\bfX\bfw^*)|\leq \epsilon$$
and $|h(\sqrt{m_1}{\bfw_i^{(0)}}^\top\bfw^* , b_{1(n,i)})|\in[0,1]$. Note that here the $h$ function is rescaled by $1/C$.\\
Then, applying Lemma A.4 of \citep{ALL19}, we define
$$I_i=I(h(\sqrt{m_1}{\bfw_i^{(0)}}^\top\bfw^* , B_{1(n,i)}))\subset [-2,2]$$
$$S=\{i\in[m_1]: \sqrt{m_2}v_{1,i}^{(0)}\in I_i\}$$
$$s_i=s(h(\sqrt{m_1}{\bfw_i^{(0)}}^\top\bfw^* , B_{1(n,i)})), \sqrt{m_2}v_{1,i}^{(0)})$$
$$u_i=\begin{cases}\frac{s_i}{\sqrt{|S|}}, &\text{if }i\in S\\
0, &\text{if }s\notin S\end{cases}$$
where $u_i,\ i\in[m_1]$ is independent of $\bfW^{(0)}$. We can write
$$\bfW^{(0)}=\alpha \bfe_d\bfu^\top+\boldsymbol{\beta},$$
where $\alpha=\bfu^\top\bfe_d^\top\bfW^{(0)}\sim\mathcal{N}(0,1/m_1)$ and $\beta\in\mathbb{R}^{d\times m_1}$ are two independent random variables given $\bfu$. We know $\alpha$ is independent of $\bfu$. Since each $i\in S$ with probability $\tau$, we know with high probability, 
\begin{equation}
    |S|=\tilde{\Theta}(\tau m_1)\label{S_cardinality}
\end{equation}
Since $\alpha=\sum_{i\in S}u_i[\bfe_d^\top\bfW^{(0)}]_i$ and $|u_i[\bfe_d^\top\bfW^{(0)}]_i|\leq \tilde{O}(1/\sqrt{m_1|S|})$, by (\ref{S_cardinality}) and the Wasserstein distance bound of central limit theorem we know there exists $g\sim\mathcal{N}(0,\frac{1}{m_1})$ such that
$$\mathcal{W}_2(\alpha|_{\bfW^{(0)}, \bfB_{1(n)}^{(0)}},g)\leq \tilde{O}(\frac{1}{\sqrt{\tau}m_1})$$
Then,
\begin{equation}
\begin{aligned}
    r_{n,1}(\bfX)&=\sum_{j=1}^N a_{j,n}\sum_{i=1}^{m_1}v_{i,1}^{(0)}\sigma(\bfa_j\bfX\bfw_1^{(0)}+B_{1(n,i)}^{(0)})\\
    &= \sum_{j=1}^N a_{j,n}\sum_{i\notin S}v_{i,1}^{(0)}\sigma(\bfa_j\bfX\bfw_1^{(0)}+B_{1(n,i)}^{(0)})+\sum_{j=1}^N a_{j,n}\sum_{i\in S}v_{i,1}^{(0)}\sigma(\bfa_j\bfX\bfw_1^{(0)}+B_{1(n,i)}^{(0)})\\
    &=J_1+\sum_{j=1}^N a_{j,n}\sum_{i\in S}v_{i,1}^{(0)}\sigma(\bfa_j\bfX\bfw_1^{(0)}+B_{1(n,i)}^{(0)})
\end{aligned}
\end{equation}
\begin{equation}
    \begin{aligned}
    &r_{n,1}(\bfX)-J_1\\
    =& \sum_{j=1}^N a_{j,n}\sum_{i\in S}v_{i,1}^{(0)}\mathbb{1}[\bfa_j\bfX\bfw_1^{(0)}+B_{1(n,i)}^{(0)}]\frac{s_i}{2\sqrt{|S|}}\alpha +\sum_{j=1}^N a_{j,n}\sum_{i\in S}v_{i,1}^{(0)}\mathbb{1}[\bfa_j\bfX\bfw_1^{(0)}+B_{1(n,i)}^{(0)}](\bfa_j\bfX\boldsymbol{\beta}_i+B_{1(n,i)}^{(0)})\\
    =&P_1+P_2
    \end{aligned}
\end{equation}
Here, we know that since
\begin{equation}
    \mathbb{E}[v_{i,1}^{(0)}\sigma(\bfa_j\bfX\bfw_1^{(0)}+B_{1(n,i)}^{(0)})]=\mathbb{E}[v_{i,1}^{(0)}]\cdot\mathbb{E}[\sigma(\bfa_j\bfX\bfw_1^{(0)}+B_{1(n,i)}^{(0)})]=0
\end{equation}
Hence,
\begin{equation}
    \mathbb{E}[J_1]=\mathbb{E}[\sum_{j=1}^N a_{j,n}\sum_{i\notin S}v_{i,1}^{(0)}\sigma(\bfa_j\bfX\bfw_1^{(0)}+B_{1(n,i)}^{(0)})]=0
\end{equation}
Then we can derive
\begin{equation}
    P_1=\sum_{j=1}^N a_{j,n}\sum_{i\in S}\mathbb{1}[\bfa_j\bfX\bfw_1^{(0)}+B_{1(n,j)}^{(0)}]\frac{\alpha}{2\sqrt{|S|m_2}}h(\sqrt{m_1}{\bfw_i^{(0)}}^\top\bfw^* , B_{1(n,i)}) +R_1
\end{equation}
where $|R_1|\leq \tilde{O}(\sqrt{\frac{|S|}{m_1m_2}})$. We write $P_3=\frac{P_1-R_1}{\alpha}$. Then,
$$|P_3-\frac{\sqrt{|S|}}{\sqrt{m_2}C}\sum_{j=1}^N a_{j,n}\phi_\epsilon(\bfa_j\bfX\bfw^*)|\leq \tilde{O}(\|\bfA\|_\infty\frac{1}{\sqrt{m_2}})$$
$$|\frac{C\sqrt{m_2}}{\sqrt{\tau m_1}}P_1-\sum_{j=1}^N a_{j,n}\phi_\epsilon(\bfa_j\bfX\bfw^*)|\leq \tilde{O}(\|\bfA\|_\infty\frac{C}{\sqrt{\tau m_1}})$$
Define
$$\rho(\bfv_1^{(0)},\bfW^{(0)}, \bfB_{1(n)}^{(0)})=\frac{\sqrt{\tau m_1}}{C\sqrt{m_2}}\alpha\sim\mathcal{N}(0, \frac{\tau}{C^2 m_2})$$
Then,
$$P_1=\rho(\bfv_1^{(0)},\bfW^{(0)}, \bfB_{1(n)}^{(0)})\cdot \sum_{j=1}^N a_{j,n}\phi_\epsilon(\bfa_j\bfX\bfw^*)+R_1+R_2(\bfX, \bfv_1^{(0)},\bfW^{(0)}, \bfB_{1(n)}^{(0)})$$
where $|R_2|\leq \tilde{O}(\frac{1}{\sqrt{m_1m_2}})$.\\
We can also define
$$\tilde{\rho}(v_1^{(0)})=\frac{\sqrt{\tau m_1}}{C\sqrt{m_2}}g\sim\mathcal{N}(0,\frac{\tau}{C^2m_2})$$
Therefore,
$$\mathcal{W}_2(\rho|_{\bfW^{(0)},\bfB_{1(n)}^{(0)}},\tilde{\rho})\leq \tilde{O}(\frac{1}{C\sqrt{m_2}})$$
Meanwhile,
$$\bfa_j\bfX\bfw_i^{(0)}=\alpha\frac{s_i}{\sqrt{|S|}}\bfa_j\bfX\bfe_d+\bfa_j\bfX\boldsymbol{\beta}_i+B_{1(n,i)}^{(0)}=\bfa_j\bfX\boldsymbol{\beta}_i+B_{1(n,i)}^{(0)}\pm \tilde{O}(\frac{1}{\sqrt{|S|m_1}})$$
we have
$$P_2=\sum_{j=1}^N a_{j,n}\sum_{i\in S}v_{i,1}^{(0)}\mathbb{1}[\bfa_j\bfX\boldsymbol{\beta}_i+b_{1(n,j)}^{(0)}](\bfa_j\bfX\boldsymbol{\beta}_i+b_{1(n,i)}^{(0)})+R_3=J_2+R_3$$
\begin{equation}
    \mathbb{E}[J_2]=0
\end{equation}
with $|R_3|\leq \tilde{O}(\frac{\|\bfA\|_\infty}{\sqrt{m_1m_2}})$.\\
Let $J=J_1+J_2$, $R=R_1+R_2+R_3$. Then, w.h.p., $\mathbb{E}[J]=0$, $|J|\leq \tilde{O}(\frac{\|\bfA\|_\infty(1+\|\bfA\|_\infty)}{\sqrt{m_2}})$, $|R|\leq \tilde{O}(\frac{\|\bfA\|_\infty}{\sqrt{m_1m_2}})$. 

\begin{lemma}\label{lm: C3}
For every $\epsilon\in(0,\frac{1}{\mathcal{C}_s(\phi,\|\bfA\|_\infty)\sqrt{\|\bfA\|_\infty^2+1}})$, there exists real-valued functions $\phi_{1,j.\epsilon}(\cdot)$ such that 
$$|\phi_{1,j,\epsilon}(\tilde{\bfa}_n\bfX\bfw_{1,j}^*)-\phi_{1,j}(\tilde{\bfa}_n\bfX\bfw_{1,j}^*)|\leq \epsilon$$
for $j\in[p_2]$. Denote by
$$C=\mathcal{C}_\epsilon(\phi, \|\bfA\|_\infty)\sqrt{\|\bfA\|_\infty^2+1},\ C'=10C\sqrt{p_2},\ \phi_{1,j,\epsilon}(\bfa_j\bfX\bfw_{1,i}^*)=\frac{1}{C'}\phi_{1,j,\epsilon}(\bfa_j\bfX\bfw_{1,i}^*)$$
For every $i\in[m_2]$, there exist independent Gaussians

$$\alpha_{i,j}\sim\mathcal{N}(0,\frac{1}{m_2}),\ \beta_i(\bfX)\sim\mathcal{N}(0, \frac{1}{m_2}),$$
satisfying
$$\mathcal{W}_2(r_{n,i}(\bfX), \sum_{j\in[p_2]}\alpha_{i,j}\sum_{m=1}^N a_{m,n}\phi_{1,j,\epsilon}(\bfa_m\bfX\bfw_{1,i}^*)+C_i\beta_i(\bfX))\leq \tilde{O}( \frac{p_2^\frac{2}{3}}{m_1^\frac{1}{6}\sqrt{m_2}})$$
\end{lemma}
\textbf{Proof:}\\ 
Define $p_2 S$ many chunks of the first layer with each chunk corresponding to a set $S_{j,l}$, where $|S_{j,l}|=m_1/(p_2 S)$ for $j\in[p_2]$ and $l\in[S]$, such that
$$\mathcal{S}_{j,l}=\{(j-1)\frac{m_1}{p_2}+(l-1)\frac{m_1}{p_2} S+k|k\in[\frac{m_1}{p_2 S]}\}\subset[m_1]$$
By Lemma \ref{lm: approximation}, we have
\begin{equation}
    \begin{aligned}
    r_{n,i}(\bfX)=&\sum_{j\in[p_2], l\in[S]}\rho(\bfv_i^{(0)}[j,l], \bfW^{(0)}[j,l], \bfB_{1(n)}^{(0)}[j,l])\sum_{m=1}^N a_{m,n}\phi_\epsilon(\bfa_m\bfX\bfw_{1,j}^*)\\
    &+\sum_{j\in[p_2], l\in[S]}J_j(\bfX, \bfv_i^{(0)}[j,l], \bfW^{(0)}[j,l], \bfB_{1(n)}^{(0)}[j,l])+R_j(\bfX, \bfv_i^{(0)}[j,l], \bfW^{(0)}[j,l], \bfB_{1(n)}^{(0)}[j,l]),
    \end{aligned}
\end{equation}
where $\rho(\bfv_i^{(0)}[j,l], \bfW^{(0)}[j,l], \bfB_{1(n)}^{(0)}[j,l])\sim \mathcal{N}(0,\frac{1}{100C^2m_2p_2S})$. Then $\rho_j=\sum_{l\in[S]}\rho_{j,l}\sim\mathcal{N}(0,\frac{1}{C'^2m_2})$ for $C'=10C\sqrt{p_2}$. 
Define
$$J_j^S(\bfX)=\sum_{l\in[S]}J_j(\bfX, \bfv_i^{(0)}[j,l], \bfW^{(0)}[j,l], \bfB_{1(n)}^{(0)}[j,l])$$
$$R_j^S(\bfX)=\sum_{l\in[S]}R_j(\bfX, \bfv_i^{(0)}[j,l], \bfW^{(0)}[j,l], \bfB_{1(n)}^{(0)}[j,l])$$
Then there exists Gaussian random variables $\beta_j(\bfX)$ and $\beta'(\bfX)=\sum_{i\in[p_2]}\beta_j(\bfX)$ such that
$$\mathcal{W}_2(J_j^S(\bfX),\beta_j(\bfX))\leq \frac{\|\bfA\|_\infty(1+\|\bfA\|_\infty)}{\sqrt{m_2 p S}}$$
$$\mathcal{W}_2(r_{n,i}(\bfX), \sum_{j\in[p_2]}\rho_j\sum_{m=1}^N a_{m,n}\phi_{1,j,\epsilon}(\bfa_m\bfX\bfw_{1,j}^*)+\beta'(\bfX))\leq \tilde{O}(\frac{Sp_2}{\sqrt{m_1m_2}}+\frac{\sqrt{p_2}\|\bfA\|_\infty(1+\|\bfA\|_\infty)}{m_2 S})$$
We know there exists a positive constant $C_i$ such that $\beta'/ C_i\sim\mathcal{N}(0,\frac{1}{m_2})$. Let 
$\alpha_{i,j}=C'\rho_j$, $\beta_i'=\beta'/C_i$. Notice that $\mathbb{E}\big[\sum_{l\in[S], i\in[p_2]}[J_j^2(\bfX, \bfv_i^{(0)}[j,l], \bfW^{(0)}[j,l], b_1^{(0)}[j,l])]\big]=\tilde{O}(\|\bfA\|_\infty^2(1+\|\bfA\|_\infty)^2/m_2)$. Hence, we have
$$C_i\leq \tilde{O}(\|\bfA\|_\infty(1+\|\bfA\|_\infty)$$
Let $S=(m_1/p_2)^\frac{1}{3}$, we can obtain
$$\mathcal{W}_2(r_{n,i}(\bfX), \sum_{j\in[p_2]}\alpha_{i,j}\sum_{m=1}^N a_{m,n}\phi_{1,j,\epsilon}(\bfa_m\bfX\bfw_{1,i}^*)+C_i\beta_i(\bfX))\leq \tilde{O}(\frac{p_2^\frac{2}{3}}{m_1^\frac{1}{6}\sqrt{m_2}})$$

\begin{lemma}\label{lm: C4}
There exists function $h: \mathbb{R}^2\rightarrow[-C'',C'']$ for $C''=\mathcal{C}_\epsilon(\Phi,C')\sqrt{\|\bfA\|_\infty^2+1}$ such that
\begin{equation}
\begin{aligned}
    &\mathbb{E}[\mathbb{1}_{r_{n,i}(\bfX)+b_{2(n,i)}^{(0)}\geq0}h(\sqrt{m_2}\sum_{j\in[p_2]}v_{1,j}^*\alpha_{i,j}, b_{2(n,i)}^{(0)})(\sum_{j\in[p_2]}v_{2,j}^*\phi_{2,j}(\tilde{\bfa}_n\bfX\bfw_{2,j}^*))]\\
    =&\Phi(\sum_{j\in[p_2]}v_{1,j}^*\sum_{m=1}^N a_{m,n}\phi_{1,j})\sum_{j\in[p_2]}v_{2,j}^*\phi_{2,j}(\tilde{\bfa}_n\bfX\bfw_{2,j}^*))\pm \tilde{O}(p_2^2\mathcal{C}_s(\Phi, p_2\mathcal{C}_s(\phi,\|\bfA\|_\infty))\mathcal{C}_s(\phi,\|\bfA\|_\infty)\sqrt{\|\bfA\|_\infty^2+1}\epsilon)
\end{aligned}
\end{equation}
\end{lemma}
\textbf{Proof:}\\
Choose $\bfw=(\alpha_{i,1},\cdots,\alpha_{i,p_2}, \beta_i)$, $\bfx=(\sum_{m=1}^N a_{m,n}\phi_{1,1,\epsilon}, \cdots, \sum_{m=1}^N a_{m,n}\phi_{1,p_2,\epsilon}, C_i)$ and $\bfw^*=(v_{1,1}^*,\cdots, v_{1,p_2}^*,0)$. Then, $\|\bfx\|\leq O(\|\bfA\|_\infty^2+\|\bfA\|_\infty)$. By Lemma \ref{lm: h_function}, there exists $h: \mathbb{R}^{2}\rightarrow[-C'',C'']$ for $C''=\mathcal{C}_s(\Phi,C')\sqrt{\|\bfA\|_\infty^2+1}$ such that
\begin{equation}\label{capital_phi}
    \begin{aligned}
    &\mathbb{E}[\mathbb{1}_{\bfw\bfX+b_{2(n,i)}^{(0)}\geq0}h(\sqrt{m_2}\bfw^\top\bfw^*, b_{2(n,i)}^{(0)})(\sum_{j\in[p_2]}v_{2,j}^*\phi_{2,j}(\tilde{\bfa}_n\bfX\bfw_{2,j}^*))]\\
    =&\mathbb{E}_{\alpha_i,\beta_i}[\mathbb{1}_{\sum_{j\in[p_2]}\alpha_{i,j}\sum_{m=1}^N a_{m,n}\phi_{1,j,\epsilon}(\tilde{\bfa}_n\bfX\bfw_{1,i}^*)+C_i\beta_i'+b_{2(n,i)}^{(0)}\geq0}h(\sqrt{m_2}\bfw^\top\bfw^*, b_{2(n,i)}^{(0)})(\sum_{j\in[p_2]}v_{2,j}^*\phi_{2,j}(\tilde{\bfa}_n\bfX\bfw_{2,j}^*))]\\
    =&\Phi(C'\sum_{j\in[p_2]}v_{1,j}^*\sum_{m=1}^N a_{m,n}\phi_{1,j,\epsilon})\sum_{j\in[p_2]}v_{2,j}^*\phi_{2,j}(\tilde{\bfa}_n\bfX\bfw_{2,j}^*))\pm \epsilon C'''
    \end{aligned}
\end{equation}
where $$C'''=\sup |\sum_{j\in[p_2]}v_{2,j}^*\phi_{2,j}(\tilde{\bfa}_n\bfX\bfw_{2,j}^*)|\leq p_2\mathcal{C}_s(\phi,\|\bfA\|_\infty)\sqrt{\|\bfA\|_\infty^2+1}$$
By Lemma \ref{lm: C3}, we know
$$\mathcal{W}_2(r_{n,i}(\bfX), \sum_{j\in[p_2]}\alpha_{i,j}\sum_{m=1}^N a_{m,n}\phi_{1,j,\epsilon}(\bfa_m\bfX\bfw_{1,i}^*)+C_i\beta_i(\bfX))\leq \tilde{O}( \frac{p_2^\frac{2}{3}}{m_1^\frac{1}{6}\sqrt{m_2}})$$
Denote $\mathcal{H}=\{i\in[m_1]: |\sum_{j\in[p_2]}\alpha_{i,j}\sum_{m=1}^N a_{m,n}\phi_{1,j,\epsilon}(\tilde{\bfa}_n\bfX\bfw_{1,i}^*)+C_i\beta_i'|\geq \tilde{O}( \frac{2 p_2^\frac{2}{3}}{m_1^\frac{1}{6}\sqrt{m_2}})\}$. Then, for every $i\in[\mathcal{H}]$, we have that
\begin{equation}
    \mathbb{1}_{r_{n,i}(\bfX)+b_{2(n,i)}^{(0)}\geq0}=\mathbb{1}_{\sum_{j\in[p_2]}\alpha_{i,j}\sum_{m=1}^N a_{m,n}\phi_{1,j,\epsilon}(\tilde{\bfa}_n\bfX\bfw_{1,i}^*)+C_i\beta_i'+b_{2(n,i)}^{(0)}\geq0}\label{indicator_equal_3l}
\end{equation}
\begin{equation}\Pr\Big(\big|\sum_{j\in[p_2]}\alpha_{i,j}\sum_{m=1}^N a_{m,n}\phi_{1,j,\epsilon}(\tilde{\bfa}_n\bfX\bfw_{1,i}^*)+C_i\beta_i'\big|\leq \tilde{O}( \frac{2 p_2^\frac{2}{3}}{m_1^\frac{1}{6}\sqrt{m_2}})\Big)\leq \tilde{O}( \frac{2 p_2^\frac{2}{3}}{m_1^\frac{1}{6}\sqrt{m_2}})\cdot \sqrt{m_2}=\tilde{O}(\frac{2 p_2^\frac{2}{3}}{m_1^\frac{1}{6}}),\label{prob_indicator_neq}
\end{equation}
which implies with probability at least $1-2p_2^{2/3}/m_1^{1/6}$, (\ref{indicator_equal_3l}) holds.  
Therefore,
\begin{equation}
\begin{aligned}
    &\mathbb{E}[\mathbb{1}_{r_{n,i}(\bfX)+b_{2(n,i)}^{(0)}\geq0}h(\sqrt{m_2}\sum_{j\in[p_2]}v_{1,j}^*\alpha_{i,j}, b_{2(n,i)}^{(0)})(\sum_{j\in[p_2]}v_{2,j}^*\phi_{2,j}(\tilde{\bfa}_n\bfX\bfw_{2,j}^*))]\\
    =&\mathbb{E}[\mathbb{1}_{\sum_{j\in[p_2]}\alpha_{i,j}\sum_{m=1}^N a_{m,n}\phi_{1,j,\epsilon}(\tilde{\bfa}_n\bfX\bfw_{1,i}^*)+C_i\beta_i'+b_{2(n,i)}^{(0)}\geq0}h(\sqrt{m_2}\sum_{j\in[p_2]}v_{1,j}^*\alpha_{i,j}, b_{2(n,i)}^{(0)})(\sum_{j\in[p_2]}v_{2,j}^*\phi_{2,j}(\tilde{\bfa}_n\bfX\bfw_{2,j}^*))]\\
    \ &\pm \mathbb{E}[\mathbb{1}_{r_{n,i}(\bfX)+b_{2(n,i)}^{(0)}\geq0}\neq\mathbb{1}_{\sum_{j\in[p_2]}\alpha_{i,j}\sum_{m=1}^N a_{m,n}\phi_{1,j,\epsilon}(\tilde{\bfa}_n\bfX\bfw_{1,i}^*)+C_i\beta_i'+b_{2(n,i)}^{(0)}\geq0}]O(C'''C'')\\
    =&\mathbb{E}[\mathbb{1}_{\sum_{j\in[p_2]}\alpha_{i,j}\sum_{m=1}^N a_{m,n}\phi_{1,j,\epsilon}(\tilde{\bfa}_n\bfX\bfw_{1,i}^*)+C_i\beta_i'+b_{2(n,i)}^{(0)}\geq0}h(\sqrt{m_2}\sum_{j\in[p_2]}v_{1,j}^*\alpha_{i,j}, b_{2(n,i)}^{(0)})(\sum_{j\in[p_2]}v_{2,j}^*\phi_{2,j}(\tilde{\bfa}_n\bfX\bfw_{2,j}^*))]\\
    \ &\pm \tilde{O}(\frac{2p_2^{2/3}}{m_1^{1/6}}C'''C'')\\
    =&\Phi(\sum_{j\in[p_2]}v_{1,j}^*\sum_{m=1}^N a_{m,n}\phi_{1,j})\sum_{j\in[p_2]}v_{2,j}^*\phi_{2,j}(\bfx))\pm \tilde{O}(p_2^2\mathcal{C}_s(\Phi, p_2\mathcal{C}_s(\phi,\|\bfA\|_\infty))\mathcal{C}_s(\phi,\|\bfA\|_\infty)\sqrt{\|\bfA\|_\infty^2+1}\cdot \epsilon),
\end{aligned}
\end{equation}
where the first step is by Lemma \ref{lm: C3}, the second step is by (\ref{indicator_equal_3l}) and (\ref{prob_indicator_neq}) and the last step comes from (\ref{capital_phi}) and $m_1\geq M$.

\begin{lemma}\label{lm: C5}
\begin{equation}
\begin{aligned}
    &\frac{1}{m_2}\mathbb{E}[\sum_{i=1}^{m_2}\frac{c_{i,l}^2}{\epsilon_c^2}\mathbb{1}_{r_{n,i}(\bfX)+b_{2(n,i)}^{(0)}\geq0}h(\sqrt{m_2}\sum_{j\in[p_2]}v_{1,j}^*\alpha_{i,j}, b_{2(n,i)}^{(0)})(\sum_{j\in[p_2]}v_{2,j}^*\phi_{2,j}(\tilde{\bfa}_n\bfX\bfw_{2,j}^*))]\\
    =&\Phi(\sum_{j\in[p_2]}v_{1,j}^*\sum_{m=1}^N a_{m,n}\bfa_m\bfX\boldsymbol{\delta}\phi_{1,j})\sum_{j\in[p_2]}v_{2,j}^*\phi_{2,j}(\tilde{\bfa}_n\bfX\bfw_{2,j}^*))\\
    &\ \ \pm \tilde{O}(p_2^2\mathcal{C}_s(\Phi, p_2\mathcal{C}_s(\phi,\|\bfA\|_\infty))\mathcal{C}_s(\phi,\|\bfA\|_\infty)\sqrt{\|\bfA\|_\infty^2+1}\cdot\epsilon)
\end{aligned}
\end{equation}
\end{lemma}
\textbf{Proof:}\\
Recall $\tilde{\rho}(\bfv_1^{(0)})\sim\mathcal{N}(0,\frac{\tau}{C^2m_2})$. Define $\tilde{\rho}_{j,l}=\tilde{\rho}(\bfv_1^{(0)}[j,l])$. Therefore,
\begin{equation}\mathcal{W}_2(\rho_{j,l}|_{\bfW^{(0)},\bfB_{1(n)}^{(0)}},\tilde{\rho}_{j,l})\leq \tilde{O}(\frac{1}{C'\sqrt{m_2}S})\label{rho_jl_tilde_W2}
\end{equation}
\begin{equation}\mathcal{W}_2(\rho_j|_{\bfW^{(0)},\bfB_{1(n)}^{(0)}},\tilde{\rho}_j)\leq \tilde{O}(\frac{1}{C'\sqrt{m_2}})\label{rho_j_tilde_W2}
\end{equation}
where $\tilde{\rho}_j=\sum_{l\in[S]}\rho_{j,l}$. We then define $\tilde{\alpha}_{i,j}=C'\tilde{\rho}_{j}$\\
Next modify $r_{n,i}(\bfX)$. Define 
$$\tilde{r}_{n,i}(\bfX)=\frac{\sum_{m=1}^N a_{m,n}\sum_{j\in[m_1]}v_{j,i}^{(0)}\sigma(\bfa_m\bfX\bfw_i^{(0)}+b_{1(n,i)}^{(0)})}{\|\bfu\|}\mathbb{E}[\|\bfu\|]$$
where $u=(\sigma(\tilde{\bfa}_n\bfX\bfw_j^{(0)}+b_{1(n,j)}^{(0)}))_{j\in[m_1]}$. By definition, we know
$$\tilde{r}_{n,i}\sim\mathcal{N}(0, \frac{\|\tilde{\bfa}_n\|_\infty^2}{m_2}\mathbb{E}[\|\bfu\|]^2)$$
Then we have
\begin{equation}
\mathcal{W}_2(r_{n,i}(\bfX), \tilde{r}_{n,i}(\bfX))\leq \tilde{O}(\frac{\|\bfA\|_\infty\sqrt{\|\bfA\|_\infty^2+1}}{\sqrt{m_2}})\label{r_rtilde_W2}
\end{equation}
Combining (\ref{rho_jl_tilde_W2}), (\ref{rho_j_tilde_W2}), (\ref{r_rtilde_W2}) and Lemma \ref{lm: C4}, we have
\begin{equation}
\begin{aligned}
    &\frac{1}{m_2}\mathbb{E}[\sum_{i=1}^{m_2}\frac{c_{i,l}^2}{\epsilon_c^2}\mathbb{1}_{r_{n,i}(\bfX)+b_{2(n,i)}^{(0)}\geq0}h(\sqrt{m_2}\sum_{j\in[p_2]}v_{1,j}^*\alpha_{i,j}, b_{2(n,i)}^{(0)})(\sum_{j\in[p_2]}v_{2,j}^*\phi_{2,j}(\tilde{\bfa}_n\bfX\bfw_{2,j}^*))]\\
    =&\Phi(\sum_{j\in[p_2]}v_{1,j}^*\sum_{m=1}^N a_{m,n}\phi_{1,j})\sum_{j\in[p_2]}v_{2,j}^*\phi_{2,j}(\tilde{\bfa}_n\bfX\bfw_{2,j}^*))\pm \tilde{O}(p_2^2\mathcal{C}_s(\Phi, p_2\mathcal{C}_s(\phi,\|\bfA\|_\infty))\mathcal{C}_s(\phi,\|\bfA\|_\infty)(\sqrt{\|\bfA\|_\infty^2+1}\cdot\epsilon)
\end{aligned}
\end{equation}

\subsubsection{Coupling}
This section illustrates the coupling between the real and pseudo networks. We first define diagonal matrices $\bfD_{n,\bfw}$, $\bfD_{n,\bfw}+\bfD_{n,\bfw}''$, $\bfD_{n,\bfw}+\bfD_{n,\bfw}'$ for node $n$ as the sign of Relu's in the first layer at weights $\bfW^{(0)}$, $\bfW^{(0)}+\bfW^\rho$ and $\bfW^{(0)}+\bfW^\rho+\bfW'$, respectively. We also define diagonal matrices $\bfD_{n,\bfv}$, $\bfD_{n,\bfv}+\bfD_{n,\bfv}''$, $\bfD_{n,\bfv}+\bfD_{n,\bfv}'$ for node $n$ as the sign of Relu's in the second layer at weights $\{\bfW^{(0)}, \bfV^{(0)}\}$, $\{\bfW^{(0)}+\bfW^\rho, \bfV^{(0)}+\bfV^\rho\}$ and $\{\bfW^{(0)}+\bfW^\rho+\bfW', \bfV^{(0)}+\bfV^\rho+\bfV'\}$, respectively. For every $l\in[K]$, we then introduce the pseudo network and its semi-bias, bias-free version as
\begin{equation}g_l(\bfq,\bfA,\bfX,\bfW,\bfV)=\bfq^\top\bfA((\bfA(\bfA\bfX\bfW+\bfB_1)\odot(\bfD_\bfw+\bfD_\bfw')\bfV+\bfB_2)\odot(\bfD_{\bfv}+\bfD_{\bfv}'))\bfc_l
\end{equation}
\begin{equation}g_l^{(b)}(\bfq,\bfA,\bfX,\bfW,\bfV)=\bfq^\top\bfA((\bfA(\bfA\bfX\bfW+\bfB_1)\odot(\bfD_\bfw+\bfD_\bfw')\bfV)\odot(\bfD_{\bfv}+\bfD_{\bfv}'))\bfc_l
\end{equation}
\begin{equation}g_l^{(b,b)}(\bfq,\bfA,\bfX,\bfW,\bfV)=\bfq^\top\bfA((\bfA(\bfA\bfX\bfW)\odot(\bfD_\bfw+\bfD_\bfw')\bfV)\odot(\bfD_{\bfv}+\bfD_{\bfv}'))\bfc_l
\end{equation}
Lemma \ref{lm: coupling} gives the final result of coupling with added Drop-out noise. Lemma \ref{lm: sparse_sign_change_cross_term} states the sparse sign change in Relu and the function value changes of pseudo network by some update. To be more specific, Lemma \ref{lm: smoothed_real_pseudo} shows that the sign pattern can be viewed as fixed for the smoothed objective when a small update is introduced to the current weights. Lemma \ref{lm: bias_free_target} proves the bias-free pseudo network can also approximate the target function.
\begin{lemma}\label{lm: coupling}
Let $F_\bfA=(f_1,f_2,\cdots,f_K)$. With high probability, we have for any $\|\bfW'\|_{2,4}\leq \tau_w$, $\|\bfV'\|_F\leq \tau_v$, such that
\begin{equation}
    \begin{aligned}
    &f_l(\bfq,\bfA,\bfX,\bfW^{(0)}+\bfW'\boldsymbol{\Sigma},\bfV^{(0)}+\boldsymbol{\Sigma}\bfV')\\
    =& \bfq^\top\bfA(\bfA((\bfA\bfX\bfW^{(0)}+\bfB_1^{(0)})\odot \bfD_{\bfw,\bfx}^{(0)}\bfV^{(0)}+\bfB_2^{(0)})\odot\bfD_{\bfv,\bfx}^{(0)})\bfc+\bfq^\top\bfA(\bfA((\bfA\bfX\bfW')\odot\bfD_{\bfw,\bfx}^{(0)}\bfV')\odot\bfD_{\bfv,\bfx}^{(0)})\bfc_l\\
    &\pm \tilde{O}(\tau_v\frac{\sqrt{m_2}}{\sqrt{m_1}}+m_1^\frac{9}{5}\tau_w^\frac{16}{5}\sqrt{m_2}+\tau_w^\frac{8}{5}m_1^\frac{9}{10})\cdot \|\bfq^\top\bfA\|_1\|\bfA\|_\infty^2,
    \end{aligned}
\end{equation}
\end{lemma}
where we use $\bfD_{\bfw,\bfx}^{(0)}$ and $\bfD_{\bfv,\bfx}^{(0)}$ to denote the sign matrices at random initialization $\bfW^{(0)}$,  $\bfV^{(0)}$ and we let $\bfD_{\bfw,\bfx}^{(0)}+\bfD_{\bfw,\bfx}'$, $\bfD_{\bfv,\bfx}^{(0)}+\bfD_{\bfv,\bfx}'$ be the sign matrices at $\bfW+\bfW'\boldsymbol{\Sigma}$, $\bfV+\boldsymbol{\Sigma}\bfV'$.

\noindent \textbf{Proof:}\\
Since $\tilde{\bfa}_n\bfX\bfw_i^{(0)}+B_{1(n,i)}^{(0)}=\tilde{\bfa}_n\tilde{\bfX}\tilde{\bfw}_i^{(0)}$ where $\tilde{\bfw}_i^{(0)}=(\bfw_i^{(0)},B_{1(n,i)}^{(0)})\in\mathbb{R}^{d+1}$ and $\tilde{\bfX}=(\bfX,\boldsymbol{1})\in\mathbb{R}^{N\times (d+1)}$, we can ignore the bias term for simplicity.
Define
$$\bfZ= \bfA(\bfA\bfX\bfW^{(0)})\odot \bfD_{\bfw,\bfx}^{(0)}$$
$$\bfZ_1= \bfA(\bfA\bfX\bfW'\boldsymbol{\Sigma})\odot \bfD_{\bfw,\bfx}^{(0)}$$
$$\bfZ_2= \bfA(\bfA\bfX(\bfW^{(0)}+\bfW')\boldsymbol{\Sigma})\odot \bfD_{\bfw,\bfx}'$$
Then by Fact C.9 in \citep{ALL19} we have
\begin{equation}
    \begin{aligned}
    \|\bfZ_n\boldsymbol{\Sigma}\bfV'\|_2^2&\leq \sum_{i=1}^{m_2}(\bfZ_n\boldsymbol{\Sigma}\bfV'_i)^2\leq \sum_{i=1}^{m_2} \tilde{O}(\|\bfZ_n\|_\infty^2\cdot\|\bfV'_i\|_2^2)\\
    &\leq \tilde{O}(\|\bfA\|_\infty^2 m_1^{-1}\tau_v^2)
    \end{aligned}
\end{equation}
Therefore, we have $\|\bfZ_n\boldsymbol{\Sigma}\bfV'\|_2\leq \tilde{O}(\|\bfA\|_\infty m_1^{-\frac{1}{2}}\tau_v)$.\\
Let $s$ be the total number of sign changes in the first layer caused by adding $\bfW'$. Note that the total number of coordinated $i$ such that $|\tilde{\bfa}_n\bfX\bfw^{(0)}_i|\leq s''=\frac{2\tau_w}{s^\frac{1}{4}}$ is at most $s'' m_1^\frac{3}{2}$ with high probability. Since $\|\bfW'\|_{2,4}\leq \tau_w$, we must have $s\leq \tilde{O}(s''m^\frac{3}{2})=\tilde{O}(\frac{\tau_w}{s^\frac{1}{4}}m_1^\frac{3}{2})$. 
Therefore, 
$\|\bfZ_{2,n}\|_0\leq s=\tilde{O}(\tau_w^\frac{4}{5} m_1^\frac{6}{5})$. Then,
\begin{equation}
    \begin{aligned}
    \|\bfZ_{2,n}\|_2=&\|(\bfA(\bfA\bfX(\bfW^{(0)}+\bfW'\boldsymbol{\Sigma}))\odot \bfD_{\bfw,\bfx}')_n\|\\
    \leq & \Big(s\cdot \sum_{(\bfD_{\bfw,\bfx}')_n\neq0}(\bfA\bfA\bfX\bfW^{(0)}+\bfA\bfA\bfX\bfW'\boldsymbol{\Sigma})_{n,i}^4\Big)^\frac{1}{4}\\
    \leq & \Big(s\cdot \sum_{(\bfD_{\bfw,\bfx}')_n\neq0}(\bfA\bfA\bfX\bfW'\boldsymbol{\Sigma})_{n,i}^4\Big)^\frac{1}{4}\\
    \leq & s^\frac{1}{4} \|\bfA\|_\infty\tau_w\\
    \leq & \tilde{O}(\tau_w^\frac{6}{5}m_1^\frac{3}{10}\|\bfA\|_\infty)
    \end{aligned}
\end{equation}
Then we have
$$    \|\bfZ_{2,n}\boldsymbol{\Sigma}\bfV'\|_2\leq \tilde{O}(\tau_v\tau_w^\frac{6}{5}m_1^\frac{3}{10}\|\bfA\|_\infty)$$
With high probability, we have
$$\sum_{n=1}^N \bfq^\top\bfa_n\sum_{i=1}^{m_2}c_{i,l}(\sigma(r_{n,i}+r'_{n,i})-\sigma(r_{n,i}))\leq \tilde{O}(\|\bfq\|\sqrt{m_2})\|r'_{n,i}\|$$
\begin{equation}
    \begin{aligned}
    &f_l(\bfq,\bfA,\bfX,\bfW^{(0)}+\bfW'\boldsymbol{\Sigma},\bfV^{(0)}+\boldsymbol{\Sigma}\bfV')\\
    =& \sum_{n=1}^N \bfq^\top\bfa_n\sum_{i=1}^{m_2}c_{i,l}\sigma\Big((\bfZ+\bfZ_1+\bfZ_2)_n^\top (\bfV_i+(\boldsymbol{\Sigma})_i\bfV')\Big)\\
    =& \sum_{n=1}^N \bfq^\top\bfa_n\sum_{i=1}^{m_2}c_{i,l}\sigma\Big((\bfZ_n+\bfZ_{1,n}+\bfZ_{2,n})^\top \bfV_i+\bfZ_{1,n}^\top(\boldsymbol{\Sigma}\bfV')_i\Big)\pm \tilde{O}(\|\bfq\|\|\bfA\|_\infty\frac{\sqrt{m_2}}{\sqrt{m_1}}\tau_v+\sqrt{m_2}\|\bfq\|\|\bfA\|_\infty\tau_w^\frac{6}{5}m_1^\frac{3}{10})
    \end{aligned}
\end{equation}
We consider the difference between 
$$A_1=\bfq^\top\bfA(((\bfZ+\bfZ_1+\bfZ_2)\bfV^{(0)}+\bfZ_1\boldsymbol{\Sigma}\bfV')\odot (\bfD_{\bfv,\bfx}^{(0)}+\bfD_{\bfv,\bfx}''))\bfc_l$$
$$A_2=\bfq^\top\bfA(((\bfZ+\bfZ_1+\bfZ_2)\bfV^{(0)}+\bfZ_1\boldsymbol{\Sigma}\bfV')\odot \bfD_{\bfv,\bfx}^{(0)})\bfc_l$$
where $\bfD_{\bfv,\bfx}''$ is the diagonal sign change matrix from $\bfZ\bfV^{(0)}$ to $(\bfZ+\bfZ_1+\bfZ_2)\bfV^{(0)}+\bfZ_1\boldsymbol{\Sigma}\bfV'$. The difference includes three terms.
\begin{equation}
\|\bfZ_{1,n}\bfV^{(0)}\|_\infty\leq \tilde{O}(\|\bfA\|_\infty m_1^\frac{1}{4}\tau_w m_2^{-\frac{1}{2}})\label{z1n_v0}
\end{equation}
\begin{equation}\|\bfZ_{2,n}\bfV^{(0)}\|_\infty\leq \tilde{O}(\|\bfZ_{2,n}\|m_2^{-\frac{1}{2}}\sqrt{s})\leq \tilde{O}(\|\bfA\|_\infty m_1^\frac{9}{10}\tau_w^\frac{8}{5}m_2^{-\frac{1}{2}})
\end{equation}
\begin{equation}\|\bfZ_{1,n}\boldsymbol{\Sigma}\bfV'\|\leq \tilde{O}(\tau_v\|\bfA\|_\infty\tau_w m_1^\frac{1}{4})
\end{equation}
where (\ref{z1n_v0}) is by Fact C.9 in \citep{ALL19}. 
Then we have
$$|A_1-A_2|\leq \|\bfq^\top\bfA\|_1\cdot \tilde{O}(m_2^\frac{3}{2}\|\bfA\|_\infty^2(m_1^\frac{1}{4}\tau_w m_2^{-\frac{1}{2}}+m_1^\frac{9}{10}\tau_w^\frac{8}{5}m_2^{-\frac{1}{2}})^2+m_2^\frac{1}{2}\tau_v^\frac{4}{3}\|\bfA\|_\infty^\frac{4}{3}\tau_w^\frac{4}{3}m_1^\frac{1}{3})$$
From $A_2$ to our goal
$$A_3= \bfq^\top\bfA(\bfZ\bfV^{(0)}\odot\bfD_{\bfv,\bfx}^{(0)})\bfc_l+\bfq^\top\bfA(\bfA(\bfA\bfX\bfW'\odot\bfD_{\bfw,\bfx}^{(0)}\bfV')\odot\bfD_{\bfv,\bfx}^{(0)})\bfc_l$$
There are two more terms.
$$|\bfq^\top\bfA(\bfZ_2\bfV^{(0)}\odot\bfD_{\bfv,\bfx}^{(0)})\bfc_l|\leq \tilde{O}(\|\bfq^\top\bfA\|_1\|\bfZ_{2,n}\|\sqrt{s})\leq \tilde{O}(\|\bfq\|_1\|\bfA\|_\infty\tau_w^\frac{8}{5}m_1^\frac{9}{10})$$
$$|\bfq^\top\bfA(\bfZ_1\bfV^{(0)}\odot\bfD_{\bfv,\bfx}^{(0)})\bfc_l|\leq \tilde{O}(\|\bfq^\top\bfA\|_1\|\bfA\|_\infty\tau_w m_1^\frac{1}{4})$$
Therefore, we have
$$|A_2-A_3|\leq \tilde{O}(\|\bfq^\top\bfA\|_1\|\bfA\|_\infty\tau_w^\frac{8}{5}m_1^\frac{9}{10}+\|\bfq\|_1\|\bfA\|_\infty\tau_w m_1^\frac{1}{4}+\|\bfq\|_1\tau_v\|\bfA\|_\infty\tau_w m_1^\frac{1}{4})$$
Finally, we have
\begin{equation}
    \begin{aligned}
    &f_l(\bfq,\bfA,\bfX,\bfW^{(0)}+\bfW'\boldsymbol{\Sigma},\bfV^{(0)}+\boldsymbol{\Sigma}\bfV')\\
    =& \bfq^\top\bfA(\bfZ\bfV^{(0)}\odot\bfD_{\bfv,\bfx}^{(0)})\bfc_l+\bfq^\top\bfA(\bfA(\bfA\bfX\bfW'\odot\bfD_{\bfw,\bfx}^{(0)}\bfV')\odot\bfD_{\bfv,\bfx}^{(0)})\bfc_l\\
    &\pm \tilde{O}(\tau_v\frac{\sqrt{m_2}}{\sqrt{m_1}}+m_1^\frac{9}{5}\tau_w^\frac{16}{5}\sqrt{m_2}+\tau_w^\frac{8}{5}m_1^\frac{9}{10})\cdot \|\bfq\|_1\|\bfA\|_\infty
    \end{aligned}
\end{equation}

\begin{lemma}\label{lm: sparse_sign_change_cross_term}
Suppose $\tau_v\in(0,1]$, $\tau_w\in[\frac{1}{m_1^\frac{3}{2}},\frac{1}{m_1^\frac{1}{2}}]$, $\sigma_w\in[\frac{1}{m_1^\frac{3}{2}}, \frac{\tau_w}{m_1^\frac{1}{4}}]$, $\sigma_v\in(0,\frac{1}{m_2^\frac{1}{2}})]$. The perturbation matrices satisfies $\|\bfW'\|_{2,4}\leq \tau_w$, $\|\bfV'\|_F\leq \tau_v$, $\|\bfW''\|_{2,4}\leq \tau_w$, $\|\bfV''\|_F\leq \tau_v$ and random diagonal matrix $\boldsymbol{\Sigma}$ has each diagonal entry i.i.d. drawn from $\{\pm1\}$. Then with high probability, we have\\
(1) Sparse sign change
$$\|\bfD_{n,\bfw}'\|_0\leq \tilde{O}(\tau_w^\frac{4}{5}m_1^\frac{6}{5})$$
$$\|\bfD_{n,\bfv}'\|_0\leq \tilde{O}(m_2^\frac{3}{2}\sigma_v(\|\bfA\|_\infty+\|\bfA\|_\infty\tau_w m_1^\frac{1}{4})+m_2\|\bfA\|_\infty^\frac{2}{3}(\|\bfA\|_\infty\tau_v+\|\bfA\|_\infty\tau_w m_1^\frac{1}{4}(1+\tau_v))^\frac{2}{3})$$
(2) Cross term vanish
\begin{equation}
    \begin{aligned}
    &g_r(\bfq,\bfA,\bfX,\bfW^{(0)}+\bfW^\rho+\bfW'+\eta\bfW''\boldsymbol{\Sigma}, \bfV^{(0)}+\bfV^\rho+\bfV'+\eta\boldsymbol{\Sigma}\bfV'')\\
    = &g_r(\bfq,\bfA,\bfX,\bfW^{(0)}+\bfW^\rho+\bfW', \bfV^{(0)}+\bfV^\rho+\bfV')+g_r^{(b,b)}(\bfq,\bfA,\bfX,\eta\bfW''\boldsymbol{\Sigma}, \eta\boldsymbol{\Sigma}\bfV'')+g_r'
    \end{aligned}
\end{equation}
for every $r\in[K]$, where $\mathbb{E}_{\boldsymbol{\Sigma}}[g_r']=0$ and $|g_r'|\leq \eta\|\bfq^\top\bfA\|_1\|\bfA\|_\infty^2\tau_v$.
\end{lemma}
\textbf{Proof:}\\
(1) We first consider the sign changes by $\bfW^\rho$. Since $\tilde{\bfa}_n\bfX\bfw_i^{(0)}+B_{1(n,i)}^{(0)}=\tilde{\bfa}_n\tilde{\bfX}\tilde{\bfw}_i^{(0)}$ where $\tilde{\bfw}_i^{(0)}=(\bfw_i^{(0)},B_{1(n,i)}^{(0)})\in\mathbb{R}^{d+1}$ and $\tilde{\bfX}=(\bfX,\boldsymbol{1})\in\mathbb{R}^{N\times (d+1)}$, we can ignore the bias term for simplicity. We have
$$\tilde{\bfa}_n\tilde{\bfX}\tilde{\bfw}_i^{(0)}\sim \mathcal{N}(0,\frac{\|\tilde{\bfa}_n\tilde{\bfX}\|^2}{m_1})$$
$$\tilde{\bfa}_n\tilde{\bfX}\tilde{\bfw}_i^\rho\sim \mathcal{N}(0,\|\tilde{\bfa}_n\tilde{\bfX}\|^2\sigma_w^2)$$
Therefore,
$$\frac{\tilde{\bfa}_n\tilde{\bfX}\tilde{\bfw}_i^{(0)}}{\tilde{\bfa}_n\tilde{\bfX}\tilde{\bfw}_i^\rho}\sim p(z)=\frac{1}{\pi(\sigma_w\sqrt{m_1}z^2+\frac{1}{\sigma_w\sqrt{m_1}})}$$
\begin{equation}
    \begin{aligned}
    \Pr[|\tilde{\bfa}_n\tilde{\bfX}\tilde{\bfw}_i^{(0)}|\leq |\tilde{\bfa}_n\tilde{\bfX}\tilde{\bfw}_i^\rho|]&=\Pr[|z|\leq 1]\\
    &=\int_{-1}^1 \frac{1}{\pi(\sigma_w\sqrt{m_1}z^2+\frac{1}{\sigma_w\sqrt{m_1}})}dz\\
    &= \int_{-(\sigma_w^2m_1)^\frac{1}{2}}^{(\sigma_w^2m_1)^\frac{1}{2}} \frac{1}{\pi(t^2+1)}dt\\
    &= \frac{2}{\pi}\arctan{\sigma_w\sqrt{m_1}}\\
    &\leq \tilde{O}(\sigma_w\sqrt{m_1})
    \end{aligned}
\end{equation}
Then, we have
$$\|\bfD_{n,\bfw}''\|_0\leq \tilde{O}(\sigma_w m_1^\frac{3}{2})$$
$$\|\tilde{\bfa}_n\tilde{\bfX}\tilde{\bfW}^{(0)}\bfD_{n,\bfw}''\|_2\leq \tilde{O}(\|\tilde{\bfa}_n\tilde{\bfX}\|\sigma_w^\frac{3}{2} m_1^\frac{3}{4})$$
We then consider the sign changes by $\bfW'$. 
Let $s=\|\bfD_{n,\bfw}'-\bfD_{n,\bfw}''\|_0$ be the total number of sign changes in the first layer caused by adding $\bfW'$. Note that the total number of coordinated $i$ such that $|\tilde{\bfa}_n\tilde{\bfX}(\tilde{\bfW}^{(0)}+\tilde{\bfW}^\rho)_i|\leq s''=\frac{2\tau_w}{s^\frac{1}{4}}$ is at most $s'' m_1^\frac{3}{2}$ with high probability. Since $\|\bfW'\|_{2,4}\leq \tau_w$, we must have
$$s\leq \tilde{O}(s''m^\frac{3}{2})=\tilde{O}(\frac{\tau_w}{s^\frac{1}{4}}m_1^\frac{3}{2})$$
$$\|\bfD_{n,\bfw}'-\bfD_{n,\bfw}''\|_0=s\leq \tilde{O}(\tau_w^\frac{4}{5}m_1^\frac{6}{5})$$
$$\|\tilde{\bfa}_n\tilde{\bfX}(\tilde{\bfW}^{(0)}+\tilde{\bfW}^\rho)(\bfD_{n,\bfw}'-\bfD_{n,\bfw}'')\|_2\leq \tilde{O}(s^\frac{1}{4}\tau_w)\leq \tilde{O}(\tau_w^\frac{6}{5}m_1^\frac{3}{10})$$
To sum up, we have
$$\|\bfD_{n,\bfw}'\|_0\leq \tilde{O}(\sigma_w m_1^\frac{3}{2}+\tau_w^\frac{4}{5}m_1^\frac{6}{5})\leq \tilde{O}(\tau_w^\frac{4}{5}m_1^\frac{6}{5})$$
Denote $\bfz_{n,0}=\tilde{\bfa}_n\tilde{\bfX}\tilde{\bfW}^{(0)}\bfD_{n,\bfw}$ and $\bfz_{n,2}= \tilde{\bfa}_n\tilde{\bfX}(\tilde{\bfW}^{(0)}+\tilde{\bfW}^\rho+\bfW')(\bfD_{n,\bfw}+\bfD_{n,\bfw}')-\tilde{\bfa}_n\tilde{\bfX}\tilde{\bfW}^{(0)}\bfD_{n,\bfw}$.
With high probability, we know
\begin{equation}
    \begin{aligned}
    \|\bfz_{n,2}\|\leq &\|\tilde{\bfa}_n\tilde{\bfX}\bfW'\|+\|\tilde{\bfa}_n\tilde{\bfX}\tilde{\bfW}^\rho\|\\
    \leq & \tilde{O}(m_1^\frac{1}{4}\tau_w\|\bfA\|_\infty+\|\bfA\|_\infty\sigma_w m_1^\frac{1}{2})\\
    \leq & \tilde{O}(\|\bfA\|_\infty\tau_w m_1^\frac{1}{4})
    \end{aligned}
\end{equation}
Denote $\bfZ_0=(\bfz_{1,0}^\top,\cdots, \bfz_{N,0}^\top)^\top\in\mathbb{R}^{N\times m_1}$, $\bfZ_2=(\bfz_{1,2}^\top,\cdots, \bfz_{N,2}^\top)^\top\in\mathbb{R}^{N\times m_1}$. The sign change in the second layer is from $\tilde{\bfa}_n\bfZ_0\bfV^{(0)}$ to $\tilde{\bfa}_n(\bfZ_0+\bfZ_2)(\bfV^{(0)}+\bfV^\rho+\bfV')$. We have
$$\|\tilde{\bfa}_n(\bfZ_0+\bfZ_2)\bfV^\rho\|_\infty\leq \tilde{O}(\sigma_v\|\bfA\|_\infty(\|\bfz_{1,0}\|+\|\bfz_{1,2}\|))$$
$$\|\tilde{\bfa}_n(\bfZ_0+\bfZ_2)\bfV'+\tilde{\bfa}_n\bfZ_2\bfV^{(0)}\|_2\leq \tilde{O}(\|\bfA\|_\infty((\|\bfz_{1,0}\|+\|\bfz_{1,2}\|)\tau_v+\|\bfz_{1,2}\|))$$
Combining $\|\bfz_{1,0}\|\leq \tilde{O}(\|\bfA\|_\infty)$, by Claim C.8 in \citep{ALL19} we have
$$\|\bfD_{n,\bfv}'\|_0\leq \tilde{O}(m_2^\frac{3}{2}\sigma_v(\|\bfA\|_\infty^2+\|\bfA\|_\infty^2\tau_w m_1^\frac{1}{4})+m_2\|\bfA\|_\infty^\frac{2}{3}(\|\bfA\|_\infty\tau_v+\|\bfA\|_\infty\tau_w m_1^\frac{1}{4}(1+\tau_v))^\frac{2}{3})$$
\\
(2) Diagonal Cross terms.\\
Denote $\bfD_{\bfw}=(\text{diag}(\bfD_{1,\bfw})^\top,\cdots,\text{diag}(\bfD_{N,m_1})^\top)^\top\in\mathbb{R}^{N\times m_1}$ and define $\bfD_\bfw'$, $\bfD_{\bfw}''$, $\bfD_{\bfv}$, $\bfD_{\bfv}'$, $\bfD_\bfv''$ accordingly.\\
Recall
$$g_r(\bfq,\bfA,\bfX,\bfW,\bfV)=\bfq^\top\bfA((\bfA(\bfA\bfX\bfW+\bfB_1)\odot(\bfD_\bfw+\bfD_\bfw')\bfV+\bfB_2)\odot(\bfD_{\bfv}+\bfD_{\bfv}'))\bfc_r$$
$$g_r^{(b)}(\bfq,\bfA,\bfX,\bfW,\bfV)=\bfq^\top\bfA((\bfA(\bfA\bfX\bfW+\bfB_1)\odot(\bfD_\bfw+\bfD_\bfw')\bfV)\odot(\bfD_{\bfv}+\bfD_{\bfv}'))\bfc_r$$
$$g_r^{(b,b)}(\bfq,\bfA,\bfX,\bfW,\bfV)=\bfq^\top\bfA((\bfA(\bfA\bfX\bfW)\odot(\bfD_\bfw+\bfD_\bfw')\bfV)\odot(\bfD_{\bfv}+\bfD_{\bfv}'))\bfc_r$$
Then 
\begin{equation}
    \begin{aligned}
    &g_r(\bfq,\bfA,\bfX,\bfW^{(0)}+\bfW^\rho+\bfW'+\eta\bfW''\boldsymbol{\Sigma}, \bfV^{(0)}+\bfV^\rho+\bfV'+\eta\boldsymbol{\Sigma}\bfV'')\\
    =& g_r(\bfq,\bfA,\bfX,\bfW^{(0)}+\bfW^\rho+\bfW', \bfV^{(0)}+\bfV^\rho+\bfV')+g_r^{(b,b)}(\bfq,\bfA,\bfX,\eta\bfW''\boldsymbol{\Sigma}, \eta\boldsymbol{\Sigma}\bfV'')\\
    &+ g_r^{(b)}(\bfq,\bfA,\bfX,\bfW^{(0)}+\bfW^\rho+\bfW', \eta\boldsymbol{\Sigma}\bfV'')+g_r^{(b,b)}(\bfq,\bfA,\bfX,\eta\bfW''\boldsymbol{\Sigma}, \bfV^{(0)}+\bfV^\rho+\bfV')
    \end{aligned}
\end{equation}
where the last two terms are the error terms. We know that
$$\|\bfW^{(0)}\|\leq \max_{\|\bfa\|=1}\|\bfa^\top\bfW^{(0)}\|\leq \max_{\|\bfa\|=1}\sqrt{\sum_{i=1}^{m_1}(\bfa^\top\bfw_i^{(0)})^2}\leq\max_{\|\bfa\|=1}\sqrt{\sum_{i=1}^{m_1}(\frac{1}{\sqrt{m_1}})^2}=1$$
Therefore,
\begin{equation}
    \begin{aligned}
    &|g_r^{(b)}(\bfq,\bfA,\bfX,\bfW^{(0)}+\bfW^\rho+\bfW', \eta\boldsymbol{\Sigma}\bfV'')|\\
    =&\eta\sum_{n=1}^N \sum_{i=1}^{m_2}\bfq^\top\bfa_n c_{i,r} {D_{n, \bfv}}_{i}\sum_{k=1}^N a_{n,k}\sum_{l=1}^{m_1} {(\boldsymbol{\Sigma}\bfV)}_{l,i}'' {D_{k,\bfw}}_{l}(\bfa_k\bfX(\bfW^{(0)}+\bfW^\rho+\bfW')_l+{\bfB_1}_{(k,l)})\\
    =&\eta \sum_{n=1}^N \bfq^\top\bfa_n\sum_{k=1}^N a_{n,k} ((\bfa_k\bfX(\bfW^{(0)}+\bfW^\rho+\bfW')+{\bfB_1}_k)\odot {\bfD_{k,\bfw}})\boldsymbol{\Sigma}\bfV''\odot{\bfD_{n,\bfv}}\bfc_r\\
    \leq & \eta \|\bfq^\top\bfA\|_1\|\bfA\|_\infty (\|\bfA\|_\infty(m^{-\frac{1}{2}}+\sigma_w+\tau_w)+m^{-\frac{1}{2}})\tau_v m_2^\frac{1}{2}\\
    \leq & \tilde{O}(\eta\|\bfq^\top\bfA\|_1\|\bfA\|_\infty^2\tau_v m_2^\frac{1}{2}m_1^{-\frac{1}{2}}),\\
    \end{aligned}
\end{equation}
where the last step is by the value selection of $\sigma_w$, $\tau_w$ and $\tau_v$.
\begin{equation}
    \begin{aligned}
    &|g_r^{(b,b)}(\bfq,\bfA,\eta\bfW''\boldsymbol{\Sigma}, \bfX,\bfV^{(0)}+\bfV^\rho+\bfV')|\\
    =& |\eta \sum_{n=1}^N \bfq^\top\bfa_n\sum_{k=1}^N a_{n,k} (\bfa_k\bfX
    \bfW''\boldsymbol{\Sigma}\odot ({\bfD_{k,\bfw}}+{\bfD_{k,\bfw}}'))(\bfV^{(0)}+\bfV^\rho+\bfV'')\odot({\bfD_{n,\bfv}}+{\bfD_{n, \bfv}}')\bfc_r|\\
    \leq & |\eta \sum_{n=1}^N \bfq^\top\bfa_n\sum_{k=1}^N a_{n,k} (\bfa_k\bfX
    \bfW''\boldsymbol{\Sigma}\odot ({\bfD_{k,\bfw}}+{\bfD_{k,\bfw}}'))\bfV'\odot({\bfD_{n,\bfv}}+{\bfD_{n,\bfv}}')\bfc_r|\\
    &+|\eta \sum_{n=1}^N \bfq^\top\bfa_n\sum_{k=1}^N a_{n,k} (\bfa_k\bfX
    \bfW''\boldsymbol{\Sigma}\odot ({\bfD_{k,\bfw}}+{\bfD_{k,\bfw}}'))(\bfV^{(0)}+\bfV^\rho)\odot{\bfD_{n,\bfv}}\bfc_r|\\
     &+|\eta \sum_{n=1}^N \bfq^\top\bfa_n\sum_{k=1}^N a_{n,k} (\bfa_k\bfX
    \bfW''\boldsymbol{\Sigma}\odot ({\bfD_{k,\bfw}}+{\bfD_{k,\bfw}}'))(\bfV^{(0)}+\bfV^\rho)\odot{\bfD_{n,\bfv}}'\bfc_r|\\
    \leq & |\eta \|\bfq^\top\bfA\|_1\|\bfA\|_\infty^2\tau_w \tau_v m_2^\frac{1}{2}|+2|\eta \|\bfq^\top\bfA\|_1\|\bfA\|_\infty^2\tau_w m_1^\frac{1}{2}|\\
    \leq & \tilde{O}(|\eta \|\bfq^\top\bfA\|_1\|\bfA\|_\infty^2\tau_w m_1^\frac{1}{2}|)\\
    \end{aligned}
\end{equation}

\begin{lemma}\label{lm: smoothed_real_pseudo}
Denote 
\begin{equation}
    \begin{aligned}
    P_{\rho,\eta}=&F_\bfA(\bfq,\bfX,\bfW+\bfW^\rho+\eta\bfW''\boldsymbol{\Sigma}, \bfV+\bfV^\rho+\eta\boldsymbol{\Sigma}\bfV'')\\
    =& \bfq^\top\bfA(\bfA(\bfA\bfX(\bfW+\bfW^\rho+\eta\bfW''\boldsymbol{\Sigma})+\bfB_1)\odot\bfD_{\bfw, \rho,\eta}(\bfV+\bfV^\rho+\eta\boldsymbol{\Sigma}\bfV'')\odot\bfD_{\bfv,\rho,\eta})\bfc_r
    \end{aligned}
\end{equation}
\begin{equation}
    \begin{aligned}
    P_{\rho,\eta}'=&G(\bfq,\bfA,\bfX,\bfW+\bfW^\rho+\eta\bfW''\boldsymbol{\Sigma}, \bfV+\bfV^\rho+\eta\boldsymbol{\Sigma}\bfV'')\\
    =& \bfq^\top\bfA(\bfA(\bfA\bfX(\bfW+\bfW^\rho+\eta\bfW''\boldsymbol{\Sigma})+\bfB_1)\odot\bfD_{\bfw, \rho}(\bfV+\bfV^\rho+\eta\boldsymbol{\Sigma}\bfV'')\odot\bfD_{\bfv,\rho})\bfc_r
    \end{aligned}
\end{equation}
There exists $\eta_0=\frac{1}{\text{poly}(m_1,m_2)}$ such that for every $\eta\leq\eta_0$, for every $\bfW''$, $\bfV''$ that satisfies $\|\bfW''\|_{2,\infty}\leq \tau_{\bfw,\infty}$, $\|\bfV''\|_{2,\infty}\leq \tau_{\bfv,\infty}$, we have 
$$\mathbb{E}_{\bfW^\rho,\bfV^\rho}[\frac{|P_{\rho,\eta}-P_{\rho,\eta}'|}{\eta^2}]=\tilde{O}(\bfq^\top\bfA\boldsymbol{1}\|\bfA\|^4(\frac{\tau^2_{\bfw,\infty}}{\sigma_w}m_1+\frac{(\tau^2_{\bfw,\infty}+\tau^2_{\bfv,\infty}m_1^{-1})}{\sigma_v}m_2))+O_p(\eta),$$
where $O_p$ hides polynomial factor of $m_1$ and $m_2$.
\end{lemma}
\textbf{Proof:}\\
\begin{equation}
    \begin{aligned}
    P_{\rho,\eta}-P_{\rho,\eta}'&= \bfq^\top\bfA(\bfA(\bfA\bfX(\bfW+\bfW^\rho+\eta\bfW''\boldsymbol{\Sigma})+\bfB_1)\odot(\bfD_{\bfw, \rho,\eta}-\bfD_{\bfw,\rho})(\bfV+\bfV^\rho+\eta\boldsymbol{\Sigma}\bfV'')\odot\bfD_{\bfv,\rho})\bfc_r\\
    &\ +\bfq^\top\bfA(\bfA(\bfA\bfX(\bfW+\bfW^\rho+\eta\bfW''\boldsymbol{\Sigma})+\bfB_1)\odot\bfD_{\bfw, \rho,\eta}(\bfV+\bfV^\rho+\eta\boldsymbol{\Sigma}\bfV'')\odot(\bfD_{\bfv,\rho,\eta}-\bfD_{\bfv,\rho}))\bfc_r
    \end{aligned}\label{two terms}
\end{equation}
We write
$$\bfZ=\bfA(\bfA\bfX(\bfW+\bfW^\rho+\eta\bfW''\boldsymbol{\Sigma})+\bfB_1)\odot\bfD_{\bfw, \rho}$$
$$\bfZ+\bfZ'=\bfA(\bfA\bfX(\bfW+\bfW^\rho+\eta\bfW''\boldsymbol{\Sigma})+\bfB_1)\odot\bfD_{\bfw, \rho,\eta}$$
Since for all $n\in[N]$, $\|\eta(\bfA\bfA\bfX\bfW''\boldsymbol{\Sigma})_n\|_\infty\leq \eta\|\bfA\|_\infty\tau_{\bfw,\infty}$, we have
$$\|\bfZ_n'\|_\infty\leq \eta\|\bfA\|_\infty\tau_{\bfw,\infty}$$
$$\Pr_{\bfW^\rho}[Z_{n,i}'\neq0]\leq \tilde{O}(\frac{\eta\|\bfA\|_\infty\tau_{\bfw,\infty}}{\sigma_w}),\ i\in[m_1]$$
Then we have
$$\Pr[\|\bfZ_n'\|_0\geq2]\leq O_p(\eta^2)$$
Then we only need to consider the case $\|\bfZ_n'\|_0=1$. Let $Z_{n,n_i}'\neq0$. Then the first term in (\ref{two terms}),  $\bfq^\top\bfA(\bfZ'(\bfV+\bfV^\rho+\eta\boldsymbol{\Sigma}\bfV'')\odot\bfD_{\bfv,\rho})\bfc_r$ should be dealt with separately.\\
The term $\bfq^\top\bfA(\bfZ'\eta\boldsymbol{\Sigma}\bfV''\odot\bfD_{\bfv,\rho})\bfc_r)$ contributes to $O_p(\eta^3)$ to the whole term.\\
Then we have
$$\|\bfq^\top\bfA(\bfZ'\eta(\bfV+\bfV^\rho)\odot\bfD_{\bfv,\rho})\bfc_r\|\leq \tilde{O}(\eta\|\|\bfq^\top\bfA\|_1\|\|\bfA\|_\infty\tau_{\bfw,\infty})$$
We also have that
$$\tilde{O}((\frac{\eta\|\bfA\|_\infty\tau_{\bfw,\infty}}{\sigma_\bfw}m_1)^N)\leq \tilde{O}(\frac{\eta\|\bfA\|_\infty\tau_{\bfw,\infty}}{\sigma_\bfw}m_1)\leq 1$$
Therefore, the contribution to the first term is $\tilde{O}(\eta^2\|\bfq^\top\bfA\|_1\|\bfA\|_\infty^2\frac{\tau^2_{\bfw,\infty}}{\sigma_w}m_1)+O_p(\eta^3)$.\\
Denote
\begin{equation}
    \begin{aligned}
    \boldsymbol{\delta}=&\bfA(\bfA\bfX(\bfW+\bfW^\rho+\eta\bfW''\boldsymbol{\Sigma})+\bfB_1)\odot\bfD_{\bfw, \rho,\eta}(\bfV+\bfV^\rho+\eta\boldsymbol{\Sigma}\bfV'')\\
    &-\bfA(\bfA\bfX(\bfW+\bfW^\rho)+\bfB_1)\odot\bfD_{\bfw, \rho}(\bfV+\bfV^\rho)
    \end{aligned}
\end{equation}
$\boldsymbol{\delta}\in\mathbb{R}^{m_2}$ has the following terms:\\
1. $\bfZ'(\bfV+\bfV^\rho+\eta\boldsymbol{\Sigma}\bfV'')$. We have its n-th row norm bounded by $O_p(\eta)$.\\
2. $\bfZ\eta\boldsymbol{\Sigma}\bfV''$. We have its n-th row infinity norm bounded by $\tilde{O}(\|\bfA\|_\infty\eta\tau_{v,\infty}m_1^{-\frac{1}{2}})$.\\
3. $\bfA(\bfA\bfX\eta\bfW''\boldsymbol{\Sigma}\odot\bfD_{\bfw,\rho})(\bfV+\bfV^\rho)$, of which the n-th row infinity is bounded by $\tilde{O}(\|\bfA\|_\infty\eta\tau_{\bfw,\infty})$.\\
4. $ \bfA(\bfA\bfX\eta^2\bfW''\boldsymbol{\Sigma}\odot\bfD_{\bfw,\rho,\eta}\boldsymbol{\Sigma}\bfV'')$. Bounded by $O_p(\eta^2)$.\\
Therefore,
$$\|\boldsymbol{\delta}_n\|_\infty\leq \tilde{O}(\|\bfA\|_\infty\eta(\tau_{\bfv,\infty}m_1^{-\frac{1}{2}}+\tau_{\bfw,\infty}))+O_p(\eta^2)$$
Similarly, we can derive that the contribution to the second term is $\tilde{O}(\eta^2\|\bfq^\top\bfA\|_1\|\bfA\|_\infty^2\frac{(\tau^2_{\bfw,\infty}+\tau^2_{\bfv,\infty}m_1^{-1})}{\sigma_v}m_2)+O_p(\eta^3)$.

\begin{lemma}\label{lm: bias_free_target}
Let $\bfF^*_{\bfA}=(f_1^*,\cdots,f_K^*)$. Perturbation matrices $\bfW'$, $\bfV'$ satisfy
$$\|\bfW'\|_{2,4}\leq \tau_w,\ \ \|\bfV'\|_F\leq \tau_v$$
There exists $\widehat{\bfW}$ and $\widehat{\bfV}$ such that
$$\|\widehat{\bfW}\|_{2, \infty}\leq \frac{C_0}{m_1},\ \ \|\widehat{\bfV}\|_{2,\infty}\leq \frac{K\sqrt{m_1}}{m_2}$$
$$\mathbb{E}[\sum_{r=1}^{K}|f_r^*(\bfq,\bfA,\bfX,\widehat{\bfW}, \widehat{\bfV})-g_r^{(b,b)}(\bfq,\bfA,\bfX,\widehat{\bfW}, \widehat{\bfV})|]\leq \epsilon$$
$$\mathbb{E}[G^{(b,b)}(\bfq,\bfA,\bfX,\widehat{\bfW}, \widehat{\bfV})]\leq OPT+\epsilon$$
\end{lemma}
\textbf{Proof:}\\
By Lemma \ref{lm: sparse_sign_change_cross_term}, we have
$$\|\bfD_{n,\bfw}'\|_0\leq \tilde{O}(\tau_w^\frac{4}{5}m_1^\frac{6}{5})\ll \tilde{O}(m_1)$$
$$\|\bfD_{n,\bfv}'\|_0\leq \tilde{O}(m_2^\frac{3}{2}\sigma_v(\|\bfA\|_\infty^2+\|\bfA\|_\infty^2\tau_w m_1^\frac{1}{4})+m_2\|\bfA\|_\infty^\frac{2}{3}(\|\bfA\|_\infty\tau_v+\|\bfA\|_\infty\tau_w m_1^\frac{1}{4}(1+\tau_v))^\frac{2}{3})\leq \tilde{O}(m_2\|\bfA\|_\infty^2(\epsilon/C_0)^{\Theta(1)})$$
Applying Lemma \ref{lm: coupling}, we know
\begin{equation}
\begin{aligned}
    &\bfq^\top\bfA((\bfA(\bfA\bfX\widehat{\bfW})\odot(\bfD_\bfw')\widehat{\bfV})\odot(\bfD_{\bfv}))\bfc_r\\
    =& \sum_{n=1}^N \bfq^\top\bfa_n\sum_{k=1}^N a_{n,k} ((\bfa_k\bfX\widehat{\bfW})\odot {\bfD_{k, \bfw}}')\widehat{\bfV}\odot{\bfD_{n,\bfv}}\bfc_r\\
    \leq & \|\bfq^\top\bfA\|_1\|\bfA\|_\infty^2 m_1^\frac{3}{10} \frac{C_0}{m_1}\frac{K\sqrt{m_1}}{m_2}\cdot m_2\\
    \leq & \epsilon
\end{aligned}
\end{equation}
\begin{equation}
\begin{aligned}
    &\bfq^\top\bfA((\bfA(\bfA\bfX\widehat{\bfW})\odot(\bfD_\bfw)\widehat{\bfV})\odot(\bfD_{\bfv}'))\bfc_r\\
    =& \sum_{n=1}^N \bfq^\top\bfa_n\sum_{k=1}^N a_{n,k} ((\bfa_k\bfX\widehat{\bfW})\odot {\bfD_{k, \bfw}})\widehat{\bfV}\odot{\bfD_{n,\bfv}'}\bfc_r\\
    \leq & \|\bfq^\top\bfA\|_1\|\bfA\|_\infty^2 m_1^\frac{1}{2} \frac{C_0}{m_1}\frac{K\sqrt{m_1}}{m_2}\cdot m_2\cdot \|\bfA\|_\infty^2(\frac{\epsilon}{C_0})^{\Theta(1)}\\
    \leq & \epsilon
\end{aligned}
\end{equation}
Then, the conclusion can be derived.

\subsubsection{Optimization}
This section states the optimization process and convergence performance of the algorithm. Lemma \ref{lm: optimzation_either_one} shows that during the optimization, either there exists an updating direction that decreases the objective, or weight decay decreases the objective. Lemma \ref{lm: convergence} provides the convergence result of the algorithm.\\ 
\noindent Define 
\begin{equation}
\begin{aligned}
    &L'(\bfA^*, \bfA^*, \bfA^*, \lambda_t,\bfW_t,\bfV_t)\\
    =&\frac{1}{\Omega^t}\sum_{i=1}^{|\Omega^t|}\mathbb{E}_{\bfW^\rho,\bfV^\rho, \boldsymbol{\Sigma'}}[L(\lambda_t F_{\bfA^*}(\bfe_g,\bfX; \bfW^{(0)}+\bfW^\rho+\bfW_t\boldsymbol{\Sigma'}, \bfV^{(0)}+\bfV^\rho+\boldsymbol{\Sigma'}\bfV_t),y_i)]+R(\sqrt{\lambda_t}\bfW_t,\sqrt{\lambda_t}\bfV_t)
\end{aligned}
\end{equation}
where
$$R(\sqrt{\lambda}\bfW_t, \sqrt{\lambda}\bfV_t)=\lambda_v\|\sqrt{\lambda}\bfV_t\|_F^2+\lambda_w\|\sqrt{\lambda}\bfW_t\|_{2,4}^2$$
\begin{lemma}\label{lm: optimzation_either_one}
For every $\epsilon_0\in(0,1)$ and $\epsilon\in(0,\frac{\epsilon_0}{K\|\bfA\|_\infty p_1p_2^2\mathcal{C}_s(\Phi, p_2\mathcal{C}_s(\phi,\|\bfA\|_\infty))\mathcal{C}_s(\phi,\|\bfA\|_\infty)\sqrt{\|\bfA\|_\infty^2+1}})$ and $\gamma\in(0,\frac{1}{4}]$, consider any $\bfW_t,\bfV_t$ with
$$L'(\bfA^*, \bfA^*,\bfA^*, \lambda_t,\bfW_t,\bfV_t)\in[(1+\gamma)OPT+\Omega(\bfq^\top\bfA^*\boldsymbol{1}\|\bfA^*\|_\infty^4\epsilon_0/\gamma),\tilde{O}(1)]$$
With high probability on random initialization, there exists $\widehat{\bfW}$, $\widehat{\bfV}$ with $\|\widehat{\bfW}\|_F\leq 1$, $\|\widehat{\bfV}\|_F\leq 1$ such that for every $\eta\in(0,\frac{1}{\text{poly}(m_1,m_2)}]$,
\begin{equation}
    \begin{aligned}
    &\min\{\mathbb{E}_{\boldsymbol{\Sigma}}[L'(\bfA^*,\bfA^*,\bfA^*, \lambda_t, \bfW_t+\sqrt{\eta}\widehat{\bfW}\boldsymbol{\Sigma}, \bfV_t+\sqrt{\eta}\boldsymbol{\Sigma}\widehat{\bfV})], L'(\bfA^*,\bfA^*, \bfA^*, (1-\eta)\lambda_t,\bfW_t,\bfV_t)\}\\
    \leq& (1-\eta\gamma/4)L'(\bfA^*, \bfA^*, \bfA^*, \lambda_t,\bfW_t,\bfV_t)
    \end{aligned}
\end{equation}
\end{lemma}
\textbf{Proof:}\\
Recall the pseudo network and the real network for every $r\in[K]$ as
$$g_r(\bfq,\bfA^*, \bfX,\bfW',\bfV')=\bfq^\top\bfA^*(\bfA^*(\bfA^*\bfX(\bfW^{(0)}+\bfW^\rho+\bfW')+\bfB_1)\odot\bfD_{\bfw, \rho,t}(\bfV^{(0)}+\bfV^\rho+\bfV')\odot\bfD_{\bfv,\rho,t})\bfc_r$$
$$f_r(\bfq,\bfA^*, \bfX,\bfW',\bfV')=\bfq^\top\bfA^*(\bfA^*(\bfA^*\bfX(\bfW^{(0)}+\bfW^\rho+\bfW')+\bfB_1)\odot\bfD_{\bfw, \rho,\bfW'}(\bfV^{(0)}+\bfV^\rho+\bfV')\odot\bfD_{\bfv,\rho,\bfV'})\bfc_r$$
where $\bfD_{\bfw, \rho,t}$ and $\bfD_{\bfv, \rho,t}$ are the diagonal matrices at weights $\bfW^{(0)}+\bfW^\rho+\bfW_t$ and $\bfV^{(0)}+\bfV^\rho+\bfV_t$. $\bfD_{\bfw, \rho,\bfW'}$ and $\bfD_{\bfv, \rho,\bfV'}$ are the diagonal matrices at weights $\bfW^{(0)}+\bfW^\rho+\bfW'$ and $\bfV^{(0)}+\bfV^\rho+\bfV'$.\\
Denote $G(\bfq,\bfA^*,\bfX, \bfW',\bfV')=(g_1,\cdots, g_K)$, $F_{\bfA^*}(\bfq,\bfX, \bfW',\bfV')=(f_1,\cdots,f_K)$.\\
As long as $L'(\bfA^*, \bfA^*, \bfA^*, \lambda_t,\bfW_t,\bfV_t)\leq \tilde{O}(1)$, according to C.32 to C.34 in \citep{ALL19}, we have
$$\lambda_w\|\sqrt{\lambda_t}\widehat{\bfW}\|_{2,4}^4\leq \epsilon_0$$
$$\lambda_v\|\sqrt{\lambda_t}\widehat{\bfV}\|_{F}^2\leq \epsilon_0$$
$$\|\widehat{\bfW}\|_F\ll 1$$
$$\|\widehat{\bfV}\|_F\ll 1$$
The we need to study an update direction
$$\widetilde{\bfW}=\bfW_t+\sqrt{\eta}\widehat{\bfW}\boldsymbol{\Sigma}$$
$$\widetilde{\bfV}=\bfV_t+\sqrt{\eta}\boldsymbol{\Sigma}\widehat{\bfV}$$
\textbf{Changes in Regularizer.} Note that here $\bfW_t\in\mathbb{R}^{d\times m_1}$, $\bfV_t\in\mathbb{R}^{m_1\times m_2}$, $\boldsymbol{\Sigma}\in\mathbb{R}^{m_1\times m_1}$. We know that
$$\mathbb{E}_{\boldsymbol{\Sigma}}[\|\bfV_t+\sqrt{\eta}\boldsymbol{\Sigma}\widehat{\bfV}\|_F^2]=\|\bfV_t\|_F^2+\eta\|\widehat{\bfV}\|_F^2$$
$$\mathbb{E}_{\boldsymbol{\Sigma}}[\|\bfW_t+\sqrt{\eta}\widehat{\bfW}\boldsymbol{\Sigma}\|_{2,4}^4]=\sum_{i\in[m_1]}\mathbb{E}[\|\bfw_{t,i}+\sqrt{\eta}\widehat{\bfW}\boldsymbol{\Sigma}_i\|_2^4]$$
For each term $i\in[m_1]$, we can bound
$$\|\bfw_{t,i}+\sqrt{\eta}\widehat{\bfW}\boldsymbol{\Sigma}_i\|_2^2=\|\bfw_{t,i}\|_2^2+\eta\|\widehat{\bfW}\boldsymbol{\Sigma}_i\|_2^2+2\sqrt{\eta}{\bfw_{t,i}}^\top\widehat{\bfW}\boldsymbol{\Sigma}_i$$
\begin{equation}
    \begin{aligned}
    \|\bfw_{t,i}+\sqrt{\eta}\widehat{\bfW}\boldsymbol{\Sigma}_i\|_2^4&=\|\bfw_{t,i}\|_2^4+\eta^2\|\widehat{\bfW}\boldsymbol{\Sigma}_i\|_2^4+4\eta\|{\bfw_{t,i}}^\top\widehat{\bfW}\boldsymbol{\Sigma}_i\|^2+2\eta\|\bfw_{t,i}\|_2^2\|\widehat{\bfW}\boldsymbol{\Sigma}_i\|_2^2\\
    &\leq \|\bfw_{t,i}\|_2^4+6\eta\|\bfw_{t,i}\|_2^2\|\widehat{\bfW}\boldsymbol{\Sigma}_i\|_2^2+O_p(\eta^2)
    \end{aligned}
\end{equation}
Therefore, by Cauchy-Schwarz inequality, we have
$$\mathbb{E}_{\boldsymbol{\Sigma}}[\|\bfW_t+\sqrt{\eta}\widehat{\bfW}\boldsymbol{\Sigma}\|_{2,4}^4]\leq \|\bfW_t\|_{2,4}^4+6\eta\|\bfW_t\|_{2,4}^2\|\widehat{\bfW}\|_{2,4}^2+O_p(\eta^2)$$
Therefore, by $\lambda_w\|\sqrt{\lambda_t}\bfW_t\|_{2,4}^4\leq R(\sqrt{\lambda_t}\bfW_t,\sqrt{\lambda_t}\bfV_t)$, we have
\begin{equation}
    \begin{aligned}
    \mathbb{E}[R(\sqrt{\lambda_t}\widetilde{\bfW},\sqrt{\lambda_t}\widetilde{\bfV})]\leq& R(\sqrt{\lambda_t}\bfW_t,\sqrt{\lambda_t}\bfV_t)+6\eta\sqrt{\epsilon_0}\sqrt{R(\sqrt{\lambda_t}\bfW_t,\sqrt{\lambda_t}\bfV_t)}+\eta\epsilon_0\\
    \leq & R(\sqrt{\lambda_t}\bfW_t,\sqrt{\lambda_t}\bfV_t)+\frac{1}{4}\eta R(\sqrt{\lambda_t}\bfW_t,\sqrt{\lambda_t}\bfV_t)+143\eta\epsilon_0\label{c35}
    \end{aligned}
\end{equation}
\textbf{Changes in Objective.} Recall that here $\widehat{\bfW}$ and $\widehat{\bfV}$ satisfy $\tau_{\bfw,\infty}\leq \frac{1}{m_1^{\frac{999}{1000}}}$ and $\tau_{\bfv,\infty}\leq \frac{1}{m_2^\frac{999}{2000}}$. By Lemma \ref{lm: smoothed_real_pseudo}, we have for every $r\in[K]$
\begin{equation}
\begin{aligned}
    &\mathbb{E}_{\bfW^\rho,\bfV^\rho}[|f_r(\bfq,\bfA^*,\bfX,\bfW+\bfW^\rho+\widetilde{\bfW}\boldsymbol{\Sigma}, \bfV+\bfV^\rho+\boldsymbol{\Sigma'}\widetilde{\bfV})-g_r(\bfq,\bfA^*,\bfX,\bfW+\bfW^\rho+\widetilde{\bfW}\boldsymbol{\Sigma}, \bfV+\bfV^\rho+\boldsymbol{\Sigma}\widetilde{\bfV})|]\\
    &\leq \tilde{O}(\|\bfq^\top\bfA^*\|_1\|\bfA^*\|_\infty^2\epsilon_0\eta)+O_p(\eta^{1.5})
\end{aligned}
\end{equation}
By Lemma \ref{lm: sparse_sign_change_cross_term}, we have
\begin{equation}
    \begin{aligned}
    G(\bfq,\bfA^*,\bfX,\widetilde{\bfW}, \widetilde{\bfV})&= G(\bfq,\bfA^*,\bfX,\bfW_t, \bfV_t)+\eta G^{(b,b)}(\bfq,\bfA^*,\bfX,\widehat{\bfW}\boldsymbol{\Sigma},\boldsymbol{\Sigma}\widehat{\bfV})+\sqrt{\eta} G'\\
    &=F_{\bfA^*}(\bfq,\bfX,\bfW_t, \bfV_t)+\eta G^{(b,b)}(\bfq,\bfA^*,\bfX,\widehat{\bfW}\boldsymbol{\Sigma},\boldsymbol{\Sigma}\widehat{\bfV})+\sqrt{\eta} G'\\
    &=F_{\bfA^*}(\bfq,\bfX,\bfW_t, \bfV_t)+\eta G^{(b,b)}(\bfq,\bfA^*,\bfX,\widehat{\bfW},\widehat{\bfV})+\sqrt{\eta} G'
    \end{aligned}
\end{equation}
where $\mathbb{E}_{\boldsymbol{\Sigma}}[G']=0$ and $|G'|\leq\epsilon$ with high probability. By C.38 in \citep{ALL19}, we have
\begin{equation}
    \begin{aligned}
    &\mathbb{E}_{\bfW^\rho,\bfV^\rho,\boldsymbol{\Sigma}}[L(\lambda_t F_{\bfA^*}(\bfq,\bfX,\widetilde{\bfW}, \widetilde{\bfV}),y)]\\
    \leq & \mathbb{E}_{\bfW^\rho,\bfV^\rho}[L(\lambda_t F_{\bfA^*}(\bfq,\bfX,\widetilde{\bfW}, \widetilde{\bfV})+\eta F^*_{\bfA^*}(\bfq,\bfA^*,\bfX,\widetilde{\bfW}, \widetilde{\bfV}),y)]+O(\|\bfq^\top\bfA^*\|_1\|\bfA^*\|_\infty^2\epsilon_0\eta)+O_p(\eta^{1.5})
    \end{aligned}
\end{equation}
Following C.40 in \citep{ALL19}, we have
\begin{equation}
    \begin{aligned}
    &\mathbb{E}_{\bfW^\rho,\bfV^\rho}[L(\lambda_t F_{\bfA^*}(\bfq,\bfX,\bfW_t, \bfV_t)+\eta F^*_{\bfA^*}(\bfq,\bfA^*,\bfX,\bfW_t, \bfV_t),y)]\\
    \leq & (1-\eta)(2L(\lambda_t F_{\bfA^*}(\bfq,\bfX,\bfW_t, \bfV_t),y)-L((1-\eta)\lambda_t F_{\bfA^*}(\bfq,\bfX,\bfW_t, \bfV_t),y))+\eta L( F^*_{\bfA^*}, y)+O_p(\eta^2)
    \end{aligned}
\end{equation}
\textbf{Putting all of them together.} Denote
\begin{equation}
    c_1=\frac{1}{|\Omega^t|}\sum_{i=1}^{|\Omega|}\mathbb{E}_{\bfW^\rho,\bfV^\rho,\boldsymbol{\Sigma},\boldsymbol{\Sigma'}}[L(\lambda_t F_{\bfA^*}(\bfe_g,\bfX,\bfW^{(0)}+\bfW^\rho+\widetilde{\bfW}\boldsymbol{\Sigma'},\bfV^{(0)}+\bfV^\rho+\boldsymbol{\Sigma'}\widetilde{\bfV}),y_i)]
\end{equation}
\begin{equation}
    c_1'=\mathbb{E}_{\boldsymbol{\Sigma}}[L'(\bfA^*,\bfA^*,\bfA^*,  \lambda_t,\widetilde{\bfW},+\widetilde{\bfV})]=c_1+\mathbb{E}_{\boldsymbol{\Sigma}}[R(\sqrt{\lambda_t} \widetilde{\bfW},\sqrt{\lambda}\widetilde{\bfV})]
\end{equation}
\begin{equation}
    c_2=\frac{1}{|\Omega^t|}\sum_{i=1}^{|\Omega|}\mathbb{E}_{\bfW^\rho,\bfV^\rho}[L((1-\eta)\lambda_t F_{\bfA^*}(\bfe_g,\bfX,\bfW^{(0)}+\bfW^\rho+\bfW_t\boldsymbol{\Sigma'},\bfV^{(0)}+\bfV^\rho+\boldsymbol{\Sigma'}\bfV_t),y_i)]
\end{equation}
\begin{equation}
    c_2'=L'(\bfA^*,\bfA^*,\bfA^*,  (1-\eta)\lambda_t,\bfW_t,\bfV_t)=c_2+R(\sqrt{(1-\eta)\lambda_t} \bfW_t,\sqrt{(1-\eta)\lambda_t}\bfV_t)
\end{equation}
\begin{equation}
    c_3=\frac{1}{|\Omega^t|}\sum_{i=1}^{|\Omega|}\mathbb{E}_{\bfW^\rho,\bfV^\rho}[L(\lambda_t F_{\bfA^*}(\bfe_g,\bfX,\bfW^{(0)}+\bfW^\rho+\bfW_t\boldsymbol{\Sigma'},\bfV^{(0)}+\bfV^\rho+\boldsymbol{\Sigma'}\bfV_t),y_i)]
\end{equation}
\begin{equation}
    c_3'=L'(\bfA^*, \bfA^*,\bfA^*,  \lambda_t,\bfW_t,\bfV_t)=c_3+R(\sqrt{\lambda_t} \bfW_t,\sqrt{\lambda}\bfV_t)
\end{equation}
Then, following from C.38 to C.42 in \citep{ALL19}, we have
\begin{equation}
    c_1'\leq (1-\eta)(2c_3'-c_2')+\frac{\eta\gamma}{4}c_3'+\eta(OPT+O(\|\bfq^\top\bfA^*\|_1\|\bfA^*\|_\infty^4\epsilon_0/\gamma))+O_p(\eta^{1.5}),
\end{equation}
which implies
$$\min\{c_1',c_2'\}\leq (1-\eta\frac{1}{2}+\frac{\eta\gamma}{8})c_3'+\eta\frac{1}{2}OPT+O(\|\bfq^\top\bfA^*\|_1\|\bfA^*\|_\infty^2\eta\epsilon_0/\gamma)+O_p(\eta^{1.5})$$
Note that the equation C.35 of \citep{ALL19}, i.e, (\ref{c35}) in this work, is modified as 
\begin{equation}
    c_1'-c_1\leq (1+\frac{\eta \gamma}{4})(c_3'-c_3)+O(\eta\epsilon_0/\gamma)
\end{equation}
if $\gamma<\epsilon_0/\Omega(1)$. 
As long as $c_3'\geq (1+\gamma)OPT+\Omega(\|\bfq^\top\bfA^*\|_1\|\bfA^*\|_\infty^2)$, we have
$$\min\{c_1',c_2'\}\leq (1-\eta\frac{\gamma}{4})c_3'$$

\begin{lemma}\label{lm: convergence}
Note that the three sampled aggregation matrices in a three-layer learner network can be be different. We denote them as ${\bfA^{t(1)}}$, $\bfA^{t(2)}$ and ${\bfA^{t(3)}}$. Let $\bfW_t$, $\bfV_t$ be the updated weights trained using $\bfA^*$ and let $\bfW_t'$, $\bfV_t'$ be the updated weights trained using $\bfA^{t(i)},\ i\in[3]$. With probability at least $99/100$, the algorithm converges in $TT_w=\text{poly}(m_1,m_2)$ iterations to a point with $\eta\in(0,\frac{1}{\text{poly}(m_1, m_2, \|\bfA^*\|_\infty,K)})$
$$L'(\bfA^*, \bfA^*, \bfA^*, \lambda_t,\bfW_t,\bfV_t)\leq (1+\gamma)OPT+\epsilon_0$$
If \begin{equation}
    \begin{aligned}
    &L'(\bfA^{t(1)},\bfA^{t(2)}, \bfA^{t(3)}, \lambda_t,\bfW_t',\bfV_t')\\
    =&\frac{1}{|\Omega^t|}\sum_{i=1}^{|\Omega^t|}\mathbb{E}_{\bfW^\rho,\bfV^\rho, \boldsymbol{\Sigma}}[L(\lambda_t F_{\bfA^{(1)},\bfA^{(2)},\bfA^{(3)}}(\bfq, \bfX_i, \bfW^{(0)}+\bfW^\rho+\bfW_t'\boldsymbol{\Sigma'}, \bfV^{(0)}+\bfV^\rho+\boldsymbol{\Sigma'}\bfV_t'),y_i)]\\
    &+R(\sqrt{\lambda_t}\bfW_t',\sqrt{\lambda_t}\bfV_t'),
    \end{aligned}
\end{equation} 
where
\begin{equation}
    F_{\bfA^{t(1)}, \bfA^{t(2)},\bfA^{t(3)}}(\bfq,\bfX,\bfW,\bfV)=\bfq^\top\bfA^{t(3)}\sigma(\bfA^{t(2)}\sigma(\bfA^{t(1)}\bfX\bfW+\bfB_1)\bfV+\bfB_2)\bfC,\label{eqn:F_A1A2A3}
\end{equation} 
we also have
\begin{equation}L'(\bfA^{t(1)},\bfA^{t(2)},\bfA^{t(3)}, \lambda_{T-1}, \bfW_T', \bfV_T')\leq L'(\bfA^*, \bfA^*, \bfA^*, \lambda_T, \bfW_T,\bfV_T)+\lambda_{T-1}\cdot O(\text{poly}(\epsilon))\leq (1+\gamma)OPT+\epsilon_0\end{equation}
\end{lemma}
\textbf{Proof:}\\
By Lemma \ref{lm: optimzation_either_one}, we know that as long as $L'(\bfA^*, \bfA^*,\bfA^*,  \lambda_t,\bfW_t,\bfV_t)\in[(1+\gamma)OPT+\Omega(\bfq^\top\bfA^*\boldsymbol{1}\|\bfA^*\|_\infty^4\epsilon_0/\gamma),\tilde{O}(1)]$, then there exists $\|\widehat{\bfW}\|_F\leq 1$, $\|\widehat{\bfV}\|_F\leq 1$ such that either
\begin{equation}
    \mathbb{E}_{\boldsymbol{\Sigma},\boldsymbol{\Sigma'}}[L'(\bfA^*, \bfA^*, \bfA^*, \lambda_t, \bfW_t\boldsymbol{\Sigma'}+\sqrt{\eta}\widehat{\bfW}\boldsymbol{\Sigma}\boldsymbol{\Sigma'}, \boldsymbol{\Sigma'}\bfV_t+\sqrt{\eta}\boldsymbol{\Sigma'}\boldsymbol{\Sigma}\widehat{\bfV})]\leq (1-\eta\gamma/4)L'(\bfA^*, \bfA^*, \bfA^*, \lambda_t,\bfW_t,\bfV_t)
\end{equation}
or
\begin{equation}
    L'(\bfA^*, \bfA^*, \bfA^*, (1-\eta)\lambda_t,\bfW_t,\bfV_t)\leq (1-\eta\gamma/4)L'(\bfA^*, \bfA^*, \bfA^*, \lambda_t,\bfW_t,\bfV_t)
\end{equation}
Denote $\bfW=\bfW^{(0)}+\bfW^\rho+\bfW_t\boldsymbol{\Sigma'}+\sqrt{\eta}\widehat{\bfW}\boldsymbol{\Sigma}\boldsymbol{\Sigma'}$, $\bfV=\bfV^{(0)}+\bfV^\rho+\boldsymbol{\Sigma'}\bfV_t+\sqrt{\eta}\boldsymbol{\Sigma'}\boldsymbol{\Sigma}\widehat{\bfV}$. Note that 
\begin{equation}
    \frac{\partial L}{\partial \bfw_j}=\sum_{i=1}^K \frac{\partial L}{\partial f_i}\frac{\partial f_i}{\partial \bfw_j}
\end{equation}
\begin{equation}
\begin{aligned}
    &\frac{\partial}{\partial \bfw_j}f_r(\bfq,\bfA^*,\bfX,\bfW^{(0)}+\bfW^\rho+\bfW_t\boldsymbol{\Sigma'}+\sqrt{\eta}\widehat{\bfW}\boldsymbol{\Sigma}\boldsymbol{\Sigma'},\bfV^{(0)}+\bfV^\rho+\boldsymbol{\Sigma'}\bfV_t+\sqrt{\eta}\boldsymbol{\Sigma'}\boldsymbol{\Sigma}\widehat{\bfV} )\\
    =& \sum_{n=1}^N \bfq^\top\bfa_n\sum_{i=1}^{m_2}c_{i,r}\mathbb{1}_{r_{n,i}+B_{2(n,i)}\geq0}\sum_{k=1}^N a_{n,k}v_{j,i}\mathbb{1}_{\tilde{\bfa^*}_{n}\bfX\bfw_k+B_{1(n,k)}\geq0}(\tilde{\bfa^*}_{n}\bfX)^\top,
\end{aligned}
\end{equation}
which implies $\frac{\partial F}{\partial\bfw_t}$, $\frac{\partial^2 F}{\partial\bfw_t^2}$, $\frac{\partial^3 F}{\partial\bfw_t^3}$ are summations of $\mathbb{1}$, $\delta$, $\delta'$ functions and their multiplications. It can be found that no $\delta(x)\delta'(x)$, $\delta(x)^2$ or $\delta'^2(x)$ exist in these terms. Therefore, by $\int_{-\infty}^{\infty}\delta(t)f(t)dt=f(0)$ and  $\int_{-\infty}^{\infty}\delta'(t)f(t)dt=-f'(0)$, we can obtain that the value of the third-order derivative w.r.t. $\bfW^\rho$ of $\mathbb{E}_{\bfW^\rho,\bfV^\rho, \boldsymbol{\Sigma}}[L(\lambda_t F_{\bfA^*}(\bfe_g, \bfX, \bfW^{(0)}+\bfW^\rho+\bfW_t\boldsymbol{\Sigma}, \bfV^{(0)}+\bfV^\rho+\boldsymbol{\Sigma}\bfV_t),y)]$ is proportional to $\text{poly}(\|\bfA^*\|_\infty,K)$, some certain value of the probability density function of $\bfW^\rho$ and its derivative, i.e., $\text{poly}(\sigma_w^{-1})$. Similarly, the value of the third-order derivative w.r.t. $\bfW^\rho$ of $\mathbb{E}_{\bfW^\rho,\bfV^\rho, \boldsymbol{\Sigma}}[L(\lambda_t F_{\bfA^*}(\bfe_g, \bfX, \bfW^{(0)}+\bfW^\rho+\bfW_t\boldsymbol{\Sigma}, \bfV^{(0)}+\bfV^\rho+\boldsymbol{\Sigma}\bfV_t),y)]$ is polynomially depend on $\sigma_v^{-1}$ and $\|\bfA^*\|_\infty$. By the value selection of $\sigma_w$ and $\sigma_v$, we can conclude that $L'$ is $B=\text{poly}(m_1,m_2, \|\bfA^*\|_\infty,K)$ second-order smooth.   \\
By Fact A.8 in \citep{ALL19}, it satisfies with $\eta\in(0,\frac{1}{\text{poly}(m_1,m_2,\|\bfA^*\|_\infty,K)})$
\begin{equation}
    \lambda_{\min}(\nabla^2 L'(\bfA^*,\bfA^*, \bfA^*, \lambda_{t-1}, \bfW_t,\bfV_t))<-\frac{1}{(m_1m_2)^8}
\end{equation}
Meanwhile, for $t\geq 1$, by the escape saddle point theorem of Lemma A.9 in \citep{ALL19}, we know with probability at least $1-p$, $    \lambda_{\min}(\nabla^2 L'(\bfA^*,\bfA^*, \bfA^*, \lambda_{t-1}, \bfW_t,\bfV_t))>-\frac{1}{(m_1m_2)^8}$ holds. Choosing $p=\frac{1}{100T}$, then this holds for $t=1,2,\cdots,T$ with probability at least 0.999.Therefore, for $t=1,2,\cdots,T$, the first case cannot happen, i.e., as long as $L'(\bfA^*, \bfA^*, \bfA^*, \lambda_t,\bfW_t,\bfV_t)\geq (1+\gamma)OPT+\Omega(\bfq^\top\bfA^*\boldsymbol{1}\|\bfA^*\|_\infty^4\epsilon_0/\gamma)$,
\begin{equation}
    L'(\bfA^*,\bfA^*, \bfA^*,(1-\eta)\lambda_t,\bfW_t,\bfV_t)\leq (1-\eta\gamma/4)L'(\bfA^*, \bfA^*, \bfA^*, \lambda_t,\bfW_t,\bfV_t)
\end{equation}
On the other hand, for $t=1,2,\cdots,T-1$, as long as $L'\leq \tilde{O}(1)$, by Lemma A.9 in \citep{ALL19}, we have
\begin{equation}
    L'(\bfA^*, \bfA^*, \bfA^*, \lambda_t,\bfW_{t+1},\bfV_{t+1})\leq L'(\bfA^*, \bfA^*, \bfA^*, \lambda_t,\bfW_t,\bfV_t)+(m_1m_2)^{-1}
\end{equation}
By $L'(\bfA^*,\bfA^*,\bfA^*, \lambda_1,\bfW_0,\bfV_0)\leq \tilde{O}(1)$ with high probability, we have
$L'(\bfA^*,\bfA^*,\bfA^*, \lambda_t,\bfW_t,,\bfV_t)\leq \tilde{O}(1)$ with high probability for $t=1,2,\cdots,T$. Therefore, after $T=\tilde{\Theta}(\eta^{-1}\log\frac{\log m}{\epsilon_0})$ rounds of weight decay, we have $L'(\bfA^*, \bfA^*, \bfA^*, \lambda_t,\bfW_t,\bfV_t)\leq (1+\gamma)OPT+\Omega(\bfq^\top\bfA^*\boldsymbol{1}\|\bfA^*\|_\infty^4\epsilon_0/\gamma)$. Rescale down $\epsilon_0$ and we can obtain our final result.\\
Consider $L'(\bfA^{t(1)},\bfA^{t(2)},\bfA^{t(3)}, \lambda_t,\bfW_t',\bfV_t')$. Let $\bfw_i$, $\bfv_i$ be the output weights updated with all the aggregation matrices equal to $\bfA^*$, and let $\bfw_i'$, $\bfv_i'$ be the output weights updated with our sampling strategy in Section \ref{sec:sampling}. We know that
\begin{equation}
\begin{aligned}
    \|\bfw_i-{\bfw_i}'\|&\lesssim\sum_{t=0}^{T-1} \|\eta\sum_{l=0}^{T_w-1}\sum_{n=1}^N \bfq^\top{\bfa_n}^* \sum_{i=1}^{m_2}c_{i,r}\mathbb{1}[{\bfa^*}_n\sigma(\bfA^*\bfX\bfW)\bfv_i\geq0]\sum_{k=1}^N {a_{n,k}}^*v_{j,i}\mathbb{1}[{\bfa^*}_k\bfX\bfw_j\geq0]({\bfa^*}_k-{\bfa_k}^{t(1)})\bfX\|\\
    & \leq \frac{1}{\text{poly}(m_1, m_2)}\cdot\frac{1}{\text{poly}(\epsilon)}\text{poly}(m_1,m_2)\epsilon_c\|\bfA^*\|_\infty\cdot\text{poly}(\epsilon)=O(\epsilon)\label{w-w'}
\end{aligned}
\end{equation}
\begin{equation}
\begin{aligned}
    \|\bfv_i-\bfv_i'\|&\lesssim\sum_{t=0}^{T-1}\|\eta\sum_{l=0}^{T_w-1}\sum_{n=1}^N \bfq^\top\bfa_n^* \sum_{i=1}^{m_2}c_{i,r}\mathbb{1}[{\bfa^*}_n\sigma(\bfA^*\bfX\bfW)\bfv_i\geq0]({\bfa^*}_n\sigma(\bfA^*\bfX\bfW)-\bfa_n^{t(2)}\sigma(\bfA^{t(1)}\bfX\bfW'))\|\\
    & \leq \frac{1}{\text{poly}(m_1, m_2)}\cdot\frac{1}{\text{poly}(\epsilon)}\text{poly}(m_1,m_2)\epsilon_c\|\bfA^*\|_\infty\cdot\text{poly}(\epsilon)=O(\epsilon)\label{v-v'}
\end{aligned}
\end{equation}
With a slight abuse of notation, for $r\in[K]$, we denote
\begin{equation}
    f_r(\bfq,\bfA^{t(1)}, \bfA^{t(2)}, \bfA^{t(3)}, \bfX,\bfW_t',\bfV_t')=\bfq^\top\bfA^{t(3)}\sigma(\bfA^{t(2)}\sigma(\bfA^{t(1)}\bfX\bfW+\bfB_1)\bfV+\bfB_2)\bfc_r\label{eqn:def_f_r}
\end{equation}
The difference between $f_r(\bfq,\bfA^*,\bfX,\bfW_t,\bfV_t)$ and $f_r(\bfq,\bfA^{t(1)}, \bfA^{t(2)}, \bfA^{t(3)}, \bfX,\bfW_t',\bfV_t')$ is caused by $\|\bfA^*-\bfA^{t(1)}\|_\infty$, $\|\bfA^*-\bfA^{t(2)}\|_\infty$, $\|\bfA^*-\bfA^{t(3)}\|_\infty$, $\bfw_i^{(t)}-{\bfw_i^{(t)}}'$ and $\bfv_i^{(t)}-{\bfv_i^{(t)}}'$. Following the proof in Lemma \ref{lm: sampling}, we can easily obtain that if $|p_l-p_l^*|\leq p_l^*\cdot O(\text{poly}(\epsilon))$ and $l_i\geq |\mathcal{N}_i|/(1+\frac{c_1\cdot\text{poly}(\epsilon)}{p_l^* L \Phi(L,i)})$, it can be derived that $\|\bfA^*-\bfA^{(1)}\|_\infty\leq O(\text{poly}(\epsilon))$,  $\|\bfA^*-\bfA^{(2)}\|_\infty\leq O(\text{poly}(\epsilon))$ and $\|\bfA^*-\bfA^{(3)}\|_\infty\leq  O(\text{poly}(\epsilon))$. Then, by (\ref{w-w'}) and (\ref{v-v'}), we have
\begin{equation}
    \begin{aligned}
    &|\bfq^\top\bfA^*\sigma(\bfA^*\sigma(\bfA^*\bfX\bfW)\bfV)\bfc_r-\bfq^\top\bfA^*\sigma(\bfA^{(2)}\sigma(\bfA^{(1)}\bfX\bfW')\bfV')\bfc_r|\\
    \leq & \Big|\sum_{n=1}^N \bfq^\top{\bfa^*}_n \sum_{i=1}^{m_2}c_{i,r}|\sigma({\bfa^*}_n\sigma(\bfA^*\bfX\bfW)\bfv_i)-\sigma({\bfa_n^{(2)}}\sigma(\bfA^{(1)}\bfX\bfW')\bfv_i')|\Big|\\
    \leq & \Big|\sum_{n=1}^N \bfq^\top{\bfa^*}_n \sum_{i=1}^{m_2}c_{i,r}|{\bfa^*}_n\sigma(\bfA^*\bfX\bfW)\bfv_i-{\bfa_n^{(2)}}\sigma(\bfA^{(1)}\bfX\bfW')\bfv_i'|\Big|\\
    \leq & \Big|\sum_{n=1}^N \bfq^\top{\bfa^*}_n \sum_{i=1}^{m_2}c_{i,r}|({{\bfa^*}_n}-{\bfa_n^{(2)}})\sigma(\bfA^*\bfX\bfW)\bfv_i|\Big|+\Big|\sum_{n=1}^N \bfq^\top{\bfa^*}_n \sum_{i=1}^{m_2}c_{i,r}|{\bfa_n^{(2)}}(\sigma(\bfA^*\bfX\bfW)\bfv_i-\sigma(\bfA^{(1)}\bfX\bfW')\bfv_i'|)\Big|\\
    \leq & \Big|\sum_{n=1}^N \bfq^\top{\bfa^*}_n \sum_{i=1}^{m_2}c_{i,r}|({{\bfa^*}_n}-{\bfa_n^{(2)}})\sigma(\bfA^*\bfX\bfW)\bfv_i|\Big|\\
    &\ \ +\Big|\sum_{n=1}^N \bfq^\top{\bfa^*}_n \sum_{i=1}^{m_2}c_{i,r}|{\bfa_n^{(2)}}((\sigma(\bfA^*\bfX\bfW)-\sigma(\bfA^{(1)}\bfX\bfW'))\bfv_i+\sigma(\bfA^{(1)}\bfX\bfW')(\bfv_i-\bfv_i'))|\Big|\\
    \leq &\Big|\sum_{n=1}^N \bfq^\top{\bfa^*}_n \sum_{i=1}^{m_2}c_{i,r}|({{\bfa^*}_n}-{\bfa_n^{(2)}})\sigma(\bfA^*\bfX\bfW)\bfv_i|\Big|+\Big|\sum_{n=1}^N \bfq^\top{\bfa^*}_n \sum_{i=1}^{m_2}c_{i,r}|{\bfa_n^{(2)}}\sigma(\bfA^{(1)}\bfX\bfW')(\bfv_i-\bfv_i')|\Big|\\
    &+\Big|\sum_{n=1}^N \bfq^\top{\bfa^*}_n \sum_{i=1}^{m_2}c_{i,r}\big|\sum_{k=1}^N {a_{n,k}''}\sum_{l=1}^{m_1}v_{i,l}|({\bfa^*}_k-\bfa_k')\bfX\bfw_l+\bfa_k'\bfX(\bfw_l-\bfw_l')|\big|\Big|\\
    \leq&  O(\text{poly}(\epsilon)).
    \end{aligned}
\end{equation}
Hence, 
\begin{equation}
    \begin{aligned}
    &|\bfe_g^\top\bfA^*\sigma(\bfA^*\sigma(\bfA^*\bfX\bfW)\bfV)\bfc_r-\bfe_g^\top\bfA^{(3)}\sigma(\bfA^{(2)}\sigma(\bfA^{(1)}\bfX\bfW')\bfV')\bfc_r|\\
    \leq &|\bfe_g^\top\bfA^*\sigma(\bfA^*\sigma(\bfA^*\bfX\bfW)\bfV)\bfc_r-\bfe_g^\top\bfA^*\sigma(\bfA^{(2)}\sigma(\bfA^{(1)}\bfX\bfW')\bfV')\bfc_r|+|\bfe_g^\top(\bfA^*-\bfA^{(3)})\sigma(\bfA^{(2)}\sigma(\bfA^{(1)}\bfX\bfW)\bfV)\bfc_r|\\
    \leq & O(\text{poly}(\epsilon)).\label{fA-fA'}
    \end{aligned}
\end{equation}
which implies 
\begin{equation}
    L'(\bfA^{t(1)},\bfA^{t(2)}, \bfA^{t(3)}, \lambda_{T-1}, \bfW_T', \bfV_T')\leq L'(\bfA^*, \bfA^*, \bfA^*, \lambda_T, \bfW_T,\bfV_T)+\lambda_{T-1}\cdot O(\text{poly}(\epsilon))\leq (1+\gamma)OPT+\epsilon_0
\end{equation}

\noindent\textbf{Proof of Theorem \ref{thm: node_3}:}\\
By Lemma \ref{lm: convergence}, we have that the algorithm converges in $TT_w$ iterations to a point
$$L'(\bfA^{t(1)}, \bfA^{t(2)}, \bfA^{t(3)}, \lambda_t,\bfW_t,\bfV_t)\leq (1+\gamma)OPT+\epsilon_0$$
We know w.h.p., among $\tilde{O}(1/\epsilon_0^2)$ choices of $j$, 
\begin{equation}\label{choices_L_F}
\min_j\{\mathbb{E}_{\bfW^\rho,\bfV^\rho,\boldsymbol{\Sigma},\ \bfz\in\Omega}L(\lambda_{T-1}F_{\bfA^*}(\bfe_g,\bfX, \bfW^{(0)}+\bfW^{\rho,j}+\bfW_T\boldsymbol{\Sigma},\bfV^{(0)}+\bfV^{\rho,j}+\boldsymbol{\Sigma}\bfV_T)\}\leq (1+\gamma)OPT+\epsilon_0
\end{equation}
Then we have
\begin{equation}\label{W_t_24}
\|\bfW_T\|_{2,4}\leq \epsilon_0^\frac{1}{4}\tau_w'
\end{equation}
\begin{equation}\label{V_t_F}
    \|\bfV_T\|_F\leq \epsilon_0^\frac{1}{2}\tau_v'
\end{equation}
By Lemma \ref{lm: coupling}, we know that
\begin{equation}
\begin{aligned}
    &f_r(\bfe_g, \bfA^*,\bfX_i, \bfW^{(0)}+\bfW^\rho+\bfW_T\boldsymbol{\Sigma},\bfV^{(0)}+\bfV^\rho+\boldsymbol{\Sigma}\bfV_T, \bfB)\\
    =& f_r(\bfe_g, \bfA^*,\bfX_i, \bfW^{(0)}+\bfW^\rho,\bfV^{(0)}+\bfV^\rho, \bfB)+g_r^{(b,b)}(\bfe_g, \bfA^*,\bfX_i, \bfW_T,\bfV_T, \bfB)\pm\frac{\epsilon}{K}
\end{aligned}
\end{equation}
Denote $\bfr'=\bfA^*\sigma(\bfA^*\bfX(\bfW^{(0)}+\bfW^\rho)+\bfB_1)(\bfV^{(0)}+\bfV^\rho)$. Then, 
\begin{equation}
    \begin{aligned}
    \|\bfr'\|\leq \|\bfA^*\|(\|\bfA^*\|_\infty\cdot \tilde{O}(1))\cdot \tilde{O}(1)\leq \|\bfA^*\|_\infty
    \end{aligned}
\end{equation}
Therefore, 
\begin{equation}
    \begin{aligned}
    &|f_r(\bfe_g, \bfA^*,\bfX_i, \bfW^{(0)}+\bfW^\rho,\bfV^{(0)}+\bfV^\rho, \bfB)|\\
    =& |\bfe_g^\top \bfA^*\sigma(\bfr'+\bfB_2)\bfc_r|\\
    \leq & \tilde{O}(\|\bfA^*\|_\infty(\|\bfA^*\|_\infty+1)\epsilon_c)
    \end{aligned}
\end{equation}
We also have
\begin{equation}
    \begin{aligned}
        &|g_r^{(b,b)}(\bfe_g, \bfA^*,\bfX_i, \bfW_T,\bfV_T, \bfB)|\\
        \leq &|\bfe_g^\top\bfA^*\bfA^*(\bfA^*\bfX\bfW_T\odot\bfD_{\bfw,\bfx}^{(0)}\bfV_T)\odot\bfD_{\bfv,\bfx}^{(0)}\bfc_r|\\
        \leq & \|\bfA^*\|_\infty^2\tau_v'\tau_w' m_1^\frac{1}{4}\sqrt{m_2}\epsilon_c\\
        \leq & C_0\|\bfA^*\|_\infty^2
    \end{aligned}
\end{equation}
Hence, 
\begin{equation}
    f_r(\bfe_g, \bfA^*,\bfX_i, \bfW^{(0)}+\bfW^\rho+\bfW_T\boldsymbol{\Sigma},\bfV^{(0)}+\bfV^\rho+\boldsymbol{\Sigma}\bfV_T, \bfB)\leq \tilde{O}(\|\bfA^*\|_\infty^2(\epsilon_c+C_0))
\end{equation}
Combining (\ref{eqn:def_f_r}, \ref{fA-fA'}), we can obtain
\begin{equation}
    f_r(\bfe_g, {\bfA^t}^{(1)},{\bfA^t}^{(2)}, {\bfA^t}^{(3)}, \bfX_i, \bfW^{(0)}+\bfW^\rho+\bfW_T\boldsymbol{\Sigma},\bfV^{(0)}+\bfV^\rho+\boldsymbol{\Sigma}\bfV_T, \bfB)\leq \tilde{O}(\|\bfA^*\|_\infty^2(\epsilon_c+C_0))
\end{equation}
as long as $\|\bfA^*-{\bfA^t}^{(1)}\|_\infty\leq\text{poly}(\epsilon)$,  $\|\bfA^*-{\bfA^t}^{(2)}\|_\infty\leq\text{poly}(\epsilon)$ and $\|\bfA^*-{\bfA^t}^{(3)}\|_\infty\leq\text{poly}(\epsilon)$.\\
For any given $\{\bfX_i, y_i\}_{i=1}^{|\Omega|}$, the dependency  between $y_i$, $y_j$, where $i,j\in|\Omega|$ can be considered in two steps.
\begin{figure}[htbp]
\centering
    \subfigure{
        \begin{minipage}{0.4\textwidth}
        \centering
        \includegraphics[width=1\textwidth]{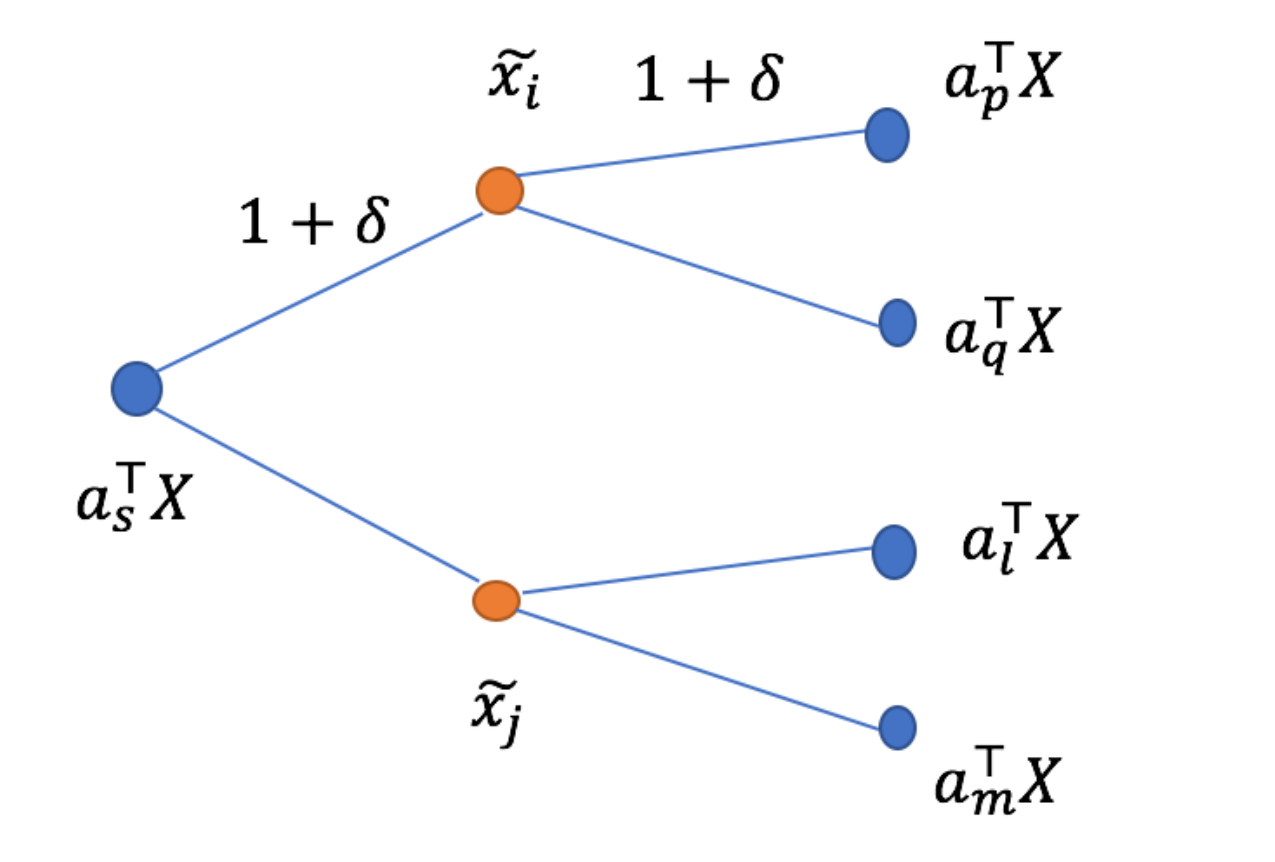}
        \end{minipage}
    }
    ~
        \subfigure{
        \begin{minipage}{0.4\textwidth}
        \centering
        \includegraphics[width=1\textwidth]{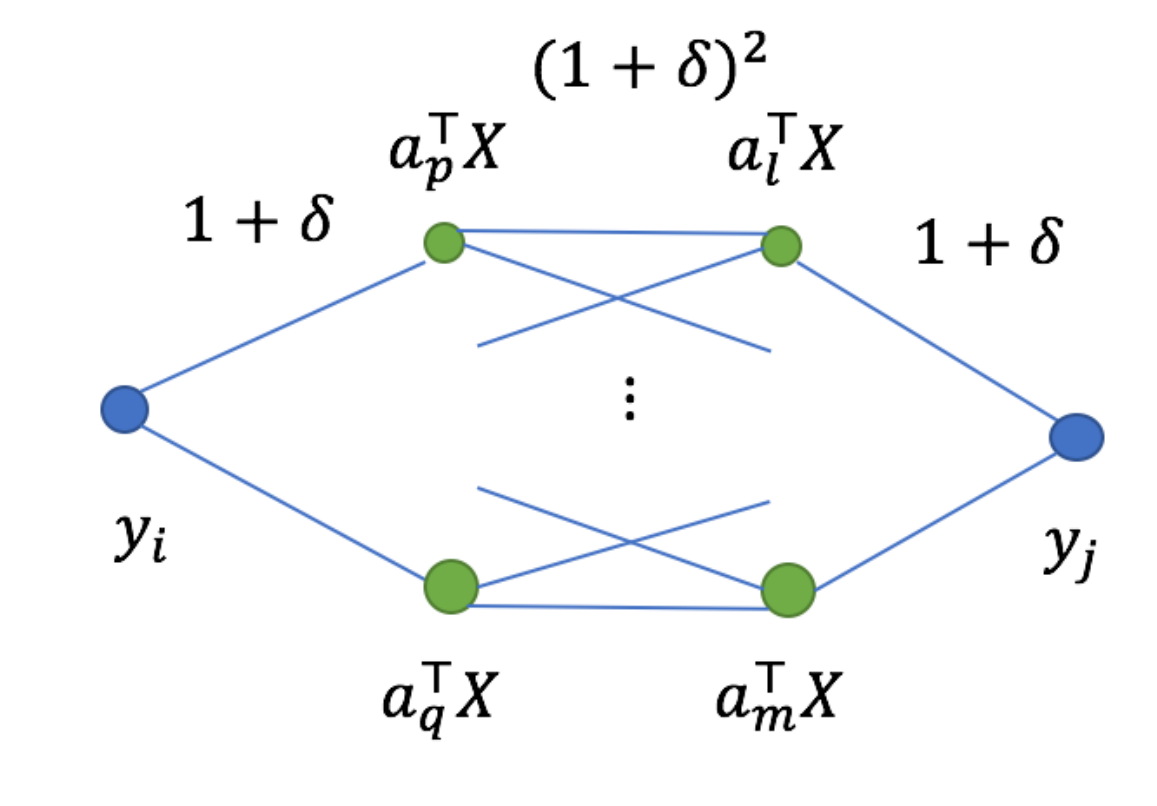}
        \end{minipage}
    }
\caption{(a) Dependency between $\bfa_s\bfX$ and $\bfa_p\bfX$ (b) Dependency between $y_i$ and $y_j$}\label{fig: dependency}
\end{figure}
Figure \ref{fig: dependency}(a) shows $\bfa_i\bfX$ is dependent with at most $(1+\delta)^2$ $\bfa_j\bfX's$. This is because each $\bfa_i\bfX$ is determined by at most $(1+\delta)$ row vector $\tilde{\bfx}_l's$, while each $\tilde{\bfx}_l$ is contained by at most $(1+\delta)$ $\bfa_p\bfX's$. Similarly, $y_i$ is determined by at most $(1+\delta)$ $\bfa_p\bfX's$ and by Figure \ref{fig: dependency}(b) we can find $y_i$ is dependent with at most $(1+\delta)^4$ $y_j$ (including $y_i$). Since the matrix $\bfA^*$ shares the same non-zero entries with $\bfA$, the output with $\bfA^*$ indicates the same dependence.\\
Denote $u_i=1/|\Omega^t|\sum_{i=1}^{|\Omega^t|}|L(\lambda_{T-1} F_{\bfA^*}(\bfe_g, \bfX, \bfW^{(0)}+\bfW^\rho+\bfW_T\boldsymbol{\Sigma},\bfV^{(0)}+\bfV^\rho+\boldsymbol{\Sigma}\bfV_T),y_i)-\mathbb{E}_{(\bfe_g,\bfX,y)\in\mathcal{D}}[L(\lambda_{T-1} F_{\bfA^*}(\bfe_g,\bfX, \bfW^{(0)}+\bfW^\rho+\bfW_T\boldsymbol{\Sigma},\bfV^{(0)}+\bfV^\rho+\boldsymbol{\Sigma}\bfV_T),y_i)]$. Then, $\mathbb{E}[u_i]=0$. Since that $L$ is 1-lipschitz smooth and $L(\boldsymbol{0}^K,y)\in[0,1]$, we have
\begin{equation}
\begin{aligned}
    &|L(\lambda_{T-1} F_{\bfA^*}(\bfe_g, \bfX, \bfW^{(0)}+\bfW^\rho+\bfW_T\boldsymbol{\Sigma},\bfV^{(0)}+\bfV^\rho+\boldsymbol{\Sigma}\bfV_T),y_i)-L(\boldsymbol{0}^K,y_i)|\\
    \leq&\|(\lambda_{T-1} F_{\bfA^*}(\bfe_g, \bfX, \bfW^{(0)}+\bfW^\rho+\bfW_T\boldsymbol{\Sigma},\bfV^{(0)}+\bfV^\rho+\boldsymbol{\Sigma}\bfV_T),y_i)-(\boldsymbol{0}^K,y_i)\|\\
    \leq &\tilde{O}(\sqrt{K}\|\bfA^*\|_\infty^2(\epsilon_c+C_0))
\end{aligned}
\end{equation}
Then,
$$|u_i|\leq 2\sqrt{K}\|\bfA^*\|_\infty^2(\epsilon_c+C_0)$$
\begin{equation}
    \mathbb{P}(|u_i|\geq t)\leq 1\leq \exp(1-\frac{t^2}{4K\|\bfA^*\|_\infty^4(\epsilon_c+C_0)^2})
\end{equation}
Then, $u_i$ is a sub-Gaussian random variable. We have $\mathbb{E} e^{s u_i}\leq e^{\|\bfA^*\|_\infty^4(\epsilon_c+C_0)^2 s^2}$. By Lemma 7 in \citep{ZWLC20}, we have
$$\mathbb{E}e^{s\sum_{i=1}^{|\Omega|} u_i}\leq e^{(1+\delta)^4K\|\bfA^*\|_\infty^4(\epsilon_c+C_0)^2|\Omega| s^2}$$
Therefore, 
\begin{equation}
    \mathbb{P}\Big(\Big|\sum_{i=1}^{|\Omega|} \frac{1}{|\Omega|}u_i\Big|\geq k\Big)\leq \exp(\|\bfA^*\|_\infty^4(\epsilon_c+C_0)^2K(1+\delta)^4|\Omega| s^2-|\Omega| ks)
\end{equation}
for any $s>0$. Let $s=\frac{k}{2\|\bfA^*\|_\infty^4(\epsilon_c+C_0)^2K(1+\delta)^4}$, $k=\|\bfA^*\|_\infty^4(\epsilon_c+C_0)^2K\sqrt{\frac{(1+\delta)^4\log N}{|\Omega|}}$, we can obtain
\begin{equation}
    \mathbb{P}\Big(\Big|\sum_{i=1}^{|\Omega|} \frac{1}{|\Omega|}u_i\Big|\geq k\Big)\leq \exp(-\|\bfA^*\|_\infty^4(\epsilon_c+C_0)^2K\log N)\leq N^{-K}
\end{equation}
Therefore, with probability at least $1-N^{-K}$, we have
\begin{equation}
\begin{aligned}
    &\Big|\mathbb{E}_{(\bfe_g,\bfX,y)\sim\mathcal{D}}[L(\lambda_{T-1} F_{\bfA^*}(\bfe_g,\bfX, \bfW^{(0)}+\bfW^\rho+\bfW_T\boldsymbol{\Sigma},\bfV^{(0)}+\bfV^\rho+\boldsymbol{\Sigma}\bfV_T),y_i)]\\
    &-\frac{1}{|\Omega^t|}\sum_{i=1}^{|\Omega^t|} L(\lambda_{T-1} F_{\bfA^*}(\bfe_g,\bfX, \bfW^{(0)}+\bfW^\rho+\bfW_T\boldsymbol{\Sigma},\bfV^{(0)}+\bfV^\rho+\boldsymbol{\Sigma}\bfV_T),y_i)\Big|\\
    \leq& \epsilon_0
\end{aligned}
\end{equation}
as long as $|\Omega|\geq \tilde{\Theta}(\epsilon_0^{-2}\|\bfA^*\|_\infty^8(1+p_1^4p_2^5\mathcal{C}_\epsilon(\phi,\|\bfA^*\|_\infty)\mathcal{C}_\epsilon(\Phi, \sqrt{p_2}\mathcal{C}_\epsilon(\phi, \|\bfA^*\|_\infty))(\|\bfA^*\|_\infty+1)^4K^6(1+\delta)^4\log N)$, i.e.,
\begin{equation}
    \mathbb{E}_{(\bfe_g, \bfX,y)\sim\mathcal{D}}[L(\lambda_{T-1} F_{\bfA^*}(\bfe_g,\bfX, \bfW^{(0)}+\bfW^\rho+\bfW_T\boldsymbol{\Sigma},\bfV^{(0)}+\bfV^\rho+\boldsymbol{\Sigma}\bfV_T),y_i)]\leq (1+\gamma)OPT+ \epsilon_0\leq (1+\epsilon_0)OPT+ \epsilon_0
\end{equation}

\end{document}